\documentclass[english]{article}
\usepackage{geometry}

\usepackage[T1]{fontenc}
\usepackage[latin9]{inputenc}
\usepackage{bm}
\usepackage{amsmath,mathtools}
\usepackage{amssymb}
\usepackage[unicode=true,
 bookmarks=false,
 breaklinks=false,pdfborder={0 0 1},colorlinks=false]
 {hyperref}
\hypersetup{
 colorlinks,citecolor=blue,filecolor=blue,linkcolor=blue,urlcolor=blue}

\makeatletter
\usepackage{amsthm}
\usepackage{cite}  
\usepackage{comment}
\usepackage{natbib}
\usepackage{booktabs}

\usepackage{graphicx}

\usepackage[linesnumbered,ruled,vlined]{algorithm2e}

\SetCommentSty{mycommfont}
\usepackage{algorithmic}

\usepackage{float}
\usepackage{multirow}
 
\usepackage{dsfont}
\usepackage{tcolorbox}

\usepackage{color}
\definecolor{yxc}{RGB}{255,0,0}
\definecolor{yjc}{RGB}{125,0,0}
\definecolor{ytw}{RGB}{255,69,0}
\definecolor{gen}{RGB}{0,0,200}

\allowdisplaybreaks

    \DeclareMathOperator{\ind}{\mathds{1}}  % Indicator

\newcommand{\mymid}{\,|\,}

\definecolor{yanxi}{RGB}{0,200,100}

\usepackage{lipsum}

\title{Instance-dependent Convergence Theory for Diffusion Models}

\author{Yuchen Jiao \thanks{Department of Statistics, The Chinese University of Hong Kong, Hong Kong; Email: \texttt{\{yuchenjiao,genli\}@cuhk.edu.hk}.}\and Gen Li \footnotemark[1]
}

\makeatother

\begin{document}

\theoremstyle{plain} \newtheorem{lemma}{\textbf{Lemma}}\newtheorem{proposition}{\textbf{Proposition}}\newtheorem{theorem}{\textbf{Theorem}}

\theoremstyle{assumption}\newtheorem{assumption}{\textbf{Assumption}}
\theoremstyle{remark}\newtheorem{remark}{\textbf{Remark}}
\theoremstyle{definition}\newtheorem{definition}{\textbf{Definition}}
\newtheorem{example}{\textbf{Example}}

\maketitle 

%%discussion about the smoothness
%%comparison with RK
%%explain nonsmoothness

\begin{abstract}
Score-based diffusion models have demonstrated outstanding empirical performance in machine learning and artificial intelligence, particularly in generating high-quality new samples from complex probability distributions. 
Improving the theoretical understanding of diffusion models, with a particular focus on the convergence analysis, has attracted significant attention.
In this work, we develop a convergence rate that is adaptive to the smoothness of different target distributions, referred to as instance-dependent bound.
Specifically, we establish an iteration complexity of $\min\{d,d^{2/3}L^{1/3},d^{1/3}L\}\varepsilon^{-2/3}$ (up to logarithmic factors), where $d$ denotes the data dimension, and $\varepsilon$ quantifies the output accuracy in terms of total variation (TV) distance.
In addition, $L$ represents a relaxed Lipschitz constant, which, in the case of Gaussian mixture models, scales only logarithmically with the number of components, the dimension and iteration number, demonstrating broad applicability.
\end{abstract}

\section{Introduction}

Score-based diffusion models have emerged as a powerful class of generative models capable of synthesizing high-quality
data from complex probability distributions \citep{Dhariwal2021Diffusion,Ho2020Denoising,SohlDickstein2015Deep,Song2019Generative,song2021denoising}.
These models, including Denoising Diffusion Probabilistic Models (DDPM) \citep{Ho2020Denoising} and Denoising Diffusion Implicit Models (DDIM) \citep{song2021denoising}, transforms pure noise into samples from the data distribution through an iterative denoising process.
This process is facilitated by a series of score functions, which approximate the gradient of the data log-density using pretrained neural networks.
In practice, diffusion models have achieved remarkable performance in various tasks
including image generation \citep{Ramesh2022Hierarchical,Rombach2021HighResolution,Saharia2022Photorealistic}, video generation \citep{villegas2022phenaki} and so on \citep{Croitoru2023Diffusion,yang2023diffusion,zhang2025improving}.
For an overview of recent developments, including both empirical and theoretical advancements, readers may refer to \citet{yang2023diffusion,tang2024score,Croitoru2023Diffusion}.

Diffusion models typically consist of two processes: the forward process and the reverse (or backward) process. The forward process is a simple stochastic process that progressively transforms a data sample $X_0$ into nearly pure noise $X_T$ by iteratively adding Gaussian noise:
\begin{align*}
    X_0 \overset{{\rm add~noise}}{\rightarrow} X_1 \overset{{\rm add~noise}}{\rightarrow} \cdots \overset{{\rm add~noise}}{\rightarrow} X_T.
\end{align*}
Here, $X_0$ is a $d$-dimensional sample from the target data distribution $p_{\mathsf{data}}$, and $X_T$ approximately follows a standard Gaussian distribution $\mathcal{N}(0,I_d)$. 
The core of diffusion models lies in the reverse process, which aims to learn a process that generates a sample resembling the target data distribution from pure Gaussian noise:
\begin{align*}
    Y_0 \overset{{\rm denoise}}{\rightarrow} Y_{1} \overset{{\rm denoise}}{\rightarrow} \cdots \overset{{\rm denoise}}{\rightarrow} Y_T,
\end{align*}
Here, $Y_0$ is initialized as white Gaussian noise, and the reverse process aims to achieve $Y_t\overset{d}{\approx} X_{T-t}$ for all $t$, such that the distribution of the final output $Y_T$ approximates the target distribution $p_{\mathsf{data}}$.
The most critical component of diffusion models is the efficient construction of this reverse process. To achieve this, diffusion models employ the time-reversal of stochastic differential equations (SDEs) to generate $Y_{t+1}$ from $Y_t$. This process relies on score functions ($s_{T-t}^\star = \nabla \log p_{X_{T-t}}$), which are gradients of the log marginal density of the forward process $X_{T-t}$. These score functions are typically pretrained using score-matching techniques \citep{Hyvarinen2005Estimation,Ho2020Denoising,Hyvrinen2007Some,Vincent2011AConnection,Song2019Generative,Pang2020Efficient}.

Due to the impressive empirical success of diffusion models, a lot of efforts in recent years have focused on analyzing their convergence properties.
Given the complexity of establishing a comprehensive end-to-end theoretical framework, most studies adopt a divide-and-conquer approach, treating the score matching step as a black box and focusing on the data generation process.
Following this framework, several studies have analyzed how factors such as data dimension, score estimation error, and the number of iterations affect the accuracy of approximating the target distribution \citep{Bortoli2022Convergence,Gao2023Wasserstein,Lee2022Convergence1,Lee2022Convergence2,Chen2022Sampling,Benton2023Nearly,Chen2023Improved,Li2023Towards,li2024sharp,Gupta2024Faster,Chen2023The,Li2024Accelerating,li2024adapting,li2024d,huang2024denoising,li2024provable,li2024improved,huang2024convergence,huang2024reverse,liang2025low}. 
For general data distributions, the best-known results are $\widetilde{O}(d\varepsilon^{-1})$ for DDPM \citep{li2024d} and $\widetilde{O}(d^{5/4}\varepsilon^{-1/2})$ for an accelerated sampler \citep{li2024provable}, where $\widetilde{O}(\cdot)$ omits logarithmic factors and $d$ denotes the data dimension.
Some works exploited smoothness conditions to further improve the iteration complexity, with the state-of-the-art achieved by \citet{li2024improved} is $\widetilde{O}(d^{1/3}L\varepsilon^{-2/3})$, where $L$ denotes the Lipschitz constant of score functions.
However, this improvement holds only when $L\lesssim \min\{d^{2/3}\varepsilon^{-1/3}, d^{11/12}\varepsilon^{1/6}\}$, which is restrictive.
This motivates us to develop an $L$-adaptive convergence bound for diffusion models, with the hope to achieve improvement over the full range of $L$.

\subsection{Our contributions}
\label{subsec:contribution}

In this paper, we investigate a sampler for SGMs based on the randomized midpoint technique,
and establish a convergence rate adaptive to the smoothness of score functions.
Specifically, up to logarithmic factors, we achieve the following iteration complexity:
%This paper focuses on convergence analysis adaptive to the smoothness $L$ of score functions, and develops an improved iteration complexity as following (up to some logarithmic factors):
\begin{align*}
\min\{d,d^{2/3}L^{1/3},d^{1/3}L\}\varepsilon^{-2/3},
\end{align*}
where $L$, as defined in Definition~\ref{def:score-lipschitz}, characterizes the smoothness of score functions.
Below, we provide a brief comparison of our results with existing convergence rates, viewed in Figure~\ref{fig:comp}:
\begin{itemize}
\item \textbf{Comparison under smoothness condition.} 
Under a uniform Lipschitz assumption, i.e., $\|s_t^{\star}(x) - s_t^{\star}(x')\|_2\le L\|x-x'\|_2$ for all $x,x'\in\mathbb{R}^d$ and all $t$, several works \citep{Chen2023Improved, Chen2023The,Gao2023Wasserstein, Gupta2024Faster, li2024improved, Lee2022Convergence1} have studied the convergence rate of SGMs. 
In comparison, our result has two advantages:
\begin{itemize}
\item The uniform Lipschitz condition used in previous works is significantly more restrictive than the non-uniform condition in Definition~\ref{def:score-lipschitz} adopted here.
Specifically, for Gaussian mixture models (GMMs), we show that $L$ scales only logarithmically with the number of components and dimension, whereas the uniform Lipschitz constant can be extremely large.
This highlights the superiority of our results (see Example~\ref{example:GMM} for details).
Achieving these advancements requires substantial technical innovations. 
%In particular, to control error propagation and bound the discretization error under the non-uniform Lipschitz conditions, we introduce two auxiliary sequences defined over a high-probability set. In addition, we refine the analysis of the derivative norm of the score functions through a careful decomposition and by leveraging the statistical structure of the derivative.
Detailed comparisons are in Section \ref{sec:mainresults} and Appendix \ref{app:comparison} of supplemental materials.

\item Among existing results, the best-known bound under smoothness conditions is $\widetilde{O}(d^{1/3}L\varepsilon^{-2/3})$, established by \citet{li2024improved}. 
Our convergence rate improves prior theory by a factor of $\max\{d^{-2/3}L, d^{-1/3}L^{2/3},1\}$, yielding substantial gains when $L\gtrsim \sqrt{d}$ even without considering the additional benefits from relaxing the Lipschitz condition.
\end{itemize}

\item \textbf{Comparison under general condition.} 
Without smoothness assumption on score functions,~\citet{Benton2023Nearly} established an iteration complexity at the order of $d\varepsilon^{-2}$, which marks the first result with linear dependency on the data dimension $d$.
This was later improved to $\tilde{O}(d\varepsilon^{-1})$ by~ \citet{li2024d} and \citet{li2024sharp}.
In comparison, our analysis further improves these bounds by a factor of $\max\{1,d^{1/3}L^{-1/3},d^{2/3}L^{-1}\}\varepsilon^{-1/3}$, which is significant for the entire range of $L$.
Notably, even in the case of $L=\infty$, our result achieves a significant improvement, except in the trivial scenario where $\varepsilon\asymp 1$.

\item \textbf{Comparison with accelerated samplers.}
Existing accelerated convergence theories often rely on additional assumptions about the target distribution or the estimation error of the higher-order score functions.
For example, results in \citet{huang2024convergence,huang2024reverse} depend on the bound of the first $p$-th derivatives of score functions or their estimates, and achieved a convergence rate faster than $\varepsilon^{-2/3}$ when $p\ge 3$.
In addition, \citet{Li2024Accelerating} assumed that the estimation errors of the jacobian matrix of score functions are bounded, and developed an improved rate of $\varepsilon^{-1/2}$.
More recently, \citet{li2024provable} achieved the state-of-the-art iteration complexity of $\widetilde{O}(d^{5/4}\varepsilon^{-1/2})$ without additional distribution or estimation assumptions.
In comparison, our result improves this bound when the number of iterations $T\lesssim \max\{d^2,d^3L^{-1},d^4L^{-3}\}$.
Notably, even in the case of $L=\infty$, a significant improvement is achieved as long as $T\lesssim d^2$.
\end{itemize}

%Finally, we use GMMs as an exmaple to illustrate the key contributions of this work.
%Given that GMMs typically do not satisfy the uniform Lipschitz continuity condition nor exhibit inherent low-dimensional structure, the best-known existing convergence rate is $\tilde{O}(d\varepsilon^{-1/2}\min\{\varepsilon^{-1/2},d^{1/4}\})$.
%In contrast, under our proposed framework, GMMs satisfy a non-uniform Lipschitz condition with $L=O(\log(H(T+d)))$, where $H$ denotes the number of components, yielding an iteration complexity of $\widetilde{O}(d^{1/3}\varepsilon^{-2/3})$.
%This achieves an improvement over prior results by a factor of $\min\{d^{2/3}\varepsilon^{-1/3},d^{11/12}\varepsilon^{1/6}\}$, which is substantially larger than $O(1)$ as long as $T\lesssim d^4$.

\noindent \textbf{Organizations:} The remaining of this paper is structured as follows.
Section \ref{sec:preliminary} provides a brief overview of fundamental concepts on SGMs, and presents the sampling algorithm.
In Section \ref{sec:mainresults}, we present our assumptions and main results, and make comparison with previous works. 
The theoretical analysis and proofs are detailed in Section \ref{sec:analysis}. Finally, Section \ref{sec:conclusion} concludes the paper and discusses potential directions for future research.

\section{Preliminary}
\label{sec:preliminary}

In this section, we present some basic concepts of generative diffusion models, introduce the sampler employed in this work, and clarify our goal.

\subsection{Score-based diffusion models}

A diffusion model typically involves two key processes: the forward process and the reverse process, which are explained below.

\noindent \textbf{Forward process.}
The forward process starts from a random instance $X_0\in\mathbb{R}^d$ sampled from the target data distribution $p_{\mathsf{data}}$, and progressively transforms it into pure Gaussian noise by iteratively adding noise at each step:
\begin{align}
X_t =\sqrt{\alpha_t}X_{t-1} +
\sqrt{1-\alpha_t} W_t, \quad 1 \le t \le T,
\end{align}
where $\{W_t\}_{1\le t\le T}$ is a sequence of independent Gaussian vectors drawn from $\mathcal{N}(0,I_d)$, and $\alpha_t\in(0,1)$ represents the step-size. 
For ease of notations, we define
\begin{align}
\overline{\alpha}_t:=\prod_{k=1}^t\alpha_k,\quad 1\le t\le T.
\end{align}
With this notation, $X_t$ can be expressed as a linear combination of the original data instance $X_0$ and Gaussian noise with variance $1-\overline{\alpha}_t$, as below.
\begin{align}\label{eq:def-Xt}
X_t = \sqrt{\overline{\alpha}_t}X_0 + \sqrt{1-\overline{\alpha}_t}~\overline{W}_t,\quad \overline{W}_t\sim\mathcal{N}(0,I_d).
\end{align}
As $\overline{\alpha}_t$ approaches zero, the distribution of  $X_t$ becomes exceedingly close to $\mathcal{N}(0,I_d)$.

Diffusion models are closely related to stochastic differential equations (SDEs). The continuous-time limit of the forward process can be modeled as:
\begin{align}\label{eq:forward-ODE}
\mathrm{d}X_{\tau} = -\frac{1}{2(1-\tau)}X_{\tau}\mathrm{d}\tau + \frac{1}{\sqrt{1-\tau}}\mathrm{d}B_{\tau},\quad\text{for }0\le \tau < 1,
\end{align}
where $B_{\tau}$ denotes some Brownian motion, $X_0\sim p_{\mathsf{data}}$, and the distribution of $X_\tau$ approaches Gaussian as $\tau$ gets close to $1$.

\noindent \textbf{Reverse process and score functions.}
The core of diffusion models lies in the reverse (or backward) process, which starts from pure Gaussian noise $Y_0\sim\mathcal{N}(0,I_d)$, and aims to generate samples $Y_T$ that resemble the target data distribution.
Many reverse processes are designed based on insights from the probability flow ODE, which is given by
\begin{align}\label{eq:ode-0}
\mathrm{d}Y_{\tau} = -\frac{1}{2(1-\tau)}\big(Y_{\tau} + \nabla \log p_{X_{\tau}}(Y)\big) \mathrm{d} \tau.
\end{align}
This ODE ensures that $Y_\tau$ follows the same distribution as $X_\tau$ defined in \eqref{eq:forward-ODE}, provided that the initial point $Y_{\tau_0}\sim p_{X_{\tau_0}}$, where $\tau_0$ is close to one and $Y_{\tau_0}$ is approximately Gaussian noise.

In ODE \eqref{eq:ode-0}, the only additional term except $Y_\tau$ itself is the gradient of the log-density of the forward process, $\nabla \log p_{X_\tau}$, known as the score function.
Its formal mathematical definition is provided below.

\begin{definition}\label{def-score} 
The score function $s^{\star}_t : \mathbb{R}^d \rightarrow \mathbb{R}^d$ for $1 \le t \le T$ is defined as:
\begin{align}\label{eq:def-score}
s_{t}^{\star}(x) &:= \nabla \log p_{X_t}(x) = -\frac{1}{1-\overline{\alpha}_t}\int_{x_0} p_{X_0 \mymid X_t}(x_0 \mymid x)(x - \sqrt{\overline{\alpha}_t}x_0)\mathrm{d}x_0. 
\end{align}
\end{definition}
For ease of notations, we also define $s_{\tau}^{\star}(x)$ with a continuous index $0< \tau < 1$ as follows:
\begin{align}\label{eq:def-stau}
s_{\tau}^{\star}(x) &:= \nabla \log p_{X_\tau}(x) = -\frac{1}{\tau}\int_{x_0} p_{X_0 \mymid X_\tau}(x_0 \mymid x)(x - \sqrt{1-\tau}x_0)\mathrm{d}x_0. 
\end{align}
It is evident that $s_{t}^{\star}(\cdot) = s_{1-\overline{\alpha}_t}^{\star}(\cdot)$.
In practice, the true score function $s_t^{\star}$ is typically unknown and must be estimated from a training dataset.
We assume access to faithful estimates $s_t$ of the score functions $s_{t}^{\star}$ across all steps $t$.
The assumption regarding score estimation errors is formally presented in Assumption \ref{assu:score-error}.

\subsection{Sampling algorithm}
\label{subsec:Algorithm}

The sampler used is the same as the one employed by \citet{li2024improved}, which is derived from the discretization of a probability flow ODE for $X_\tau$ over $\tau\in(0,1)$.
Specifically, we first discretize $\tau$ into intermediate points $\tau_{k,n}$ within the interval $(0,1)$, where $n=0,\cdots,N$ and $k=0,\cdots,K$. 
We then estimate $X_{\tau_{k,n}}$ at these intermediate points by approximating the integral of probability flow ODE.
The detailed implementation is stated below.

\noindent \textbf{Randomized schedule.}
We begin by discretizing the interval $[0,1]$ into a sequence of subintervals $(\widehat{\alpha}_t,\widehat{\alpha}_{t-1})$, where $\widehat{\alpha}_t$ is defined as
\begin{align}
\widehat{\alpha}_{T+1} = \frac{1}{T^{c_0}},
\quad
\widehat{\alpha}_{t-1} = \widehat{\alpha}_{t} + \frac{c_1\widehat{\alpha}_{t}(1-\widehat{\alpha}_{t})\log T}{T}, \qquad t=-\frac{N}{2}+1,\cdots,T+1,
\end{align}
for some sufficiently large constants $c_0, c_1 > 0$, where the ratio $c_1/c_0$ is assumed to be sufficiently large.
Subsequently, we employ a randomized learning rate schedule by setting $\overline{\alpha}_t$ in \eqref{eq:def-Xt} as
\begin{subequations}
\label{eq:learning-rate}
\begin{align}
\overline{\alpha}_{t} \sim \mathsf{Unif}(\widehat{\alpha}_{t}, \widehat{\alpha}_{t-1}),
\quad\text{for }t = -\frac{N}{2}+1, \ldots, T+1,
\end{align}
where $\mathsf{Unif}$ denotes the uniform distribution.

The algorithm operates over $K$ rounds, each consisting of  $N=\frac{2T}{K}$ steps. 
We define
%Then for $k = 0, \ldots, K-1$, we define
\begin{align}
\widehat{\tau}_{k, n} &:= 1-\widehat{\alpha}_{T-\frac{kN}{2}-n},\qquad \tau_{k, n} := 1-\overline{\alpha}_{T-\frac{kN}{2}-n+1}\qquad \text{for}\quad n = -1, \ldots, N.
%\\
%\tau_{k, n} &:= 1-\overline{\alpha}_{T-\frac{kN}{2}-n+1},\qquad \text{for}\quad n = 0, \ldots, N.
\end{align}
\end{subequations}
It follows that
%\begin{align*}
$
\tau_{k,n}\sim\mathsf{Unif}(\widehat{\tau}_{k,n},\widehat{\tau}_{k,n-1}),
$
%\end{align*}
which we use as the discretization of $\tau$.

\noindent \textbf{Sampling procedure.}
With $\tau$ discretized, we are now ready to describe the sampling procedure. As aforementioned, the sampler is implemented over $K$ rounds, each consisting of $N$ steps.
In the $k$-th round, the sampler approximates $X_{\tau_{k,N}}$ defined by the probability flow ODE with the initial point given by $X_{\tau_{k,0}}$, whose probability distribution is denoted as $q_k$.
At the end of each round, a Gaussian noise is injected to convert the total variation distance between the reverse and forward processes into an estimation error in the $\ell_2$ norm.
The sampling procedure consists of the following steps:
\begin{enumerate}
\item Initialization: 
The sampler begins with an initial sample $Y_{0} \sim \mathcal{N}(0, I_d)$.
\item Iterative update:
For each $k$ ranging from $0$ to $K-1$, the intermediate variables $Y_{k,n}$ are iteratively updated for $n=1,\cdots,N$ by discretizing the ODE as follows:
\begin{subequations}
\label{eq:sampler}
\begin{small}
\begin{align}
&\frac{Y_{k, n}}{\sqrt{1-\tau_{k, n}}} = \frac{Y_{k,0}}{\sqrt{1-\tau_{k,0}}} 
+ \frac{s_{T-\frac{kN}{2}+1}(Y_{k,0})}{2(1-\tau_{k,0})^{3/2}}(\tau_{k, 0} - \widehat{\tau}_{k,0})\notag\\
&+ \sum_{i = 1}^{n-1} \frac{s_{T-\frac{kN}{2}-i+1}(Y_{k,i})}{2(1-\tau_{k,i})^{3/2}}(\widehat{\tau}_{k,i-1} - \widehat{\tau}_{k,i}) 
+ \frac{s_{T-\frac{kN}{2}-n+2}(Y_{k,n-1})}{2(1-\tau_{k,n-1})^{3/2}}(\widehat{\tau}_{k,n-1} - \tau_{k, n}),
\end{align}
\end{small}
where $Y_{k,0}=Y_k$ and $s_{T-\frac{kN}{2}-i+1}$ is an estimation of $s_{T-\frac{kN}{2}-i+1}^{\star}$ as defined in \eqref{eq:def-score}, corresponding to $\overline{\alpha}_{T-\frac{kN}{2}-i+1} = 1-\tau_{k,i}$.
\item Noise injection:
After obtaining $Y_{k, N}$, we update $Y_{k+1}$ by injecting stochastic noise as:
\begin{align}
Y_{k+1} &= \sqrt{\frac{1-\tau_{k+1, 0}}{1-\tau_{k, N}}}Y_{k, N} + \sqrt{\frac{\tau_{k+1, 0}-\tau_{k, N}}{1-\tau_{k, N}}}Z_k,
\end{align}
where $Z_k \overset{\mathrm{i.i.d.}}{\sim} \mathcal{N}(0, I_d)$. 
\end{subequations}
\end{enumerate}
Notice that at each step of computing $Y_{k,n}$, only one additional score evaluation for $s_{T-\frac{kN}{2}-n+2}(Y_{k,n-1})$ is required.
Therefore, the total iteration complexity of the sampler is $KN = 2T$.
This sampler can be implemented in parallel, as demonstrated by \citet{li2024improved}. For clarity, we present the details in Appendix \ref{subsec:parallel-alg} of the supplemental material.

\noindent \textbf{Our goal.} The objective of this work is to improve the convergence rate for a broader class of distributions by analyzing the aforementioned sampler.
Since $X_{\tau_{K,0}}$, which is nearly the starting point of the forward process, is perturbed only by noise with a small variance of $1-\overline{\alpha}_1$, 
the performance of the sampler is evaluated using the total variation (TV) distance between $p_{Y_K}$ and $q_{K}$, 
defined as 
\begin{align*}
\mathsf{TV}(q_{K},p_{Y_K}) := %\sup_{{A}\in\mathcal{F}}|P(A) - Q(A)| = 
\frac12 \int|p_{Y_K}(x)-q_K(x)| \mathrm d x.
\end{align*}
%For ease of notations, we denote the probability distribution of $X_{\tau_{k,0}}$ as $q_k$.

\section{Main results}
\label{sec:mainresults}

In this section, we establish an instance-dependent convergence rate for diffusion models under a novel analytical framework that accommodates a relaxed Lipschitz condition, thereby covering a broader class of distributions.
Finally, we extend our results to the parallel implementation of the sampler, following an idea similar to that of \citet{li2024improved}.

\subsection{Assumptions}

We first make the following assumption on the target data distribution $p_{\mathsf{data}}$, which accommodates a broad class of data distributions.
\begin{assumption}\label{assu:distribution}
We assume that the target distribution $p_{\mathsf{data}}$ has a bounded second-order moment in the sense that
\begin{align}
\mathbb{E}_{X_0\sim p_{\mathsf{data}}}[\|X_0\|_2^2] < T^{c_R},
\end{align}
where $c_R > 0$ is an arbitrarily large constant.
\end{assumption}
The second-order moment of $X_0$ is assumed to be polynomial in the number of iterations $T$.
This assumption is mild as the exponent $c_R$ can be arbitrarily large. 
However, there exist exceptions including extremely heavy-tailed distributions. For example, densities decaying slower than $1/x^3$ may not align with our theoretical framework.

Our analysis is conducted under a relaxed smoothness condition, which is substantially weaker than the commonly used uniform Lipschitz condition and encompasses a wide range of distributions previously considered non-Lipschitz. Specifically, we define the Lipschitz constant for the normalized score functions $(1-\overline{\alpha}_t)s_t^{\star}$ as follows.
\begin{definition}[Non-uniform Lipschitz property]\label{def:score-lipschitz}
Let $L$ denote the smallest quantity, which may depend on $T$ and $d$, such that 
\begin{align*}
&\quad\mathbb{P}_{x\sim X_t}\left\{(1-\overline{\alpha}_t)\|s_{t}^{\star}(x') - s_{t}^{\star}(x)\|_2 \le L\|x' - x\|_2,~\forall~\|x'-x\|_2\le \frac{C\sqrt{d(1-\overline{\alpha}_t)\log T}}{L}\right\}\notag\\
&\ge 1-\frac{c}{(T+d)^4},
%L:=\sup_{t\ge 0}\sup_{x_1,x_2}\frac{\|s_{t}^{\star}(x_1) - s_{t}^{\star}(x_2)\|_2}{\|x_1 - x_2\|_2}.
\end{align*}
where $C$ and $c$ are some universal constants.
\end{definition}

\begin{remark}\label{rem:Lipschitz}
Previous works typically assume a uniform Lipschitz condition, requiring $(1-\overline{\alpha}_t)\|s_t^{\star}(x) - s_t^{\star}(x')\|_2\le L\|x-x'\|_2$ (or $\|s_t^{\star}(x) - s_t^{\star}(x')\|_2\le L\|x-x'\|_2$) to hold for all $x$ and $x'$.
In contrast, this work adopts a non-uniform Lipschitz condition, which is a significant relaxation compared to prior assumptions.
%In addition, this work allows the Lipschitz constant of the score function $s_t^{\star}$ to vary with the noise variance $1-\overline{\alpha}_t$, following the similar idea in \citet{Lee2022Convergence2,Bortoli2022Convergence}.
%In practice, the non-uniform and local Lipschitz constant $L$ may depend on $\log T$ and $\log d$.
In what follows, we present two examples to demonstrate the practical importance and necessity of this relaxation.
\end{remark}

\begin{example}[Gaussian distribution]
\label{example:Gaussian}
Consider Gaussian target distribution 
$X_0^{(i)}\sim \mathcal{N}(0,\sigma_i^2)$, where $X_0^{(i)}$ denotes the $i$-th entry of $X_0$, $i=1,\cdots,d$.
The score function $s_t^{\star}(x)= \nabla \log p_{X_t}(x)$ satisfies
\begin{align*}
\forall~t > 0,~\forall~x,x'\in\mathbb{R}^d,\qquad &(1-\overline{\alpha}_t)\|s_t^{\star}(x) - s_t^{\star}(x')\|_2\le \|x-x'\|_2,\\%\label{eq:example-Gaussian-1}\\
\forall~t\ge 0,~~ \exists~x,x'\in\mathbb{R}^d, \quad {\rm such~that}\qquad  &\|s_t^{\star}(x) - s_t^{\star}(x')\|_2\ge (1-\overline{\alpha}_t)^{-1}\|x-x'\|_2,%\label{eq:example-Gaussian-2}
\end{align*}
provided that $\min_i\sigma_i^2 = 0$.
This highlights the rationale for defining the Lipschitz condition for $(1-\overline{\alpha}_t)s_t^{\star}$.
The detailed derivations are presented in Appendix \ref{app:examples-Gaussian}.
\end{example}

\begin{example}[Gaussian Mixture Models]
\label{example:GMM}
Consider GMM target distribution 
$X_0\sim \sum_{h=1}^H\gamma_h\mathcal{N}(\mu_h,\sigma^2I_d)$, where $\mu_h\in\mathbb{R}^d$, $\sigma\ge 0$, $\gamma_h\ge 0$ and $\sum_{h=1}^H\gamma_h = 1$.
Then for any $t\ge 0$, we have
\begin{small}
\begin{align*}
&\mathbb{P}\left\{(1-\overline{\alpha}_t) \|s_t^{\star}(x) - s_t^{\star}(x')\|_2\le C_1\log(H(T+d))\|x-x'\|_2,~\forall~\|x-x'\|_2
\le C_2\sqrt{d(1-\overline{\alpha}_t)}\right\} \notag\\
&\ge 1- \frac{c}{(T+d)^4},
%^\label{eq:example-GMM-1}
\end{align*}
\end{small}
for some universal constants $C_1, C_2$ and $c$.
Moreover, consider a simple case that $X_0\sim \frac{1}{2}\mathcal{N}(\mu,\sigma^2I_d) + \frac{1}{2}\mathcal{N}(-\mu,\sigma^2I_d)$, then there exists some $x\in\mathbb{R}^d$, such that for $\overline{\alpha}_t>1/2$,
\begin{align*}
(1-\overline{\alpha}_t)\|\nabla s_t^{\star}(x)\|_{\mathsf{op}} \ge \frac{(1-\overline{\alpha}_t)\|\mu\|_2^2}{4(1-\overline{\alpha}_t+\sigma^2)^2}.
%\label{eq:example-GMM-2}
\end{align*}
The above example implies that the non-uniform Lipschitz constant remains relatively small, satisfying $L\le \log(H(T+d))$. In contrast, the uniform Lipschitz constant may be extremely large when $\sigma^2$ is small enough. Here, $\|\mu\|_2^2 \approx \mathbb{E}[\|X_0\|_2^2]$ is typically large in practice, which often scales on the order of $d$. 
% This implies that $L\le \log(H(T+d))$ holds for Definition \ref{def:score-lipschitz}.
% In contrast, the uniform Lipschitz constant can be extremely large, when the norm $\|\mu\|_2^2$ which corresponds to $\mathbb{E}\|X_0\|_2^2$ is large, or when $\sigma^2\lesssim 1-\overline{\alpha}_t$ and $1-\overline{\alpha}_t + \sigma^2$ is small enough.
%It is worth noting that the Bernoulli distribution is a special case of this example with $H=2$ and $\sigma=0$.
The detailed derivations are presented in Appendix \ref{app:examples-GMM}.
\end{example}

Finally, we make the following assumption about the estimation error of score functions.
%For ease of notations, we denote the probability distribution of $X_{\tau_{k,0}}$ as $q_k$.
\begin{assumption}\label{assu:score-error}
We assume access to an estimate $s_t(\cdot)$ for each $s^{\star}_t(\cdot)$, with the averaged $\ell_2$ score estimation error as
\begin{small}
\begin{align} \label{eq:score-error}
%\varepsilon_{\mathsf{score}}^2 = \frac{1}{T}\sum_{t = 1}^T \mathbb{E}_{x \sim p_{Y_t}}\big[\|s_{t}(x) - s_{t}^{\star}(x)\|_2^2\big]=: \frac{1}{T}\sum_{t = 1}^T \varepsilon_t^2.
\varepsilon_{\mathsf{score}}^2 &= \frac{1}{T}\sum_{k = 0}^{K-1} \sum_{n=0}^{N-1}\mathbb{E}_{Y_{k}\sim q_{k}}\big[\|s_{T-\frac{kN}{2}-n+1}(Y_{k,n}) - s_{T-\frac{kN}{2}-n+1}^{\star}(Y_{k,n})\|_2^2\big]\notag\\
&=: \frac{1}{T}\sum_{k = 0}^{K-1} \sum_{n=0}^{N-1} \varepsilon_{k,n}^2.
\end{align}
\end{small}
\end{assumption}

\subsection{Convergence Analysis}

We are now positioned to present the convergence guarantees --- measured by the total variation distance between the forward and the reverse processes --- for the sampler \eqref{eq:sampler}.
The proof is postponed to Section \ref{sec:analysis}.

\begin{theorem} \label{thm:main}
Suppose that Assumptions \ref{assu:distribution} and \ref{assu:score-error} hold true, and $K = c_2\min\{d\log^2 T, L\log T\}$ for some constant $c_2 > 0$.  
Then the sampling process \eqref{eq:sampler} with the learning rate schedule \eqref{eq:learning-rate} satisfies
\begin{align}\label{eq:TV-main}
\mathsf{TV}\left( q_{K},p_{Y_{K}}\right) \le \frac{C\min\{d^{3/2}, dL^{1/2},d^{1/2}L^{3/2}\}\log^{4} T}{T^{3/2}} + C\varepsilon_{\mathsf{score}}\log^{1/2} T
\end{align}
for some constant $C > 0$ large enough, where $L$ is defined in Definition \ref{def:score-lipschitz}.
\end{theorem}

We now discuss the main implications of Theorem~\ref{thm:main}.

\noindent \textbf{Relaxation of smoothness.}
Our result requires only a non-uniform Lipschitz condition, which is substantially weaker than the uniform Lipschitz condition commonly employed in prior studies. 
As demonstrated in Example~\ref{example:GMM} for Gaussian mixture models, we have shown that $L$ scales only logarithmically with the number of components, dimension and iteration number, whereas the uniform Lipschitz constant may be extremely large. Given the wide use of GMMs, this demonstrates that the uniform Lipschitz condition employed in previous works is more restrictive compared to the non-uniform condition in Definition~\ref{def:score-lipschitz} used here.

Achieving these improvements requires a lot of technical efforts. For example, the absence of a uniform Lipschitz condition poses significant challenges for controlling error propagation across multiple steps, while a naive stepwise analysis may lead to suboptimal bounds. We address this issue by introducing two auxiliary sequences based on a typical set and quantifying how error propagation affects the probability of $Y_{k,n}$ outside this set (see Step 2 in Section \ref{sec:analysis} and Lemma \ref{lem:endpoint}). In addition, lacking a uniform condition prevents the log-concavity of $p_{X_{\tau}|X_{\tau+\delta}}$ for small $\delta$, which plays a crucial rule in controlling the one-step discretization error (see~Lemma 1 and (B.2)-(B.4) in \citet{Chen2023The}). We instead directly analyze this derivative based on its definition via a careful decomposition and statistical bounds (see Lemma \ref{lem:bound-der-tau}). 
Further details are provided in Appendix~\ref{app:comparison} of supplemental material.

\noindent \textbf{Iteration complexity.}
%For the moment, we disregard the score estimation error by assuming access to the perfect score function estimation, i.e., $\varepsilon_{\mathsf{score}}=0$.
For the moment, we focus on the first term in \eqref{eq:TV-main}, which corresponds to discretization error.
%, and ignore the score estimation error.
To ensure $\mathsf{TV}\left(q_K,p_{Y_{K}}\right) \le \varepsilon$, it is sufficient to choose
\begin{align*}
T \gtrsim \frac{\min\{d,d^{2/3}L^{1/3},d^{1/3}L\}\log^{\frac83} T}{\varepsilon^{2/3}}.
\end{align*}
This result is adaptive to the non-uniform Lipschitz constant $L$ of normalized score functions and is thus referred to as \emph{instance-dependent}.
Different from previous works, which only improve iteration complexity under specific conditions on $L$, our theory improves existing results \citep{li2024provable,li2024d,Benton2023Nearly} over a full range of $L$,
and improves the state-of-the-art \citep{li2024improved} when $L\gtrsim \sqrt{d}$, even when non-uniform and uniform Lipschitz constants are equal.
A detailed comparison of iteration complexity orders has been also provided in Section \ref{subsec:contribution}.

\begin{figure*}%
\centering
{\includegraphics[width=0.9\textwidth]{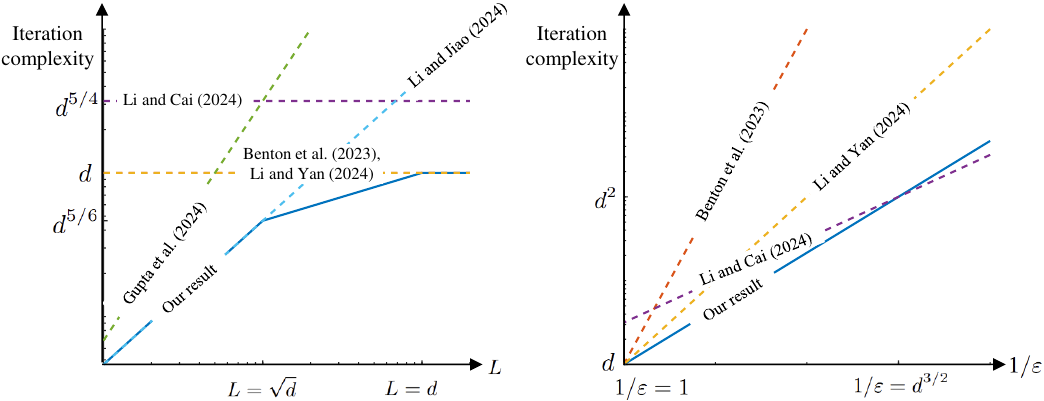}}
\caption{Comparison of Theorem \ref{thm:main} with prior results.
left: the iteration complexity as a function of $L$ when $\varepsilon = O(1)$.
right: the iteration complexity as a function of $\varepsilon$ when $L=\infty$.}\label{fig:comp}
\end{figure*}

To illustrate these improvements, we assume the non-uniform and uniform Lipschitz constants are identical and compare iteration complexities for $\varepsilon = O(1)$ across varying values of $L$, as shown in the left subplot of Figure \ref{fig:comp}.
It indicates that prior works \citep{Gupta2024Faster,Benton2023Nearly,li2024provable,li2024d} outperform others in specific regimes. 
In contrast, our instance-dependent result achieves the best result across the full range of $L$, and improves all previous works when $\sqrt{d}\lesssim L\lesssim  d$.
Furthermore, to more clearly illustrate the improvement in the case of $L=\infty$, we present the iteration complexity as a function of $\varepsilon$ in the right subplot of Figure \ref{fig:comp}.
It demonstrates clear improvement over previous results when $T\lesssim d^{2}$.
We verify our theoretical result via a numerical simulation, which is provided in Appendix \ref{app:simulations} of supplemental material.
%Finally, we remark that our result holds for a normalized Lipschitz constant $L$ (c.f. Definition \ref{def:score-lipschitz}), which represents a milder assumption compared to those in \citet{li2024improved,Gupta2024Faster}.
More comparisons are provided in Appendix \ref{app:comparison} of supplemental material.

\begin{remark}
Some studies have explored provably accelerated samplers for diffusion models \citep{huang2024convergence,huang2024reverse}, achieving convergence rate faster than the $\varepsilon^{-2/3}$ rate established in this work.
However, these results typically rely on the bound of the first $p$-th derivatives of the score functions or their estimates with $p\ge 3$.
\end{remark}

% \begin{figure}
% \begin{center}
% \includegraphics[width=0.9\textwidth]{figures/fig_comp_4.pdf}
% \end{center}
% \caption{Comparison of Theorem \ref{thm:main} with prior results.
% left: the iteration complexity as a function of $L$ when $\varepsilon = O(1)$.
% right: the iteration complexity as a function of $\varepsilon$ when $L=\infty$. \label{fig:comp}}
% \end{figure}

%\noindent \textbf{Dependency of score estimation error.}
%Turning attention to the second term, the sampling accuracy scales as $\varepsilon_{\mathsf{score}}\log^{1/2} T$, suggesting that our sampler is stable to imperfect score estimation.
%In comparison, the existing results replying on smoothness constant scale as $L^{1/2}d^{1/12}\varepsilon_{\mathsf{score}}\log T$ and $L^{1/2}\varepsilon_{\mathsf{score}}\log T$ in \citet{Gupta2024Faster} and \citet{Chen2023The}, respectively.
%Our theory exhibits a better dependency on score estimation error, which is similar to the results in~\citet{Chen2022Sampling,Benton2023Nearly,li2024d,li2024improved}.

\subsection{Extension to parallel sampling}

In this section, we present a theoretical guarantee for the parallel implementation described in Appendix~\ref{subsec:parallel-alg} of the supplementary material, following an approach similar to that of \citet{li2024improved,Gupta2024Faster}. Let $N$ denote the number of parallel processors and $MK$ the total number of parallel rounds. The convergence result is stated below, with a detailed proof provided in Appendix~\ref{sec:proof-thm-parallel} of the supplementary material.
\begin{theorem}
\label{thm:parallel}
Under the same assumptions as Theorem~\ref{thm:main}, it is sufficient to choose
\begin{align}
N &\gtrsim \frac{(\min\{d^{2/3}L^{-2/3},d^{1/3}\}+1)\log^{5/3} T}{\varepsilon^{2/3}},\notag\\
MK &\gtrsim \min\{d\log T,L\}\log^2 T,
\quad
\varepsilon_{\mathsf{score}}^2 \lesssim \varepsilon^2\log^{-1} T
\end{align}
to achieve $\mathsf{TV}\left(q_K,p_{Y_{K}}\right) \lesssim \varepsilon$ for the parallel sampler,
%~\eqref{eq:sampler-parallel},
where $T = KN/2$.
\end{theorem}

Finally, let us briefly compare our theory with the prior works.
This theorem states that the parallel sampler achieves $\varepsilon$-accuracy with respect to total variation distance using $O(\min\{L,d\}\log^2(Ld/\varepsilon))$ parallel rounds, which is consistent with the results in \citet{Gupta2024Faster,chen2024accelerating,li2024improved}.
Moreover, our sampler requires only $\widetilde{O}((\min\{d^{2/3}L^{-2/3},d^{1/3}\}+1)\varepsilon^{-2/3})$ parallel processors, which achieves a significant improvement over previous results.

\section{Analysis}
\label{sec:analysis}

This section is devoted to establishing Theorem~\ref{thm:main}.
%Before diving into the details of the proof, we provide an overview of the approach.
%First, we introduce auxiliary sequences based on high-probability typical sets, thereby focusing attention on samples within these sets.
%Second, by injecting noise into the probability flow ODE (c.f.\eqref{eq:ODE}), we transfer the sampling task to one task of discretizing the ODE  with small error (see Lemma \ref{lem:ODE}).
%This step contributes to the main improvement of our work over prior works, which provides a simple framework for transferring the sampling task to discretizing the ODE \eqref{eq:ODE} with a small error. Once this reduction is established, we leverage the randomized midpoint method, as in prior works, to achieve a more efficient discretization of the ODE and attain the desired bounds.
% Finally, we combine these results to obtain the final bound.
% The subsequent subsections will show each of these four steps in detail.
%
Before proceeding, 
we rewrite the sampling process with the continuous index as following:
\begin{align*}
\frac{Y_{\tau_{k, n}}}{\sqrt{1-\tau_{k, n}}} &= \frac{Y_{\tau_{k,0}}}{\sqrt{1-\tau_{k,0}}} 
+ \frac{s_{\tau_{k,0}}(Y_{\tau_{k,0}})}{2(1-\tau_{k,0})^{3/2}}(\tau_{k, 0} - \widehat{\tau}_{k,0})\\
&\quad
+ \sum_{i = 1}^{n-1} \frac{s_{\tau_{k,i}}(Y_{\tau_{k,i}})}{2(1-\tau_{k,i})^{3/2}}(\widehat{\tau}_{k,i-1} - \widehat{\tau}_{k,i}) + \frac{s_{\tau_{k,n-1}}(Y_{\tau_{k,n-1}})}{2(1-\tau_{k,n-1})^{3/2}}(\widehat{\tau}_{k,n-1} - \tau_{k, n}),
\end{align*}
and
\begin{align*}
Y_{\tau_{k+1,0}} &= \sqrt{\frac{1-\tau_{k+1, 0}}{1-\tau_{k, N}}}Y_{\tau_{k, N}} + \sqrt{\frac{\tau_{k+1, 0}-\tau_{k, N}}{1-\tau_{k, N}}}Z_k,
\end{align*}
where $Y_{\tau_{0,0}} \sim \mathcal{N}(0, I_d)$, $Z_k \overset{\mathrm{i.i.d.}}{\sim} \mathcal{N}(0, I_d)$.
Here, $Y_{\tau_{k,0}}$ corresponds to the $Y_{k}$ in discrete index,
and $s_\tau(\cdot)$ denotes the estimate of score function $s_\tau^{\star}(\cdot)$ defined in \eqref{eq:def-stau}.
For ease of notations, we denote estimation error at the $(k,n)$-th step as 
$$
\widetilde{\varepsilon}_{k,n} = \left\|s_{\tau_{k,n}}(Y_{\tau_{k,n}})-s_{\tau_{k,n}}^{\star}(Y_{\tau_{k,n}})\right\|_2.
$$
Moreover, we shall show that $Y_{\tau_{k,n}}$ follows a distribution similar to that of $X_{\tau_{k,n}}$ defined in \eqref{eq:forward-ODE}, which can also be expressed as
% Without abuse of notations, we use $X_{\tau}$ to denote the continuous process as follows:
% \eqref{eq:forward-ODE}
\begin{align}\label{eq:def-Xtau}
X_{\tau} \overset{\mathrm{d}}{=} \sqrt{1-\tau}X_0 + \sqrt{\tau}Z,\text{ with }Z\sim \mathcal{N}(0, I_d),\text{ for }0\le\tau\le1.
\end{align}

\subsection{Step 1: introduce the auxiliary sequence}

We first introduce an auxiliary sequence $\widehat{X}_k$, $1\le k\le K$, which has identical distribution with $X_{\tau_{k,0}}$.
Let $\Phi_{\tau_1 \to \tau_2}(x) := x_{\tau_2}\mymid_{x_{\tau_1} = x}$ defined through the following ODE
\begin{align} \label{eq:ODE}
\mathrm{d} \frac{x_{\tau}}{\sqrt{1-\tau}} = -\frac{s_{\tau}^{\star}(x_{\tau})}{2(1-\tau)^{3/2}} \mathrm{d} \tau.
\end{align}
Then, we have the following result, which has been presented in \citet{li2024improved}.
For completeness, we present its proof in Appendix~\ref{sec:proof-lem:ODE} in supplemental material:

\begin{lemma}\label{lem:ODE}
It can be shown that
\begin{align}\label{eq:lem-ODE}
\Phi_{\tau_1 \to \tau_2}(X_{\tau_1}) \overset{\mathrm{d}}{=} X_{\tau_2}.
\end{align}
Moreover, assume that $\widehat{X}_0 \overset{\mathrm{d}}{=} X_{\tau_{0, 0}}$.
Then we have for $0 \le k < K$,
\begin{align}\label{eq:lem-distribution}
\widehat{X}_{k+1} = \sqrt{\frac{1-\tau_{k+1, 0}}{1-\tau_{k, N}}}\Phi_{\tau_{k, 0} \to \tau_{k, N}}(\widehat{X}_k) + \sqrt{\frac{\tau_{k+1, 0}-\tau_{k, N}}{1-\tau_{k, N}}}Z_k \overset{\mathrm{d}}{=} X_{\tau_{k+1, 0}},
\end{align}
where $Z_k \overset{\mathrm{i.i.d.}}{\sim} \mathcal{N}(0, I_d)$.
\end{lemma}

%\vspace{-20ex}

Before introducing other auxiliary sequences, we introduce two notations for ease of presentation. Let
\begin{align}
\frac{y_{\tau_{k, n}}(x_{\tau_{k,0}})}{\sqrt{1-\tau_{k, n}}} &= \frac{x_{\tau_{k, 0}}}{\sqrt{1-\tau_{k,0}}} 
+ \frac{s_{\tau_{k,0}}(x_{\tau_{k, 0}})}{2(1-\tau_{k,0})^{3/2}}(\tau_{k, 0} - \widehat{\tau}_{k,0}) \notag\\
&\quad
+ \sum_{i = 1}^{n-1} \frac{s_{\tau_{k,i}}(y_{\tau_{k,i}})}{2(1-\tau_{k,i})^{3/2}}(\widehat{\tau}_{k,i-1} - \widehat{\tau}_{k,i})+ \frac{s_{\tau_{k,n-1}}(y_{\tau_{k,n-1}})}{2(1-\tau_{k,n-1})^{3/2}}(\widehat{\tau}_{k,n-1} - \tau_{k, n}),\nonumber\\
\frac{x_{\tau_{k, n}}(x_{\tau_{k,0}})}{\sqrt{1-\tau_{k, n}}} 
&= \frac{x_{\tau_{k, 0}}}{\sqrt{1-\tau_{k, 0}}} + \int_{\tau_{k,n}}^{\tau_{k,0}} \frac{s_\tau^{\star}(x_\tau)}{2(1-\tau)^{3/2}} \mathrm d \tau,\label{eq:def-xtau}
\end{align}
for $n = 1, \ldots, N$, which satisfies $x_{\tau_{k,n}}(x_{\tau_{k,0}}) = \Phi_{\tau_{k,0}\to\tau_{k,n}}(x_{\tau_{k,0}})$.
Moreover, we define typical sets $\mathcal{E}_k$ for $0\le k\le K-1$ as
\begin{small}
\begin{align*}
\mathcal{E}_k: = \left\{
\begin{array}{ll}
   \{x_{\tau_{k,0}}: x_{\tau_{k,n}}(x_{\tau_{k,0}})\in\widetilde{\mathcal{S}}_{\tau_{k,n}}\cap \mathcal{L}_{\tau_{k,n}}, &\\
   \qquad \quad y_{\tau_{k,n}}(x_{\tau_{k,0}})\in\mathcal{S}_{\tau_{k,n}}, \forall 0\le n \le N-1\},  & {\rm if}~L> d\log T, \\
   \emptyset  &  {\rm if}~L\le d\log T,
\end{array}
\right.
\end{align*}
\end{small}
where $\widetilde{\mathcal{S}}_{\tau}$, $\mathcal{S}_{\tau}$ and $\mathcal{L}_{\tau}$ denote high probability sets
\begin{small}
\begin{align}\label{eq:typical-set}
\mathcal{S}_{{\tau}}
&: = \{x: -\log p_{X_{\tau}}(x)\le \theta d\log T\},\quad
\widetilde{\mathcal{S}}_{{\tau}}
: = \{x: -\log p_{X_{\tau}}(x)\le \theta d\log T -\log 2\},\\
\mathcal{L}_{\tau} &:= \left\{x:\tau\|s_{\tau}^{\star}(x') - s_{\tau}^{\star}(x)\|_2 \le L\|x' - x\|_2, \forall \|x'-x\|_2\le \frac{C\sqrt{d\tau\log T}}{L}\right\},
\end{align}
\end{small}
with $\theta$ a sufficiently large constant.
Moreover, we define $\mathcal{E}_K = \mathbb{R}^d$.

Based on the above definitions, we introduce a new auxiliary reverse process $\widetilde{X}_k$ for $0\le k\le K$, which transforms similar with ODE in Lemma \ref{lem:ODE}, but removes samples out of the typical set $\mathcal{E}_k$.
Specifically, let $\widetilde{X}_0 = \widehat{X}_{0}$ for $\widehat{X}_{0}\in\mathcal{E}_{0}$ and $\widetilde{X}_0=\infty$ otherwise.
Then it transits following the probability
\begin{align}\label{eq:def-Xtilde}
p_{\widetilde{X}_{k+1}\mymid\widetilde{X}_{k}}(x|x_k) = \left\{
\begin{array}{ll}
p_{\widehat{X}_{k+1}\mymid \widehat{X}_{k}}(x\mymid x_k) \mathds{1}(x\in\mathcal{E}_{k+1}) & \\
\quad + \int_{\mathcal{E}_{k+1}^{\rm c}} p_{\widehat{X}_{k+1}\mymid \widehat{X}_{k}}(x_{k+1}\mymid x_k) \mathrm d x_{k+1} \delta_{\infty},& x_k\neq \infty,\\
\delta_{\infty},&x_k = \infty,
\end{array}
\right.
\end{align}
where according to Lemma \ref{lem:ODE},
$
p_{\widehat{X}_{k+1}\mymid \widehat{X}_{k}}(x\mymid x_k) = \phi\left(x|x_{\tau_{k,N}}(x_{k}), \sigma_k^2\right),
$
and $\phi(x|\mu,\sigma^2)$ denotes the probability density function of Gaussian distribution with mean vector $\mu$ and covariance matrix $\sigma^2I_d$, $\mathcal{E}_k^{\rm c}$ denotes the complementary set of $\mathcal{E}_k$ ($\infty\notin\mathcal{E}_{k}^{\rm c}$), and
\begin{align}\label{eq:def-sigma-k}
\sigma_k^2 = \frac{\tau_{k+1,0}-\tau_{k,N}}{1-\tau_{k,N}}.
\end{align}

Moreover, we define another reverse sequence $\widetilde{Y}_k$ similar with $Y_k$ as follows.
First, $\widetilde{Y}_0$ is initialized as ${Y}_0$ for $Y_0\in\mathcal{E}_{0}$ and $\widetilde{Y}_0 = \infty$ otherwise.
Then for $k=0,\cdots,K-1$, the conditional density of $\widetilde{Y}_{k+1}$ given $\widetilde{Y}_{k} = y_k$ is
\begin{align}\label{eq:def-Ytilde}
p_{\widetilde{Y}_{k+1}\mymid\widetilde{Y}_{k}}(y\mymid y_k) = \left\{
\begin{array}{ll}
p_{{Y}_{k+1}\mymid {Y}_{k}}(y\mymid y_k) \mathds{1}(y\in\mathcal{E}_{k+1}) & \\
\quad + \int_{\mathcal{E}_{k+1}^{\rm c}} p_{{Y}_{k+1}\mymid {Y}_{k}}(y_{k+1}\mymid y_k) \mathrm d y_{k+1} \delta_{\infty},& y_k\neq \infty,\\
\delta_{\infty},&y_k = \infty,
\end{array}
\right.
\end{align}
where according to the algorithm design
%\begin{align*}
$
P_{Y_{k+1}|Y_k}(y|{y}_k) = \phi\left(y|y_{\tau_{k,N}}({y}_k),\sigma_k^2\right),
$
%\end{align*}
with $\sigma_k^2$ defined in \eqref{eq:def-sigma-k}.
The following lemma states some basic properties about the above auxiliary sequences.
The proof is postponed to Appendix \ref{subsec:proof-lem-prop-seq}  in supplemental material.
\begin{lemma}\label{lem:prop-seq}
The following properties hold:
\begin{align}
p_{\widetilde{X}_{k}}(x) &= 0, \quad p_{\widetilde{Y}_{k}}(y) = 0,\quad {\rm for}~x,y\in\mathcal{E}_k^{\rm c};\label{eq:lem-prop-seq-1}\\
p_{\widetilde{X}_{k}}(x)&\le p_{\widehat{X}_{k}}(x),\quad 
p_{\widetilde{Y}_{k}}(y)\le p_{Y_{k}}(y),\quad {\rm for}~ x,y\neq \infty.\label{eq:lem-prop-seq}
%p_{\overline{X}_{k}}(\overline{x}_{k})&\le p_{\widetilde{X}_k}(\overline{x}_{k}),\qquad \overline{x}_{k}\neq \infty.\label{eq:lem-prop-seq-xbar}
\end{align}
\end{lemma}

\subsection{Step 2: decompose the error terms}

Recalling the definition of $\widehat{X}_k$, 
and the fact that $p_{\widetilde{Y}_{K}}(x)\le p_{Y_{K}}(x)$ and $p_{\widetilde{X}_{K}}(x)\le p_{\widehat{X}_{K}}(x)$, we have
\begin{align}\label{eq:proof-decomp-error-1}
\mathsf{TV}\big(q_K,p_{Y_{K}}\big) &= \mathsf{TV}\big(p_{\widehat{X}_{K}}, p_{Y_{K}}\big) 
= \int(p_{\widehat{X}_{K}}(x)-p_{Y_{K}}(x))\mathds{1}\{p_{\widehat{X}_{K}}(x)> p_{Y_{K}}(x)\} \mathrm d x \nonumber\\
&\le \int(p_{\widehat{X}_{K}}(x)-p_{\widetilde{Y}_{K}}(x))\mathds{1}\{p_{\widehat{X}_{K}}(x)> p_{\widetilde{Y}_{K}}(x)\} \mathrm d x \nonumber\\
&= \int(p_{\widetilde{Y}_{K}}(x)-p_{\widehat{X}_{K}}(x))\mathds{1}\{p_{\widetilde{Y}_{K}}(x)>p_{\widehat{X}_{K}}(x)\} \mathrm d x + P(\widetilde{Y}_{K} = \infty)\nonumber\\
&\le \int(p_{\widetilde{Y}_{K}}(x)-p_{\widetilde{X}_{K}}(x))\mathds{1}\{p_{\widetilde{Y}_{K}}(x)>p_{\widetilde{X}_{K}}(x)\} \mathrm d x + P(\widetilde{Y}_{K} = \infty)\nonumber\\
& \overset{\text{(a)}}{\le}  \mathsf{TV}(p_{\widetilde{Y}_{K}},p_{\widetilde{X}_{K}}) + P(\widetilde{X}_{K} = \infty)\nonumber\\
&\overset{\text{(b)}}{\le}\mathsf{TV}(p_{\widetilde{Y}_{K}},p_{\widetilde{X}_{K}})+ \mathsf{TV}(p_{\widetilde{X}_K},p_{\widehat{X}_K}),
\end{align}
where (a) uses the fact that
\begin{align*}
\mathsf{TV}(p_{\widetilde{Y}_{K}},p_{\widetilde{X}_{K}})
&=\int(p_{\widetilde{Y}_{K}}(x)-p_{\widetilde{X}_{K}}(x))\mathds{1}\{p_{\widetilde{Y}_{K}}(x)>p_{\widetilde{X}_{K}}(x)\} \mathrm d x\nonumber\\
&\quad + \max\{P(\widetilde{Y}_{K} = \infty) - P(\widetilde{X}_{K} = \infty),0\},
\end{align*}
and (b) uses the fact that 
\begin{align*}%\label{eq:proof-decomp-error-2}
P(\widetilde{X}_{K} = \infty) \le \mathsf{TV}(p_{\widetilde{X}_{K}},p_{\widehat{X}_K}).
\end{align*}

Now we intend to bound the two terms in the right-hand-side of \eqref{eq:proof-decomp-error-1} separately.
For the first term, by using Pinsker's inequality, we have
\begin{align}
&\quad \mathsf{TV}^2(p_{\widetilde{Y}_{K}},p_{\widetilde{X}_{K}})
\le \frac12\mathsf{KL}\left(p_{\widetilde{X}_{K}}\Vert p_{\widetilde{Y}_{K}}\right) \le \frac12\mathsf{KL}\left(p_{\widetilde{X}_{0},\ldots,\widetilde{X}_{K}}\Vert p_{\widetilde{Y}_{0},\ldots,\widetilde{Y}_{K}}\right) \nonumber\\
&= \frac12\mathsf{KL}\left(p_{\widetilde{X}_{0}}\Vert p_{\widetilde{Y}_{0}}\right) + \sum_{k=0}^{K-1}\mathbb{E}_{x_k\sim\widetilde{X}_k}\mathsf{KL}(p_{\widetilde{X}_{k+1}\mymid\widetilde{X}_k}(\cdot\mymid x_k)\Vert p_{\widetilde{Y}_{k+1}\mymid\widetilde{Y}_k}(\cdot\mymid x_k))\nonumber\\
&\overset{\text{(a)}}{\le} \frac12\mathsf{KL}\left(p_{\widehat{X}_{0}}\Vert p_{{Y}_{0}}\right) + \sum_{k=0}^{K-1}\mathbb{E}_{x_k\sim\widetilde{X}_k}\mathsf{KL}(p_{\widehat{X}_{k+1}\mymid\widehat{X}_k}(\cdot\mymid x_k)\Vert p_{{Y}_{k+1}\mymid{Y}_k}(\cdot\mymid x_k)),\label{eq:KL-infty}
\end{align}
where (a) is proved in Appendix \ref{subsec:proof-eq-KL-infty} of supplemental material.
For the second term, considering $x\in \mathcal{E}_k$, we have
\begin{align}
p_{\widetilde{X}_k}(x) 
&= \int_{\mathcal{E}_{k-1}} p_{\widetilde{X}_k\mymid\widetilde{X}_{k-1}}(x\mymid x_{k-1})p_{\widetilde{X}_{k-1}}(x_{k-1}) \mathrm d x_{k-1}\nonumber\\
&= \int_{\mathcal{E}_{k-1}} p_{\widetilde{X}_k\mymid\widetilde{X}_{k-1}}(x\mymid x_{k-1})p_{\widehat{X}_{k-1}}(x_{k-1}) \mathrm d x_{k-1} - \Delta_{k}(x)\nonumber\\
&= \int_{\mathcal{E}_{k-1}} p_{\widehat{X}_k\mymid\widehat{X}_{k-1}}(x\mymid x_{k-1})p_{\widehat{X}_{k-1}}(x_{k-1}) \mathrm d x_{k-1} - \Delta_{k}(x)\nonumber\\
&=p_{\widehat{X}_k}(x) - \int_{\mathcal{E}_{k-1}^{\rm c}} p_{\widehat{X}_k\mymid\widehat{X}_{k-1}}(x\mymid x_{k-1})p_{\widehat{X}_{k-1}}(x_{k-1}) \mathrm d x_{k-1} - \Delta_{k}(x),\label{eq:proof-temp-6}
\end{align}
where 
$$
\Delta_k(x) = \int_{\mathcal{E}_{k-1}} p_{\widetilde{X}_k\mymid\widetilde{X}_{k-1}}(x\mymid x_{k-1})\left(p_{\widehat{X}_{k-1}}(x_{k-1})-p_{\widetilde{X}_{k-1}}(x_{k-1})\right) \mathrm d x_{k-1}.
$$
Thus we have
\begin{align}\label{eq:proof-temp-7}
\mathsf{TV}(p_{\widetilde{X}_k},p_{\widehat{X}_k}) 
&= \int p_{\widehat{X}_k}(x)-p_{\widetilde{X}_k}(x) \mathrm d x\nonumber\\
%\mathds{1}\{p_{\widehat{X}_k}(x)>p_{\widetilde{X}_k}(x)\} \mathrm d x
&= \int_{\mathcal{E}_k^{\rm c}} p_{\widehat{X}_k}(x)\mathrm d x + \int_{\mathcal{E}_k} p_{\widehat{X}_k}(x)-p_{\widetilde{X}_k}(x) \mathrm d x\nonumber\\
&\overset{\text{(a)}}{=} P(\widehat{X}_k\in\mathcal{E}_k^{\rm c}) + \int_{\mathcal{E}_k} \int_{\mathcal{E}_{k-1}^{\rm c}} p_{\widehat{X}_k\mymid\widehat{X}_{k-1}}(x\mymid x_{k-1})p_{\widehat{X}_{k-1}}(x_{k-1}) \mathrm d x_{k-1} \mathrm d x\nonumber\\
&+ \int_{\mathcal{E}_k}\Delta_k(x) \mathrm d x,
\end{align}
where (a) inserts \eqref{eq:proof-temp-6}.
Notice that
\begin{align*}
&\int_{\mathcal{E}_k} \int_{\mathcal{E}_{k-1}^{\rm c}} p_{\widehat{X}_k\mymid\widehat{X}_{k-1}}(x\mymid x_{k-1})p_{\widehat{X}_{k-1}}(x_{k-1}) \mathrm d x_{k-1} \mathrm d x
\le \int_{\mathcal{E}_{k-1}^{\rm c}} p_{\widehat{X}_{k-1}}(x_{k-1}) \mathrm d x_{k-1} \notag\\
&\qquad\qquad \qquad\qquad\qquad\qquad\qquad\qquad\qquad\qquad= P(\widehat{X}_{k-1}\in\mathcal{E}_{k-1}^{\rm c}),\nonumber\\
&\int_{\mathcal{E}_k}\Delta_k(x) \mathrm d x
=\int_{\mathcal{E}_k}\int_{\mathcal{E}_{k-1}} p_{\widetilde{X}_k\mymid\widetilde{X}_{k-1}}(x\mymid x_{k-1})\left(p_{\widehat{X}_{k-1}}(x_{k-1})-p_{\widetilde{X}_{k-1}}(x_{k-1})\right) \mathrm d x_{k-1}\mathrm d x\nonumber\\
&\qquad \qquad \qquad \le \int_{\mathcal{E}_{k-1}} p_{\widehat{X}_{k-1}}(x_{k-1})-p_{\widetilde{X}_{k-1}}(x_{k-1}) \mathrm d x_{k-1} \le \mathsf{TV}(p_{\widetilde{X}_{k-1}},p_{\widehat{X}_{k-1}}).
\end{align*}
Inserting into \eqref{eq:proof-temp-7} and by using recursion, we have
\begin{align}\label{eq:proof-decomp-error-4}
\mathsf{TV}(p_{\widetilde{X}_K},p_{\widehat{X}_K}) 
&\le  \mathsf{TV}(p_{\widetilde{X}_{K-1}},p_{\widehat{X}_{K-1}}) + P(\widehat{X}_{K-1}\in\mathcal{E}_{K-1}^{\rm c}) + P(\widehat{X}_{K}\in\mathcal{E}_{K}^{\rm c})\notag\\
&\le 2\sum_{k=0}^{K-1} P(\widehat{X}_{k}\in\mathcal{E}_{k}^{\rm c})
\end{align}
considering that
\begin{align*}
\mathsf{TV}(p_{\widetilde{X}_0},p_{\widehat{X}_0}) = P(\widehat{X}_0\in\mathcal{E}_0),\quad P(\widehat{X}_{K}\in\mathcal{E}_{K}^{\rm c})=0.
\end{align*}

Inserting \eqref{eq:KL-infty} and \eqref{eq:proof-decomp-error-4} into \eqref{eq:proof-decomp-error-1}, we have
\begin{small}
\begin{align}\label{eq:error-decomposition-result}
\mathsf{TV}\big(q_K, p_{Y_{K}}\big)
&\le \sqrt{\frac12\mathsf{KL}\big(p_{\widehat{X}_{0}}\Vert p_{{Y}_{0}}\big)+\frac12\sum_{k=0}^{K-1}\mathbb{E}_{x_{k}\sim p_{\widetilde{X}_{k}}}\left[\mathsf{KL}\big(p_{\widehat{X}_{k+1}|\widehat{X}_{k}}\left(\,\cdot\mymid x_{k}\right)\,\Vert\,p_{{Y}_{k+1}|{Y}_{k}}\left(\,\cdot\mymid x_{k}\right)\big)\right]}\nonumber\\
&\quad + 2\sum_{k=0}^{K-1} P(\widehat{X}_{k}\in\mathcal{E}_{k}^{\rm c}).
\end{align}
\end{small}

\subsection{Step 3: control the KL divergence between $p_{\widehat{X}_{k+1}|\widehat{X}_{k}}$ and $p_{Y_{k+1}|Y_{k}}$}

% Notice that given an initialization point $x_{\tau_{k,0}}=x_k$, the term $\theta(\tau;x_k):=\frac{s_\tau^{\star}(x_\tau)}{2(1-\tau)^{3/2}}$ depends only on $\tau$. Solving \eqref{eq:ODE} is equivalent to calculating the integral of $\theta(\tau;x_k)$. Consequently, it is natural for the algorithm to approximate the integral by discretizing $\tau$ and estimating $\theta(\tau;x_k)$ at some discrete time points. 
%Due to the injection of noise, the approximation error determines 
The KL divergence between conditional distributions $p_{\widehat{X}_{k+1}|\widehat{X}_k}$ and $p_{{Y}_{k+1}|{Y}_k}$ is
\begin{small}
\begin{align}\label{eq:KL-condition}
\mathsf{KL}\big(p_{\widehat{X}_{k+1}|\widehat{X}_{k}}\left(\,\cdot\mymid x_{k}\right)\,\Vert\,p_{Y_{k+1}|Y_{k}}\left(\,\cdot\mymid x_{k}\right)\big)
&= \frac{1-\tau_{k+1, 0}}{2(\tau_{k+1, 0}-\tau_{k, N})}\|y_{\tau_{k, N}}(x_k) - x_{\tau_{k, N}}(x_k)\|_2^2.
\end{align}
\end{small}
This can be immediately verified by recalling that  $\widehat{X}_{k+1}|\widehat{X}_{k}=x_{k}$ and $Y_{k+1}|Y_{k}=x_{k}$ are both normal distributions with the same variance $\frac{\tau_{k+1,0}-\tau_{k,N}}{1-\tau_{k,N}}$ and different means $\sqrt{\frac{1-\tau_{k+1, 0}}{1-\tau_{k, N}}}x_{\tau_{k, N}}(x_k)$ and $\sqrt{\frac{1-\tau_{k+1, 0}}{1-\tau_{k, N}}}y_{\tau_{k, N}}(x_k)$, respectively.

Thus the core of this step is to control the estimation error.
Before proceeding, we introduce additional two sequences of auxiliary variables
\begin{align*}
\frac{y_{\tau_{k, n}}^{\star}(x_{\tau_{k,0}})}{\sqrt{1-\tau_{k, n}}} &= \frac{x_{\tau_{k, 0}}}{\sqrt{1-\tau_{k,0}}} 
+ \frac{s_{\tau_{k,0}}^{\star}(x_{\tau_{k, 0}})}{2(1-\tau_{k,0})^{3/2}}(\tau_{k, 0} - \widehat{\tau}_{k,0})\\
&\quad
+ \sum_{i = 1}^{n-1} \frac{s_{\tau_{k,i}}^{\star}(y_{\tau_{k,i}})}{2(1-\tau_{k,i})^{3/2}}(\widehat{\tau}_{k,i-1} - \widehat{\tau}_{k,i}) + \frac{s_{\tau_{k,n-1}}^{\star}(y_{\tau_{k,n-1}})}{2(1-\tau_{k,n-1})^{3/2}}(\widehat{\tau}_{k,n-1} - \tau_{k, n}),\\
\frac{z_{\tau_{k,n}}^{\star}(x_{\tau_{k,0}})}{\sqrt{1-\tau_{k,n}}} &= \frac{x_{\tau_{k, 0}}}{\sqrt{1-\tau_{k,0}}}
+ \frac{s_{\tau_{k,0}}^{\star}(x_{\tau_{k, 0}})}{2(1-\tau_{k,0})^{3/2}}(\tau_{k, 0} - \widehat{\tau}_{k,0})\\
&\quad + \sum_{i = 1}^{n-1} \frac{s_{\tau_{k,i}}^{\star}(x_{\tau_{k,i}})}{2(1-\tau_{k,i})^{3/2}}(\widehat{\tau}_{k,i-1} - \widehat{\tau}_{k,i})+ \frac{s_{\tau_{k,n-1}}^{\star}(x_{\tau_{k,n-1}})}{2(1-\tau_{k,n-1})^{3/2}}(\widehat{\tau}_{k,n-1} - \tau_{k, n}),
\end{align*}
for $n = 1, \ldots, N$.
For convenience, in the following proof we shall omit the dependence of $x_{\tau_{k,n}}$, $y_{\tau_{k,n}}$, $y_{\tau_{k,n}}^{\star}$, and $z_{\tau_{k,n}}^{\star}$ on $x_{\tau_{k,0}}$ without ambiguity.
Based on these notations, we define the estimation error
%the discretization and estimation errors of the ODE process~\eqref{eq:ODE} as
\begin{align}\label{eq:def-xi}
\xi_{k, n}(x_{\tau_{k,0}}) &:= \frac{y^{\star}_{\tau_{k, n}} - y_{\tau_{k, n}}}{\sqrt{1-\tau_{k,n}}}+\frac{x_{\tau_{k, n}}-z^{\star}_{\tau_{k, n}}}{\sqrt{1-\tau_{k,n}}}.
\end{align}

Moreover, we define the matrix
\begin{align*}
\Sigma_\tau(x) = \mathsf{Cov}[Z\mymid \sqrt{1-\tau}X_0 + \sqrt{\tau}Z = x],
\end{align*}
where $\mathsf{Cov}[\cdot]$ denotes the covariance matrix.
The following lemma controls the estimation error $\xi_{k,n}$. 
The proof is postponed to Appendix~\ref{sec:proof-lem:discrete}.
% Similar errors are also analyzed in \citet{Gupta2024Faster} (Appendix A.1) under an exponential integrator formulation. Our derivation adapts these ideas to the context of the probability flow ODE discretization in this work, and additional efforts are made to overcome the challenges brought by non-uniform Lipschitz condition, as discussed in Appendix \ref{app:comparison} of supplemental material.
\begin{lemma}
\label{lem:discrete}
For any $k$ and $n$, with probability at least $1 - T^{-100}$,
\begin{align}
\mathbb{E}_{x_{\tau_{k,0}} \sim p_{\widehat{X}_{k}}}\big[\|\xi_{k, n}(x_{\tau_{k,0}})\|_2^2\big] 
&\lesssim \frac{d\log^4 T}{T^3}\min\Big\{\frac{Nd\widehat{\tau}_{k,-1}\log T}{T(1-\widehat{\tau}_{k,-1})}+\int_{{\tau}_{k,n}}^{{\tau}_{k,0}} \frac{\mathbb{E}[\mathsf{Tr}(\Sigma_\tau^2(x_\tau))]}{(1-\tau)^2}\mathrm d \tau,\nonumber\\
&\quad\frac{NL^2\widehat{\tau}_{k,-1}\log T}{T(1-\widehat{\tau}_{k,-1})}\Big\} +\frac{N\log^2 T}{T^2}\sum_{i = 0}^{n-1} \widehat{\tau}_{k,i}(1-\widehat{\tau}_{k,i})^{-1}\varepsilon_{k,i}^2,
\end{align}
where $\varepsilon_{k,i}^2$ is defined in~\eqref{eq:score-error}.
\end{lemma}

With the above relation, we can bound the divergence as following.
The proof is postponed to Appendix~\ref{sec:proof-lem:KL} in supplemental material.
\begin{lemma}
\label{lem:KL}
According to Lemma~\ref{lem:discrete}, it can be shown that
\begin{align} \label{eq:KL}
&\quad \sum_{k=0}^{K-1}\mathbb{E}_{x_{\tau_{k,0}}\sim p_{\widetilde{X}_{k}}}\left[\mathsf{KL}\left(p_{\widehat{X}_{k+1}|\widehat{X}_{k}}\left(\,\cdot\mymid x_{\tau_{k,0}}\right)\,\Vert\,p_{Y_{k+1}|Y_{k}}\left(\,\cdot\mymid x_{\tau_{k,0}}\right)\right)\right]\nonumber\\
&\lesssim \frac{Kd\log^5 T}{T^3}\min\Big\{d,L^2\Big\}
+\varepsilon_{\mathsf{score}}^2\log T,
\end{align}
where $\varepsilon_{\mathsf{score}}^2$ is defined in~\eqref{eq:score-error}.
\end{lemma}

\subsection{Step 4: putting everything together}

The remaining terms in~\eqref{eq:error-decomposition-result} can be bounded through the following lemma, whose proof can be found in Appendix~\ref{sec:proof-lem-endpoint} in supplemental material.
\begin{lemma} 
\label{lem:endpoint}
Under our choice of learning schedule~\eqref{eq:learning-rate}, we have
\begin{align} 
\mathsf{KL}\big(p_{\widehat{X}_{0}}\Vert p_{\widehat{Y}_{0}}\big) &\le \frac{1}{T^{10}},\label{eq:prob-endpoints}\\
\sum_{k=0}^{K-1}P(\widehat{X}_{k}\in\mathcal{E}_{k}^{\rm c})
&\lesssim \left(\frac{d^2\log^5 T}{T^2} + \varepsilon_{\mathsf{score}}^2\log T\right)\mathds{1}(d\log T<L).
\label{eq:prob-setE}
\end{align}
\end{lemma}
Inserting~\eqref{eq:KL}, \eqref{eq:prob-endpoints}, and \eqref{eq:prob-setE} into~\eqref{eq:error-decomposition-result} leads to
\begin{align*}
\mathsf{TV}\left(q_K, p_{Y_{K}}\right) 
&\lesssim \sqrt{\frac{1}{T^{10}} +  \frac{Kd\log^5 T}{T^3}\min\Big\{d,L^2\Big\}
+\varepsilon_{\mathsf{score}}^2\log T}\nonumber\\
&\quad +\left(\frac{d^2\log^5 T}{T^2} + \varepsilon_{\mathsf{score}}^2\log T\right)\mathds{1}(d\log T<L) \nonumber\\
%&\lesssim \sqrt{\frac{1}{T^{10}} +  \frac{Kd\log^4 T}{T^3}\min\Big\{d,L^2\Big\}+\varepsilon_{\mathsf{score}}^2\log T}\nonumber\\
&\lesssim \frac{\min\{d^{3/2}, dL^{1/2},d^{1/2}L^{3/2}\}\log^{4} T}{T^{3/2}} + \varepsilon_{\mathsf{score}}\log^{1/2} T,
\end{align*}
and we conclude the proof here.

\section{Discussion}
\label{sec:conclusion}

In this paper, we establish a faster convergence rate for generative diffusion models under a relaxed Lipschitz condition, which requires a number of $\min\{d,d^{2/3}L^{1/3},d^{1/3}L\}\varepsilon^{-2/3}\log^{8/3}T$ iterations to achieve an $\varepsilon$ accuracy in terms of TV distance,
where $L$ denotes the non-uniform Lipschitz constant.
%Our theory is instance-dependent and adapts to the smoothness $L$ of score functions normalized by the noise variances.
As a result, our results demonstrate improvements over the convergence theory without exploiting smoothness across the entire range of $L$, and accommodates a broader class of distributions, e.g., Gaussian mixture model.
In addition, our analysis requires only $\varepsilon$ estimation errors, implying the robustness of the algorithm to imperfect score estimations.
Furthermore, we extend the result to the parallel implementation of the sampler.
However, several open questions remain. For example, deriving instance-dependent bounds for other settings, such as accelerated samplers or data distributions with low-dimensional structures, which we leave for future research.
Moreover, the benefit of the randomized design in our work relies heavily on the deterministic nature of ODE process. It is still unclear for us how to adapt this approach to deal with the inherent stochasticity in SDEs while maintaining similar improvements. This will also be left as future work.
In addition, it may be feasible to apply your analysis framework to improve the bound for other variants of samplers such as the Langevin algorithm.
Furthermore, estimating the Lipschitz constant in real-world cases would be highly beneficial, potentially broadening the applicability of our results.

\subsection*{Acknowledgments}

Gen Li is supported in part by the Chinese University of Hong Kong Direct Grant for Research and the Hong Kong Research Grants Council ECS 2191363.

\appendix
\section{Numerical experiments}
\label{app:simulations}

We conduct numerical experiments to validate our theoretical results.
For ease of computing, we select a Gaussian distribution as the target distribution.
This choice ensures that all $Y_{k,n}$ in the implementation of the proposed sampler follow a Gaussian distribution and that the KL divergence between $Y_{k,0}$ and $X_1$ has a closed-form expression. 
Moreover, since our primary focus is on the convergence rate, we assume access to the exact score function $s_t^{\star}(\cdot)$.

The target distribution $p_0$ is a $d$-dimensional Gaussian distribution with zero mean and a diagonal covariance matrix. The first $k$ diagonal entries are uniformly distributed within the interval $[0,10]$, while the remaining $d-k$ diagonal entries are set to zero.
We implement the proposed sampler in Section \ref{subsec:Algorithm} with $K=10$, and $N=2T/K$. For different number of iterations $T$, we compute the distribution of output $Y_K$, and its KL divergence with the distribution $q_K$ of $X_{\tau_{K,0}}$, which is approximately the starting point of the forward process.

\begin{figure}
\begin{center}
\includegraphics[width=\textwidth]{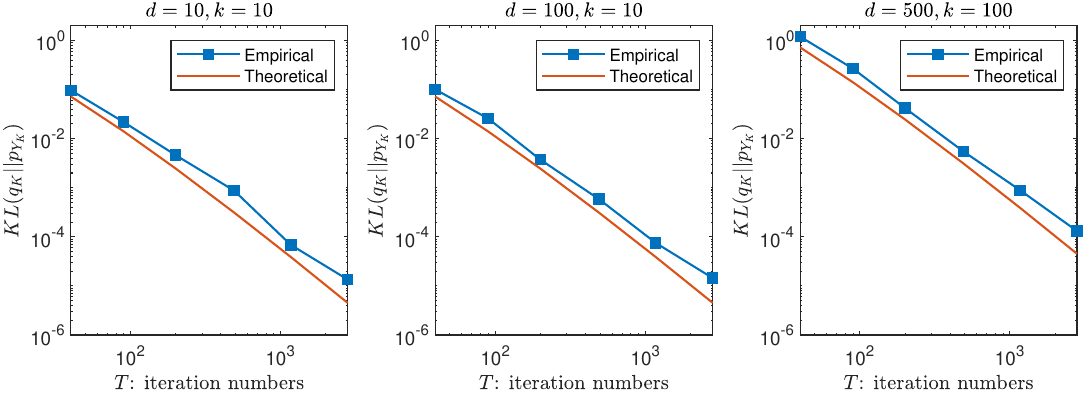}
\end{center}
\caption{Sampling error of the proposed sampler and fitted rate $T \rightarrow \Theta(\log^4 T/T^3)$:
(a) $d = 10, k = 10$; (b) $d = 100, k = 10$; (c) $d = 500, k = 100$. \label{fig:exp}}
\end{figure}

The results are presented in Figure \ref{fig:exp}.
The blue line represents the empirical results, and the black line corresponds to the theoretical rate $O(\log^4T/T^3)$.
According to Theorem \ref{thm:main}, 
our theoretical analysis predicts a convergence rate of $O(\mathsf{poly}(\log T)/T^3)$ in terms of KL divergence, which is consistent with empirical observations. This further confirms that our sampler achieves a KL divergence convergence rate of $O(\log^4 T/T^3)$ in terms of KL divergence, implying a total variation(TV) distance convergence rate of $O(\log^2 T/T^{3/2})$.
Finally, we remark that Theorem \ref{thm:main} establishes a convergence rate of $O(\log^{4}T/T^{3/2})$.
Compared to empirical results, this bound is suboptimal in terms of its dependence on logarithmic factors. Refining this dependency requires further effort and is left for future work.

\section{Comparison with previous works}
\label{app:comparison}

\begin{figure}
\begin{center}
\includegraphics[width=0.9\textwidth]{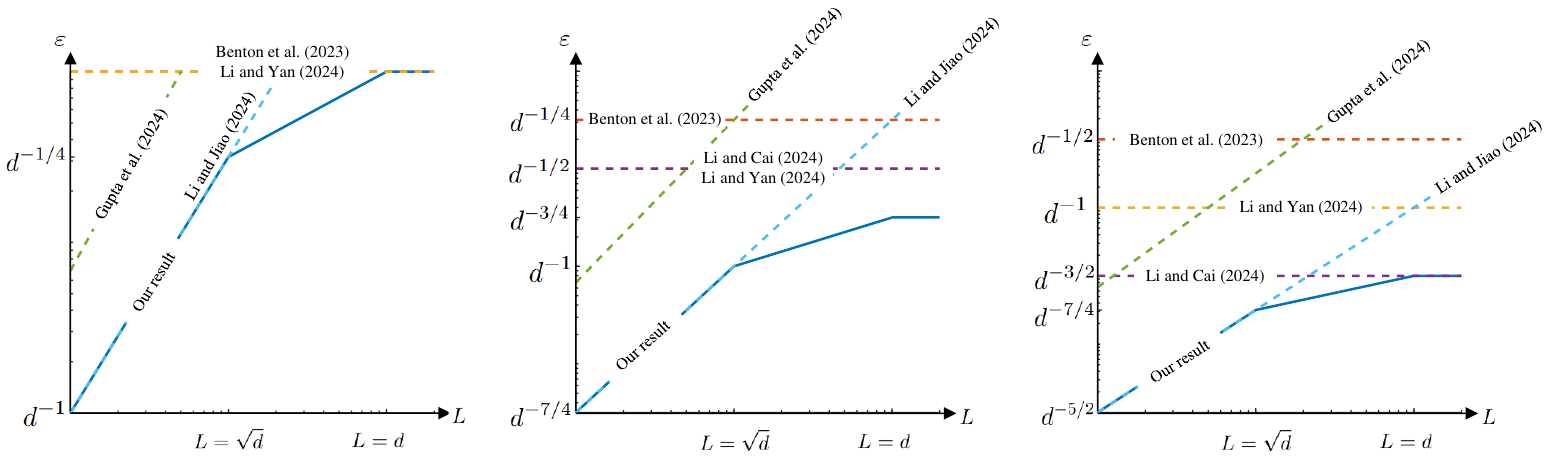}
\end{center}
\caption{TV distance $\varepsilon$ achieved by Theorem \ref{thm:main} and previous results with left: $T = O(d)$; middle: $T=O(d^{3/2})$; right: $T=O(d^2)$. \label{fig:comp-app}}
\end{figure}

To compare Theorem \ref{thm:main} with previous results, we illustrate the TV distance achieved by various theories for a fixed number of iterations $T$.
The corresponding results are presented in Figure \ref{fig:comp-app}.
Notably, when $T=O(d)$, the results from \citet{li2024provable,Benton2023Nearly,li2024d} reduce to a trivial bound $\varepsilon=1$.
In comparison, for both $T=O(d)$ and $T=O(d^2)$, our result achieves the best result across a full range of~$L$, and improves previous results when $\sqrt{d}\lesssim L\lesssim d$.

A lot of technical efforts have been devoted to achieving the improvements stated in Section \ref{sec:mainresults}.
\begin{itemize}
\item[a.] The absence of a uniform Lipschitz bound (see Definition \ref{def:score-lipschitz}) brings new challenges to the discretization analysis.
Prior works \citep{Chen2023The,li2024improved} leveraged the log-concavity of $p_{X_{\tau}|X_{\tau+\delta}}$ for some small $\delta$ to control the one-step discretization error (see~Lemma 1 and (B.2)-(B.4) in \citet{Chen2023The}).
However, this approach fails under the high-probability bound employed in this work.
To address this issue, we directly handle this derivative based on its definition, decomposing it into two components: one depending on the operator norm of the Jacobian matrix and the other on the data dimension $d$, and then statistically bounding each term (see Lemma \ref{lem:bound-der-tau}).
Although the resulting bound is of a similar order, the analytical techniques under the uniform bound assumption and the high-probability bound assumption are fundamentally different.

\item[b.] The lack of a uniform Lipschitz condition poses significant challenges for controlling error propagation across multiple steps, while a naive stepwise analysis may result in suboptimal bounds.
To overcome this challenge, we introduce two auxiliary sequences, $\widetilde{X}_k$ and $\widetilde{Y}_k$, which constrain $\widehat{X}_k$ and $Y_k$ to lie within a typical set (see Step 1 in Section \ref{sec:analysis} and Lemma \ref{lem:prop-seq}). We then relate the TV distance between $\widehat{X}_k$ and $Y_k$ to that between $\widetilde{X}_k$ and $\widetilde{Y}_k$ (see Step 2 in Section \ref{sec:analysis}), by analyzing how error propagation affects the probability that $Y_{k,n}|{Y}_{k,0} = x_{\tau_{k,0}}$ falls outside the typical set (see Lemma \ref{lem:endpoint}). Within this typical set, we establish a uniform bound for score functions that depends explicitly on the data dimension $d$ (see Lemma \ref{lem:lipschitz}).
As a result, the error at the $(k,n)$-th step, $\|y_{\tau_{k,n}} - x_{\tau_{k,n}}\|$, is effectively controlled by the discretization error $\zeta_{k,n}$ and estimation error terms related to $\widetilde{\varepsilon}_{k,i}$ (see \eqref{eq:lem-KL-pre-1}), thereby ensuring stable error propagation throughout the process.

\item[c.] To make the result adaptive to $L$ and applicable for the minimal condition on the target data distribution (i.e., $L=\infty$), new bounds for the discretization error and the number of rounds $K$ are derived.
Based on the error propagation analysis discussed earlier, the lower bound for $K$ is $O(\min\{d\log T,L\}\log T)$, which remains bounded even when $L=\infty$.
Additionally, to obtain a discretization error adaptive to $L$, we derive a new bound for the expectation of the product $\|J_\tau(x_\tau)s_\tau^{\star}(x_\tau)\|_2^2$ by carefully analyzing the structure of the Jacobian matrix $J_\tau(\cdot)$ (see Lemma \ref{lem:bound-s-J}).
The new bound $\widetilde{O}(d\min\{d,L^2\})$ helps to derive a corresponding discretization error of $\widetilde{O}(d\min\{d,L^2\})$ (see Lemma \ref{lem:discrete-pre}), which depends only on $d$ when $L$ is large.
%This improvement enables the minimal condition on the target data distribution, specifically when $L=\infty$.
\end{itemize}

\section{Computations of Examples }
\label{app:examples}

\subsection{Computation of Example \ref{example:Gaussian}}
\label{app:examples-Gaussian}

It is easy to check that the score function is
\begin{align*}
s_t^{\star}(x) = -\Sigma_t^{-1}x,
\end{align*}
where $\Sigma_t$ is a diagonal matrix with the $(i,i)$-th entry equal to 
$$
(\Sigma_t)_{i,i} = \overline{\alpha}_t\sigma_i^2 + 1-\overline{\alpha}_t.
$$
Thus we have
\begin{align*}
\|s_t^{\star}(x) - s_t^{\star}(y)\|_2 = \|\Sigma_t^{\rm -1}(x-y)\|_2,
\end{align*}
and 
\begin{align*}
%J_t(X) = -\Sigma_t^{-1}\qquad {\rm and} \qquad 
\|\Sigma_t^{-1}\|_2 = \frac{1}{\overline{\alpha}_t\min\sigma_i^2 + 1-\overline{\alpha}_t} = \frac{1}{1-\overline{\alpha}_t}.
\end{align*}
Then we complete the proof.
%have \eqref{eq:example-Gaussian-1} and \eqref{eq:example-Gaussian-2} hold obviously.

\subsection{Computation of Example \ref{example:GMM}}
\label{app:examples-GMM}

It is easy to check that the score function is
\begin{align*}
s_t^{\star}(x) = -\frac{1}{\sigma_t^2}\sum_{h=1}^H \pi_h(x)(x-\sqrt{\overline{\alpha}_t}\mu_h) = -\frac{x}{\sigma_t^2} + \frac{\sqrt{\overline{\alpha}_t}}{\sigma_t^2}\sum_{h=1}^H\pi_h(x)\mu_h,
\end{align*}
where
\begin{align*}
\pi_h(x) = \frac{\gamma_h\exp\left(-\frac{1}{\sigma_t^2}\|x-\sqrt{\overline{\alpha}_t}\mu_h\|^2\right)}{\sum_{i=1}^H\gamma_i\exp\left(-\frac{1}{\sigma_t^2}\|x-\sqrt{\overline{\alpha}_t}\mu_i\|^2\right)},\qquad{\rm and}\qquad  \sigma_t^2 = \overline{\alpha}_t\sigma^2 + 1-\overline{\alpha}_t.
\end{align*}

\subsubsection{Proof of the upper bound}

For ease of notations, we prove that for any $\mu_h,\sigma$, and $\gamma_h$, the following inequality holds:
\begin{align*}
\mathbb{P}\left\{\exists x'\in\mathbb{R}^d, (1-\overline{\alpha}_t)\|s_t^{\star}(x)-s_t^{\star}(x')\|_2 > C\log(HT)\|x-x'\|_2\right\} \lesssim \frac{1}{T^4}.
\end{align*}
The upper bound can be proved by replacing $T$ with $T+d$. 
%One can easily prove that if $d\ge T$, then
% \begin{align*}
% \mathbb{P}\left\{\exists x'\in\mathbb{R}^d, (1-\overline{\alpha}_t)\|s_t^{\star}(x)-s_t^{\star}(x')\|_2 > C\log(Hd)\|x-x'\|_2\right\} \lesssim \frac{1}{d^4}.
% \end{align*}
% Combining these two inequalities can complete the proof.

According to the definition of $s_t^{\star}(x)$, We have
\begin{align*}
&\quad (1-\overline{\alpha}_t)\|s_t^{\star}(x)-s_t^{\star}(x')\|_2 \notag\\
&\overset{\text{(a)}}{=} \left\|-\frac{1-\overline{\alpha}_t}{\sigma_t^2}(x-x') + \frac{\sqrt{\overline{\alpha}_t}(1-\overline{\alpha}_t)}{\sigma_t^2}\sum_{i=1}^H (\pi_i(x)-\pi_i(x'))(\mu_i-\mu_h)\right\|_2\nonumber\\
&\le \frac{1-\overline{\alpha}_t}{\sigma_t^2}\|x-x'\|_2 + \frac{\sqrt{\overline{\alpha}_t}(1-\overline{\alpha}_t)}{\sigma_t^2}\sum_{i=1}^H \|(\pi_i(x)-\pi_i(x'))(\mu_i-\mu_h)\|_2\nonumber\\
&\overset{\text{(b)}}{\le} \|x-x'\|_2 + \sqrt{\overline{\alpha}_t}\sum_{i=1}^H \|(\pi_i(x)-\pi_i(x'))(\mu_i-\mu_h)\|_2,
\end{align*}
where (a) uses the fact that $\sum_{i=1}^H (\pi_i(x)-\pi_i(x'))(\mu_i-\mu_h) = \sum_{i=1}^H (\pi_i(x)-\pi_i(x'))\mu_i$,
and (b) use the fact that $1-\overline{\alpha}_t\le \sigma_t^2$.
Thus we have
\begin{align}\label{eq:proof-examples-temp-1}
&\quad \mathbb{P}\left\{\exists x'\in\mathbb{R}^d, (1-\overline{\alpha}_t)\|s_t^{\star}(x)-s_t^{\star}(x')\|_2 >C\log(HT)\|x-x'\|_2\right\}\nonumber\\
&\le \sum_{h=1}^H\gamma_h\mathbb{P}_h\left\{\exists x'\in\mathbb{R}^d, (1-\overline{\alpha}_t)\|s_t^{\star}(x)-s_t^{\star}(x')\|_2 >C\log(HT)\|x-x'\|_2\right\}\nonumber\\
&\le \sum_{h=1}^H\gamma_h\mathbb{P}_h\left\{\exists x'\in\mathbb{R}^d, \sqrt{\overline{\alpha}_t}\sum_{i=1}^H \|(\pi_i(x)-\pi_i(x'))(\mu_i-\mu_h)\|_2 >\frac{C\log(HT)}{2}\|x-x'\|_2\right\}\nonumber\\
&\le \sum_{h:\gamma_h\ge\frac{1}{HT^4}}^H\gamma_h\mathbb{P}_h\left\{\exists x'\in\mathbb{R}^d, \sqrt{\overline{\alpha}_t}\sum_{i=1}^H \|(\pi_i(x)-\pi_i(x'))(\mu_i-\mu_h)\|_2 >\frac{C\log(HT)}{2}\|x-x'\|_2\right\}\notag\\
&\quad + \frac{1}{T^4},
\end{align}
where $\mathbb{P}_h\{\cdot\}$ denotes the probability of the event when $x$ follows the Gaussian distribution $\mathcal{N}(\sqrt{\overline{\alpha}_t}\mu_h,\sigma_t^2I_d)$.

Define the set
\begin{align*}
\mathcal{T}_t^h = \left\{x\in\mathbb{R}^d:\sqrt{\overline{\alpha}_t}|(x-\sqrt{\overline{\alpha}_t}\mu_h)^{\top}(\mu_i-\mu_h)|\le C_2\sigma_t\sqrt{\overline{\alpha}_t\log(HT)}\|\mu_i-\mu_h\|,\forall i\right\}.
\end{align*}
According to the concentration inequality of Gaussian distribution, it is easy to check that for sufficiently large $C_2$, we have
\begin{align*}
\mathbb{P}\left\{x\sim\mathcal{N}(\sqrt{\overline{\alpha}_t}\mu_h,\sigma_t^2I_d), x\notin\mathcal{T}_t^h\right\}\lesssim \frac{1}{HT^4},\qquad \forall h.
\end{align*}
Below we shall prove that for $h$ satisfying $\gamma_h\ge \frac{1}{HT^4}$,
\begin{align}\label{eq:proof-examples-temp-2}
\begin{small}
\mathcal{T}_t^h\subset \left\{x\in\mathbb{R}^d:\forall x'\in\mathbb{R}^d, \sqrt{\overline{\alpha}_t}\sum_{i=1}^H \|(\pi_i(x)-\pi_i(x'))(\mu_i-\mu_h)\|_2 \le\frac{C\log(HT)}{2}\|x-x'\|_2\right\}.
\end{small}
\end{align}
Inserting it into \eqref{eq:proof-examples-temp-1}, we have
\begin{align*}
&\quad\mathbb{P}\left\{\exists x'\in\mathbb{R}^d, (1-\overline{\alpha}_t)\|s_t^{\star}(x)-s_t^{\star}(x')\|_2 >C\log(HT)\|x-x'\|_2\right\}\nonumber\\
&\le \sum_{h=1}^H\gamma_h \mathbb{P}_h\left\{x\notin\mathcal{T}_t^h\right\}\mathds{1}\left(\gamma_h\ge \frac{1}{HT^4}\right) + \frac{1}{T^4}\lesssim \frac{1}{T^4},
%\nonumber\\
%&\le \sum_{h=1}^H\pi_h P\left\{J_t(X_t^{(h)}\notin\mathcal{T}_t^h\right\}+ \frac{1}{T^4}\lesssim \frac{1}{T^4},
\end{align*}
and complete the proof.

\noindent \textbf{Proof of \eqref{eq:proof-examples-temp-2}.}
To this end, we introduce an auxiliary set
\begin{align*}
\mathcal{F}_h = \left\{i: \|x-x'\|_2\le c\sqrt{\overline{\alpha}_t}\|\mu_i - \mu_h\|_2\right\},
\end{align*}
where $c$ is a sufficiently small constant.
Then we have
\begin{align}\label{eq:proof-example-GMM-decomp-i}
&\quad\sqrt{\overline{\alpha}_t}\sum_{i=1}^H \|(\pi_i(x)-\pi_i(x'))(\mu_i-\mu_h)\|_2\notag\\
&\le \sqrt{\overline{\alpha}_t}\sum_{i\in\mathcal{F}_h} \|(\pi_i(x)-\pi_i(x'))(\mu_i-\mu_h)\|_2 + \sqrt{\overline{\alpha}_t}\sum_{i\in\mathcal{F}_h^{\rm c}} \|(\pi_i(x)-\pi_i(x'))(\mu_i-\mu_h)\|_2\nonumber\\
&\le\sqrt{\overline{\alpha}_t}\sum_{i\in\mathcal{F}_h} \|(\pi_i(x)-\pi_i(x'))(\mu_i-\mu_h)\|_2 + \frac{2}{c}\|x-x'\|_2.
\end{align}
For $i\in\mathcal{F}_h$, we make a decomposition on the term $|\pi_i(x) - \pi_i(x')|$.
Define 
$
\widetilde{x}_i = x\mathds{1}(\|x-\sqrt{\overline{\alpha}_t}\mu_i\|_2\le\|x'-\sqrt{\overline{\alpha}_t}\mu_i\|_2) + x'\mathds{1}(\|x'-\sqrt{\overline{\alpha}_t}\mu_i\|_2\le\|x-\sqrt{\overline{\alpha}_t}\mu_i\|_2)
$ and $
\widetilde{x}'_i = x'\mathds{1}(\|x-\sqrt{\overline{\alpha}_t}\mu_i\|_2\le\|x'-\sqrt{\overline{\alpha}_t}\mu_i\|_2) + x\mathds{1}(\|x'-\sqrt{\overline{\alpha}_t}\mu_i\|_2\le\|x-\sqrt{\overline{\alpha}_t}\mu_i\|_2)
$
%If $\|x-\sqrt{\overline{\alpha}_t}\mu_i\|_2\le\|x'-\sqrt{\overline{\alpha}_t}\mu_i\|_2$, then according to the definition of $\pi_i(x)$, we have
\begin{small}
\begin{align}\label{eq:proof-example-GMM-diffpi}
|\pi_i(x) - \pi_i(x')|
&\le \pi_i(\widetilde{x}_i)\left|1-\exp\left(-\frac{\|\widetilde{x}'_i-\sqrt{\overline{\alpha}_t}\mu_i\|_2^2}{2\sigma_t^2}+\frac{\|\widetilde{x}_i-\sqrt{\overline{\alpha}_t}\mu_i\|_2^2}{2\sigma_t^2}\right)\right|\nonumber\\
&\quad + \sum_{\ell=1}^H\frac{\gamma_\ell\gamma_i\exp\left(-\frac{\|\widetilde{x}'_i-\sqrt{\overline{\alpha}_t}\mu_i\|_2^2}{2\sigma_t^2}\right)\left|\exp\left(-\frac{\|\widetilde{x}'_i-\sqrt{\overline{\alpha}_t}\mu_\ell\|_2^2}{2\sigma_t^2}\right)-\exp\left(-\frac{\|\widetilde{x}_i-\sqrt{\overline{\alpha}_t}\mu_\ell\|_2^2}{2\sigma_t^2}\right)\right|}{\left(\sum_{\ell=1}^H\gamma_\ell\exp\left(-\frac{\|\widetilde{x}_i-\sqrt{\overline{\alpha}_t}\mu_\ell\|_2^2}{2\sigma_t^2}\right)\right)\left(\sum_{\ell=1}^H\gamma_\ell\exp\left(-\frac{\|\widetilde{x}'_i-\sqrt{\overline{\alpha}_t}\mu_\ell\|_2^2}{2\sigma_t^2}\right)\right)}.
\end{align}
\end{small}
For the first term, we have
\begin{align}
&\quad \pi_i(\widetilde{x}_i)\left|1-\exp\left(-\frac{\|\widetilde{x}'_i-\sqrt{\overline{\alpha}_t}\mu_i\|_2^2}{2\sigma_t^2}+\frac{\|\widetilde{x}_i-\sqrt{\overline{\alpha}_t}\mu_i\|_2^2}{2\sigma_t^2}\right)\right|\nonumber\\
&\le \frac{\pi_i(\widetilde{x}_i)}{\sigma_t^2}\left(|(x-\sqrt{\overline{\alpha}_t}\mu_h)^{\top}(\widetilde{x}'_i-\widetilde{x}_i)| + \sqrt{\overline{\alpha}_t}|(\widetilde{x}'_i-\widetilde{x}_i)^{\top}(\mu_h-\mu_i)| + \frac{\|x-x'\|_2^2}{2}\right)\label{eq:proof-example-GMM-diffprob}\\
&\overset{\text{(a)}}\le\frac{\max\{\pi_i(x),\pi_i(x')\}}{\sigma_t^2}\left(|(x-\sqrt{\overline{\alpha}_t}\mu_h)^{\top}(x'-x)| + \left(1+\frac{c}{2}\right)\sqrt{\overline{\alpha}_t}\|x'-x\|_2\|\mu_h-\mu_i\|_2
%+ \frac{c\sqrt{\overline{\alpha}_t}\|\mu_i-\mu_h\|_2\|x-x'\|_2}{2}
\right),\label{eq:proof-example-GMM-diffpi-1}
\end{align}
where (a) uses the fact that $|(\widetilde{x}'_i-\widetilde{x}_i)^{\top}(\mu_k-\mu_i)|\le \|x'-x\|_2\|\mu_k-\mu_i\|_2$ and $\|x-x'\|_2\le c\sqrt{\overline{\alpha}_t}\|\mu_i-\mu_h\|_2$.

For the second term, we have
\begin{align}
&\quad\frac{\gamma_\ell\gamma_i\exp\left(-\frac{\|\widetilde{x}'_i-\sqrt{\overline{\alpha}_t}\mu_i\|_2^2}{2\sigma_t^2}\right)\left|\exp\left(-\frac{\|\widetilde{x}'_\ell-\sqrt{\overline{\alpha}_t}\mu_\ell\|_2^2}{2\sigma_t^2}\right)-\exp\left(-\frac{\|\widetilde{x}_\ell-\sqrt{\overline{\alpha}_t}\mu_\ell\|_2^2}{2\sigma_t^2}\right)\right|}{\left(\sum_{\ell=1}^H\gamma_\ell\exp\left(-\frac{\|\widetilde{x}_i-\sqrt{\overline{\alpha}_t}\mu_\ell\|_2^2}{2\sigma_t^2}\right)\right)\left(\sum_{\ell=1}^H\gamma_\ell\exp\left(-\frac{\|\widetilde{x}'_i-\sqrt{\overline{\alpha}_t}\mu_\ell\|_2^2}{2\sigma_t^2}\right)\right)}\nonumber\\
&\le\frac{\gamma_i\exp\left(-\frac{\|\widetilde{x}'_i-\sqrt{\overline{\alpha}_t}\mu_i\|_2^2}{2\sigma_t^2}\right)}{\min\{s(x),s(x')\}}
\frac{\gamma_\ell\left|\exp\left(-\frac{\|\widetilde{x}'_\ell-\sqrt{\overline{\alpha}_t}\mu_\ell\|_2^2}{2\sigma_t^2}\right)-\exp\left(-\frac{\|\widetilde{x}_\ell-\sqrt{\overline{\alpha}_t}\mu_\ell\|_2^2}{2\sigma_t^2}\right)\right|}{\max\{s(x),s(x')\}}\nonumber\\
&\overset{\text{(a)}}{\le} \max\{\pi_i(x),\pi_i(x')\}\frac{\max\{\pi_\ell(x),\pi_\ell(x')\}}{\sigma_t^2}\Bigg(|(x-\sqrt{\overline{\alpha}_t}\mu_h)^{\top}(x'-x)|\nonumber\\
&\qquad\qquad\qquad\qquad + \sqrt{\overline{\alpha}_t}\|x'-x\|_2\|\mu_k-\mu_l\|_2 + \frac{c\sqrt{\overline{\alpha}_t}\|\mu_i-\mu_h\|\|x-x'\|_2}{2}\Bigg),\label{eq:proof-example-GMM-diffpi-2}
\end{align}
where
\begin{align*}
s(x): = \sum_{\ell=1}^H\gamma_\ell\exp\left(-\frac{\|x-\sqrt{\overline{\alpha}_t}\mu_\ell\|_2^2}{2\sigma_t^2}\right),
\end{align*}
and (a) comes from the fact that
\begin{align*}
\frac{\gamma_i\exp\left(-\frac{\|\widetilde{x}'_i-\sqrt{\overline{\alpha}_t}\mu_i\|_2^2}{2\sigma_t^2}\right)}{\min\{s(x),s(x')\}}
&\le \max\left\{\frac{\gamma_i\exp\left(-\frac{\|\widetilde{x}'_i-\sqrt{\overline{\alpha}_t}\mu_i\|_2^2}{2\sigma_t^2}\right)}{s(\widetilde{x}_i)},\frac{\gamma_i\exp\left(-\frac{\|\widetilde{x}'_i-\sqrt{\overline{\alpha}_t}\mu_i\|_2^2}{2\sigma_t^2}\right)}{s(\widetilde{x}'_i)}\right\}\nonumber\\
&\le \max\left\{\frac{\gamma_i\exp\left(-\frac{\|\widetilde{x}_i-\sqrt{\overline{\alpha}_t}\mu_i\|_2^2}{2\sigma_t^2}\right)}{s(\widetilde{x}_i)},\frac{\gamma_i\exp\left(-\frac{\|\widetilde{x}'_i-\sqrt{\overline{\alpha}_t}\mu_i\|_2^2}{2\sigma_t^2}\right)}{s(\widetilde{x}'_i)}\right\}\notag\\
&= \max\left\{\pi_i(x),\pi_i(x')\right\},
\end{align*}
and the following inequality holds according to \eqref{eq:proof-example-GMM-diffprob}:
\begin{align*}
&\quad\frac{\gamma_\ell\left|\exp\left(-\frac{\|\widetilde{x}'_\ell-\sqrt{\overline{\alpha}_t}\mu_\ell\|_2^2}{2\sigma_t^2}\right)-\exp\left(-\frac{\|\widetilde{x}_\ell-\sqrt{\overline{\alpha}_t}\mu_\ell\|_2^2}{2\sigma_t^2}\right)\right|}{\max\{s(x),s(x')\}}\nonumber\\
&\le\frac{\max\{\pi_\ell(x),\pi_\ell(x')\}}{\sigma_t^2}\left(|(x-\sqrt{\overline{\alpha}_t}\mu_h)^{\top}(x'-x)| + \sqrt{\overline{\alpha}_t}|(x'-x)^{\top}(\mu_h-\mu_\ell)| + \frac{\|x-x'\|_2^2}{2}\right)\nonumber\\
&\le\frac{\max\{\pi_\ell(x),\pi_\ell(x')\}}{\sigma_t^2}\Big(|(x-\sqrt{\overline{\alpha}_t}\mu_h)^{\top}(x'-x)| + \sqrt{\overline{\alpha}_t}\|x'-x\|_2\|\mu_h-\mu_\ell\|_2\notag\\
&\quad + \frac{c\sqrt{\overline{\alpha}_t}\|\mu_i-\mu_h\|\|x-x'\|_2}{2}\Big).
\end{align*}

Inserting \eqref{eq:proof-example-GMM-diffpi-1} and \eqref{eq:proof-example-GMM-diffpi-2} into \eqref{eq:proof-example-GMM-diffpi}, for $c\le 2$, we have
\begin{align}\label{eq:proof-example-decomp-i}
\sqrt{\overline{\alpha}_t}|\pi_i(x) - \pi_i(x')|\|\mu_i-\mu_h\|_2
&\le \epsilon_{i,1} + \sum_{\ell=1}^H \epsilon_{i,2}^{(\ell)},
\end{align}
where
\begin{align*}
\epsilon_{i,1}&:=\frac{\max\{\pi_i(x),\pi_i(x')\}}{\sigma_t^2}\left(\sqrt{\overline{\alpha}_t}|(x-\sqrt{\overline{\alpha}_t}\mu_h)^{\top}(\mu_i-\mu_h)| + 2\overline{\alpha}_t\|\mu_h-\mu_i\|_2^2\right)\|x'-x\|_2,\nonumber\\
\epsilon_{i,2}^{(\ell)}&:=\max\{\pi_i(x),\pi_i(x')\}\frac{\max\{\pi_\ell(x),\pi_\ell(x')\}}{\sigma_t^2}\Bigg(\sqrt{\overline{\alpha}_t}|(x-\sqrt{\overline{\alpha}_t}\mu_h)^{\top}(\mu_i-\mu_h)|\nonumber\\
&\qquad\qquad\qquad\qquad + \overline{\alpha}_t\|\mu_h-\mu_\ell\|_2\|\mu_h-\mu_i\|_2 + \overline{\alpha}_t\|\mu_i-\mu_h\|_2^2\Bigg)\|x'-x\|_2.
\end{align*}
We use the following lemma to continue the proof.
\begin{lemma}\label{lem:example-GMM-bounds}
For any $i$ and $h$,
suppose that $x\in\mathcal{T}_t^h$, $\|x-x'\|_2\le c\sqrt{\overline{\alpha}_t}\|\mu_i-\mu_h\|_2$, and $\gamma_h \ge \frac{1}{HT^4}$, where $c$ is a sufficiently small constant.
Then we have
\begin{align*}
\frac{\overline{\alpha}_t}{\sigma_t^2}\max\{\pi_i(x),\pi_i(x')\}\|\mu_i-\mu_h\|_2^2
&\lesssim \max\{\gamma_i,\pi_i(x),\pi_i(x')\}\log(HT),\nonumber\\
\frac{\sqrt{\overline{\alpha}_t}}{\sigma_t^2}\max\{\pi_i(x),\pi_i(x')\}|(x-\sqrt{\overline{\alpha}_t}\mu_h)^{\top}(\mu_i-\mu_h)|
&\lesssim \max\{\gamma_i,\pi_i(x),\pi_i(x')\}\log(HT).
\end{align*}
\end{lemma}

By using Lemma \ref{lem:example-GMM-bounds}, we have
\begin{align*}
\epsilon_{i,1}
&\lesssim \max\{\gamma_i,\pi_i(x),\pi_i(x')\}\log(HT)\|x-x'\|_2,\nonumber\\
\epsilon_{i,2}^{(\ell)}
&\lesssim \max\{\pi_\ell(x),\pi_\ell(x')\}\max\{\gamma_i,\pi_i(x),\pi_i(x')\}\log(HT)\|x-x'\|_2\nonumber\\
&\quad + \max\{\pi_\ell(x),\pi_\ell(x')\}\max\{\pi_i(x),\pi_i(x')\}\frac{\overline{\alpha}_t}{\sigma_t^2}\|\mu_h-\mu_\ell\|_2\|\mu_h-\mu_i\|_2\|x-x'\|_2.
\end{align*}
If $\|x-x'\|_2\le c\sqrt{\overline{\alpha}_t}\|\mu_\ell-\mu_h\|_2$, then we have
\begin{align*}
&\quad\max\{\pi_\ell(x),\pi_\ell(x')\}\max\{\pi_i(x),\pi_i(x')\}\frac{\overline{\alpha}_t}{\sigma_t^2}\|\mu_h-\mu_\ell\|_2\|\mu_h-\mu_i\|_2\|x-x'\|_2\notag\\
&\le \max\{\pi_\ell(x),\pi_\ell(x')\}\max\{\pi_i(x),\pi_i(x')\}\frac{\overline{\alpha}_t}{\sigma_t^2}\max\{\|\mu_h-\mu_\ell\|_2^2,\|\mu_h-\mu_i\|_2^2\}\|x-x'\|_2\notag\\
&\le \max\{\gamma_\ell,\pi_\ell(x),\pi_\ell(x')\}\max\{\gamma_i,\pi_i(x),\pi_i(x')\}\log(HT)\|x-x'\|_2.
\end{align*}
Otherwise, that is $\|x-x'\|_2> c\sqrt{\overline{\alpha}_t}\|\mu_\ell-\mu_h\|_2$, then we have
\begin{align*}
&\quad\max\{\pi_\ell(x),\pi_\ell(x')\}\max\{\pi_i(x),\pi_i(x')\}\frac{\overline{\alpha}_t}{\sigma_t^2}\|\mu_h-\mu_\ell\|_2\|\mu_h-\mu_i\|_2\|x-x'\|_2\notag\\
&\le\max\{\pi_\ell(x),\pi_\ell(x')\}\max\{\pi_i(x),\pi_i(x')\}\frac{\overline{\alpha}_t}{\sigma_t^2}\frac{\|x-x'\|_2}{c\sqrt{\overline{\alpha}_t}}\|\mu_h-\mu_i\|_2 c\sqrt{\overline{\alpha}_t}\|\mu_i-\mu_h\|_2\notag\\
&=\max\{\pi_\ell(x),\pi_\ell(x')\}\max\{\pi_i(x),\pi_i(x')\}\frac{\overline{\alpha}_t}{\sigma_t^2}\|x-x'\|_2\|\mu_h-\mu_i\|_2^2\notag\\
&\le \max\{\pi_\ell(x),\pi_\ell(x')\}\max\{\gamma_i,\pi_i(x),\pi_i(x')\}\log(HT)\|x-x'\|_2.
\end{align*}
Thus we have
\begin{align*}
\epsilon_{i,2}^{(\ell)}
&\overset{\text{}}{\lesssim} \max\{\gamma_\ell,\pi_\ell(x),\pi_\ell(x')\}\max\{\gamma_i,\pi_i(x),\pi_i(x')\}\log(HT)\|x-x'\|_2.
\end{align*}
Inserting into \eqref{eq:proof-example-decomp-i}, we have
\begin{align}
&\quad \sqrt{\overline{\alpha}_t}|\pi_i(x) - \pi_i(x')|\|\mu_i-\mu_h\|_2\notag\\
&\lesssim \max\{\gamma_i,\pi_i(x),\pi_i(x')\}\log(HT)\|x-x'\|_2\left(1 + \sum_{\ell=1}^H\max\{\gamma_\ell,\pi_\ell(x),\pi_\ell(x')\}\right)\nonumber\\
&\lesssim  \max\{\gamma_i,\pi_i(x),\pi_i(x')\}\log(HT)\|x-x'\|_2.
\end{align}
Inserting it into \eqref{eq:proof-example-GMM-decomp-i}
we have
\begin{align*}
&\quad \sqrt{\overline{\alpha}_t}\sum_{i=1}^H \|(\pi_i(x)-\pi_i(x'))(\mu_i-\mu_h)\|_2\notag\\
&\lesssim \sum_{i\in\mathcal{F}_h}\max\{\gamma_i,\pi_i(x),\pi_i(x')\}\log(HT)\|x-x'\|_2 + \frac{2}{c}\|x-x'\|_2\notag\\
&\lesssim \log(HT)\|x-x'\|_2,
\end{align*}
and complete the proof.

\noindent \textbf{Proof of Lemma \ref{lem:example-GMM-bounds}.}
We first bound $\pi_i(x)$ and $\pi_i(x')$.
% According to \eqref{eq:proof-examples-temp-5}, we first make the following decomposition
% \begin{align}\label{eq:proof-examples-temp-6}
% \mathsf{Tr}\left(J_t(X_t^{(h)})+\frac{1-\overline{\alpha}_t}{\sigma_t^2}I_d\right) = \frac{\overline{\alpha}_t}{\sigma_t^2}\left(\sum_{h=1}^H\gamma_h\mu_h\mu_h^{\top} - \overline{\mu}\overline{\mu}^{\top}\right) 
% &= \frac{\overline{\alpha}_t}{\sigma_t^2}\sum_{i=1}^H\gamma_i\left\|\mu_i - \overline{\mu}\right\|^2\nonumber\\
% &\le \frac{\overline{\alpha}_t}{\sigma_t^2}\sum_{i=1}^H\gamma_i\left\|\mu_i - \mu_h\right\|^2.
% \end{align}
Noticing that for any $h$ such that $\gamma_h\ge \frac{1}{HT^4}$, any $i$, and any $x\in\mathcal{T}_t^h$, we have
\begin{align*}
\pi_i(x)
&\le \frac{\pi_i(x)}{\pi_h(x)} = \frac{\gamma_i}{\gamma_h}\exp\left(-\frac{1}{2\sigma_t^2}\|x-\sqrt{\overline{\alpha}_t}\mu_i\|^2 + \frac{1}{2\sigma_t^2}\|x-\sqrt{\overline{\alpha}_t}\mu_h\|^2\right)\nonumber\\
&=\frac{\gamma_i}{\gamma_h}\exp\left(-\frac{\overline{\alpha}_t\|\mu_i-\mu_h\|_2^2}{2\sigma_t^2} + \frac{\sqrt{\overline{\alpha}_t}}{\sigma_t^2}(x-\sqrt{\overline{\alpha}_t}\mu_h)^{\top}(\mu_i-\mu_h)\right)\nonumber\\
&\overset{\text{(a)}}{\le}
\gamma_i\exp\left(-\frac{\overline{\alpha}_t\|\mu_i-\mu_h\|_2^2}{2\sigma_t^2} + \frac{\sqrt{\overline{\alpha}_t}}{\sigma_t^2}|(x-\sqrt{\overline{\alpha}_t}\mu_h)^{\top}(\mu_i-\mu_h)| + 4\log(HT)\right),
%&\overset{\text{(a)}}{\le}
%\gamma_i\exp\left(-\frac{\overline{\alpha}_t}{2\sigma_t^2} + \frac{C_5}{\sigma_t}\sqrt{\overline{\alpha}_t\log(HT)}\|\mu_i-\mu_h\| + 4\log(KT)\right),
\end{align*}
where (a) uses the fact that $\gamma_h\ge \frac{1}{HT^4}$.
Similarly, we bound the $\pi_i(x')$ as 
\begin{align*}
\pi_i(x')
&\le \frac{\pi_i(x')}{\pi_h(x')} =\frac{\gamma_i}{\gamma_h}\exp\Big(-\frac{\overline{\alpha}_t\|\mu_i-\mu_h\|_2^2}{2\sigma_t^2} + \frac{\sqrt{\overline{\alpha}_t}}{\sigma_t^2}(x-\sqrt{\overline{\alpha}_t}\mu_h)^{\top}(\mu_i-\mu_h)\notag\\
&\qquad\qquad\qquad\qquad\qquad +\frac{\sqrt{\overline{\alpha}_t}}{\sigma_t^2}(x'-x)^{\top}(\mu_i-\mu_h)\Big)\nonumber\\
&\overset{\text{(a)}}{\le}
\gamma_i\exp\left(-\frac{11\overline{\alpha}_t\|\mu_i-\mu_h\|_2^2}{24\sigma_t^2} + \frac{\sqrt{\overline{\alpha}_t}}{\sigma_t^2}|(x-\sqrt{\overline{\alpha}_t}\mu_h)^{\top}(\mu_i-\mu_h)| + 4\log(HT)\right),
%&\overset{\text{(a)}}{\le}
%\gamma_i\exp\left(-\frac{\overline{\alpha}_t}{2\sigma_t^2} + \frac{C_5}{\sigma_t}\sqrt{\overline{\alpha}_t\log(HT)}\|\mu_i-\mu_h\| + 4\log(KT)\right),
\end{align*}
where (a) uses the fact that for $c\le 1/24$,
$$
\frac{\sqrt{\overline{\alpha}_t}}{\sigma_t^2}(x'-x)^{\top}(\mu_i-\mu_h)\le \frac{\sqrt{\overline{\alpha}_t}}{\sigma_t^2}\|x'-x\|_2\|\mu_i-\mu_h\|_2 \le \frac{c\overline{\alpha}_t}{\sigma_t^2}\|\mu_i-\mu_h\|_2^2\le \frac{\overline{\alpha}_t}{24\sigma_t^2}\|\mu_i-\mu_h\|_2^2.
$$
Let us discuss two cases:
\begin{itemize}
\item If $\sqrt{\overline{\alpha}_t}\|\mu_i-\mu_h\|_2\ge 6\sigma_tC_2\sqrt{\log(HT)}$, then
\begin{align*}
\max\{\pi_i(x),\pi_i(x')\} 
&\le \gamma_i\exp\Big(-\frac{11\overline{\alpha}_t\|\mu_i-\mu_h\|_2^2}{24\sigma_t^2} + \frac{\sqrt{\overline{\alpha}_t}}{\sigma_t^2}|(x-\sqrt{\overline{\alpha}_t}\mu_h)^{\top}(\mu_i-\mu_h)|\notag\\
&\qquad\qquad\quad+ 4\log(HT)\Big)\nonumber\\
&\overset{\text{(a)}}{\le} \gamma_i\exp\left(-\frac{5\overline{\alpha}_t\|\mu_i-\mu_h\|_2^2}{12\sigma_t^2} + \frac{\sqrt{\overline{\alpha}_t}}{\sigma_t^2}|(x-\sqrt{\overline{\alpha}_t}\mu_h)^{\top}(\mu_i-\mu_h)|\right),
\end{align*}
where (a) holds as long as $C_2\ge\sqrt{8/3}$ and 
$$
4\log(HT)\le \frac{\overline{\alpha}_t\|\mu_i-\mu_h\|_2^2}{9C_2^2\sigma_t^2}\le \frac{\overline{\alpha}_t\|\mu_i-\mu_h\|_2^2}{24\sigma_t^2}.
$$
Furthermore, we have
\begin{align*}
&\quad\frac{\overline{\alpha}_t}{\sigma_t^2}\max\{\pi_i(x),\pi_i(x')\}\|\mu_i-\mu_h\|_2^2\notag\\
&\le 24\gamma_i\exp\left(-\frac{5\overline{\alpha}_t\|\mu_i-\mu_h\|_2^2}{12\sigma_t^2} + \frac{\sqrt{\overline{\alpha}_t}}{\sigma_t^2}|(x-\sqrt{\overline{\alpha}_t}\mu_h)^{\top}(\mu_i-\mu_h)|+\frac{\overline{\alpha}_t\|\mu_i-\mu_h\|_2^2}{24\sigma_t^2}\right)\nonumber\\
&\lesssim \gamma_i\exp\left(-\frac{3\overline{\alpha}_t\|\mu_i-\mu_h\|_2^2}{8\sigma_t^2} + \frac{\sqrt{\overline{\alpha}_t}}{\sigma_t^2}|(x-\sqrt{\overline{\alpha}_t}\mu_h)^{\top}(\mu_i-\mu_h)|\right)
% \nonumber\\
% &\lesssim \gamma_i\exp\left(-\frac{\overline{\alpha}_t\|\mu_i-\mu_h\|_2^2}{8\sigma_t^2} + \frac{\sqrt{\overline{\alpha}_t}}{\sigma_t^2}|(x-\sqrt{\overline{\alpha}_t}\mu_h)^{\top}(\mu_i-\mu_h)|\right)
\end{align*}
and 
\begin{align*}
&\quad\frac{\sqrt{\overline{\alpha}_t}}{\sigma_t^2}\max\{\pi_i(x),\pi_i(x')\}|(x-\sqrt{\overline{\alpha}_t}\mu_h)^{\top}(\mu_i-\mu_h)|\notag\\
&\le \gamma_i\exp\left(-\frac{5\overline{\alpha}_t\|\mu_i-\mu_h\|_2^2}{12\sigma_t^2} + \frac{2\sqrt{\overline{\alpha}_t}}{\sigma_t^2}|(x-\sqrt{\overline{\alpha}_t}\mu_h)^{\top}(\mu_i-\mu_h)|\right)
\end{align*}
Thus to establish Lemma \ref{lem:example-GMM-bounds}, it suffices to prove that
\begin{align*}
\exp\left(-\frac{3\overline{\alpha}_t\|\mu_i-\mu_h\|_2^2}{8\sigma_t^2} + \frac{2\sqrt{\overline{\alpha}_t}}{\sigma_t^2}|(x-\sqrt{\overline{\alpha}_t}\mu_h)^{\top}(\mu_i-\mu_h)|\right)\le 1.
\end{align*}
Recalling that $x\in\mathcal{T}_t^h$, which implies that $\sqrt{\overline{\alpha}_t}|(x-\sqrt{\overline{\alpha}_t}\mu_h)^{\top}(\mu_i-\mu_h)|\le C_2\sigma_t\sqrt{\overline{\alpha}_t\log(HT)}\|\mu_i-\mu_h\|_2$, we have
\begin{align*}
\frac{2\sqrt{\overline{\alpha}_t}}{\sigma_t^2}|(x-\sqrt{\overline{\alpha}_t}\mu_h)^{\top}(\mu_i-\mu_h)|\le \frac{2C_2}{\sigma_t}\sqrt{\overline{\alpha}_t\log(HT)}\|\mu_i-\mu_h\|_2\le \frac{\overline{\alpha}_t}{3\sigma_t^2}\|\mu_i-\mu_h\|_2^2,
\end{align*}
and complete the proof.
\item If $\sqrt{\overline{\alpha}_t}\|\mu_i-\mu_h\|< 6C_2\sigma_t\sqrt{\log(HT)}$, then we have
\begin{align}\label{eq:proof-example-temp-4}
&\quad\frac{\overline{\alpha}_t}{\sigma_t^2}\max\{\pi_i(x),\pi_i(x)\}\|\mu_i-\mu_h\|_2^2
\le 36C_2^2\max\{\pi_i(x),\pi_i(x)\}\log(HT),\nonumber\\
&\quad \frac{\sqrt{\overline{\alpha}_t}}{\sigma_t^2}\max\{\pi_i(x),\pi_i(x')\}|(x-\sqrt{\overline{\alpha}_t}\mu_h)^{\top}(\mu_i-\mu_h)|\notag\\
&\le \frac{C_2\sqrt{\overline{\alpha}_t\log(HT)}}{\sigma_t}\max\{\pi_i(x),\pi_i(x')\}\|\mu_i-\mu_h\|_2\nonumber\\
&\le 6C_2^2\log(HT)\max\{\pi_i(x),\pi_i(x')\}.
\end{align}
% Inserting \eqref{eq:proof-example-temp-3} and \eqref{eq:proof-example-temp-4} into \eqref{eq:proof-examples-temp-6}, we have
% \begin{align*}
% \mathsf{Tr}\left(J_t(X_t^{(h)})+\frac{1-\overline{\alpha}_t}{\sigma_t^2}I_d\right) \le 
% \sum_{i=1}^H 12\pi_i\exp\left(-3C_5^2\log(HT)\right) + \sum_{i=1}^H 36C_5^2\gamma_i\log(HT)
% \lesssim \log(HT),
% \end{align*}
% and we complete the proof.

\end{itemize}

% \begin{align*}
% \gamma_i\le \pi_i\exp\left(-\frac{\overline{\alpha}_t}{6\sigma_t^2}\|\mu_h-\mu_i\|^2\right),
% \end{align*}
% as long as $C_5\ge \sqrt{2/3}$.
% Then we have
% \begin{align}\label{eq:proof-example-temp-3}
% \frac{\gamma_i\overline{\alpha}_t}{\sigma_t^2}\|\mu_i-\mu_h\|^2
% &\le \frac{\pi_i\overline{\alpha}_t}{\sigma_t^2}\|\mu_i-\mu_h\|^2\exp\left(-\frac{\overline{\alpha}_t}{6\sigma_t^2}\|\mu_h-\mu_i\|^2\right)\le 12\pi_i\exp\left(-\frac{\overline{\alpha}_t}{12\sigma_t^2}\|\mu_h-\mu_i\|^2\right) \nonumber\\
% &\overset{\text{(a)}}{=} 12\pi_i\exp\left(-3C_5^2\log(HT)\right),
% \end{align}
% where (a) uses the condition $\sqrt{\overline{\alpha}_t}\|\mu_i-\mu_h\|\ge 6C_5\sigma_t\sqrt{\log(HT)}$.

\subsubsection{Proof of the lower bound}

By calculation, we have
\begin{align}\label{eq:def-J-GMM}
J_t(x) = \frac{1}{\sigma_t^2}\left[-I_d + \frac{\overline{\alpha}_t}{\sigma_t^2}\left(\sum_{h=1}^H\gamma_h\mu_h\mu_h^{\top} - \overline{\mu}\overline{\mu}^{\top}\right)\right],\qquad {\rm where}\qquad \overline{\mu} = \sum_{h=1}^H\gamma_h\mu_h.
\end{align}
In the case of $X_0\sim\frac12\mathcal{N}(\mu,\sigma^2I_d) + \frac12\mathcal{N}(-\mu,\sigma^2I_d)$, we have
\begin{align*}
(1-\overline{\alpha}_t)\nabla s_t^{\star}(x) = (1-\overline{\alpha}_t)J_t(x) = \frac{(1-\overline{\alpha}_t)}{\sigma_t^2}\left[-I_d + \frac{\overline{\alpha}_t}{\sigma_t^2}\mu\mu^{\top}\right].
\end{align*}
Thus we have
\begin{align*}
\left\|(1-\overline{\alpha}_t)\nabla s_t^{\star}(x)\right\|_2 
&= \frac{(1-\overline{\alpha}_t)}{\sigma_t^2}\left\|-I_d + \frac{\overline{\alpha}_t}{\sigma_t^2}\mu\mu^{\top}\right\|_\mathsf{op} \nonumber\\
&=\frac{(1-\overline{\alpha}_t)}{\sigma_t^2}\max\left\{\frac{\overline{\alpha}_t\|\mu\|_2^2}{\sigma_t^2}-1,1\right\}
\overset{\text{(a)}}{\ge} \frac{\overline{\alpha}_t(1-\overline{\alpha}_t)\|\mu\|_2^2}{2\sigma_t^4}\nonumber\\
&\ge\frac{(1-\overline{\alpha}_t)\|\mu\|_2^2}{4(1-\overline{\alpha}_t+\sigma^2)^2},
%&=\frac{\|\mu\|_2^2}{2(\sigma^4\vartheta_t + 2\sigma^2 + \frac{1}{\vartheta_t})} \ge \frac{1}{1+2\sigma^2}\frac{1}{1-\overline{\alpha}_t+\sigma^2},
\end{align*}
where the last inequality uses the fact that $\overline{\alpha}_t\ge \frac12$.

\section{Proof of key lemmas in Theorem~\ref{thm:main}}

Before diving into the proof details, we first present a preliminary lemma about the learning schedule $\tau_{k,n}$.
It has been stated in \citet{li2024improved}.
For completeness, we rewrite its proof in Appendix~\ref{sec:proof-lem:learning} in supplemental material.
\begin{lemma} \label{lem:learning}
Our choice of learning schedules~\eqref{eq:learning-rate} satisfies
\begin{align}\label{eq:diff-tau-1}
1-\tau_{0, 0} \le \widehat{\alpha}_{T} \le \frac{2}{T^{c_0}},
\qquad
\tau_{K, 0} \le 1-\widehat{\alpha}_{1} \le \frac{1}{T^{c_0}},
\quad\text{and}\quad
\frac{\widehat{\tau}_{k,n-1} - \widehat{\tau}_{k,n}}{\widehat{\tau}_{k,n-1}(1-\widehat{\tau}_{k, n-1})} = \frac{c_1\log T}{T}.
\end{align}
Moreover, we have
\begin{align}\label{eq:diff-tau-2}
\frac{\widehat{\tau}_{k,i-1}}{\widehat{\tau}_{k,i}}\le \frac{\widehat{\tau}_{k,-1}}{\widehat{\tau}_{k,N}}\lesssim 1,\qquad \frac{1-\widehat{\tau}_{k,i}}{1-\widehat{\tau}_{k,i-1}}\le\frac{1-\widehat{\tau}_{k,N}}{1-\widehat{\tau}_{k,-1}}
\lesssim 1.
\end{align}
\end{lemma}

\subsection{Proof of Lemma~\ref{lem:discrete}}
\label{sec:proof-lem:discrete}

We present a more preliminary conclusion stated as below.
\begin{lemma}
\label{lem:discrete-pre}
For any $k$ and $n$, with probability at least $1 - T^{-100}$,
\begin{align}
&\quad \frac{\|y^{\star}_{\tau_{k, n}}(x_k) - y_{\tau_{k, n}}(x_k)\|_2^2}{1-\tau_{k,n}}
\lesssim \frac{N\log^2 T}{T^2}\sum_{i=0}^{n-1}\frac{\widehat{\tau}_{k,i}\widetilde{\varepsilon}_{k,i}^2(x_k)}{1-\widehat{\tau}_{k,i}},\label{eq:lem-discrete-pre-1}\\
&\quad \frac{\mathbb{E}_{x_k\sim\widehat{X}_k}\|x_{\tau_{k, n}}(x_k) - z_{\tau_{k, n}}^{\star}(x_k)\|_2^2}{1-\tau_{k,n}}\nonumber\\
&
\lesssim\frac{d\log^5 T}{T^3}\min\Big\{\frac{Nd\widehat{\tau}_{k,-1}\log T}{T(1-\widehat{\tau}_{k,-1})}+\int_{{\tau}_{k,n}}^{{\tau}_{k,0}} \frac{\mathbb{E}[\mathsf{Tr}(\Sigma_\tau^2(x_\tau))]}{(1-\tau)^2}\mathrm d \tau,\frac{NL^2\widehat{\tau}_{k,-1}\log T}{T(1-\widehat{\tau}_{k,-1})}\Big\},\label{eq:lem-discrete-pre-2}
% \|\xi_{k, n}(x_{\tau_{k,0}})\|_2^2 
% &\le \zeta_{k,n} + \frac{N\log^2 T}{T^2}\sum_{i = 0}^{n-1} \widehat{\tau}_{k,i}(1-\widehat{\tau}_{k,i})^{-1}\|s_{\tau_{k,i}}(y_{\tau_{k,i}})-s_{\tau_{k,i}}^{\star}(y_{\tau_{k,i}})\|^2,\nonumber\\
% \mathbb{E}_{x_{\tau_{k,0}} \sim p_{\widehat{X}_{k}}}\big[\zeta_{k,n}\big]
% &\lesssim \sum_{i = 0}^{n} \frac{\widehat{\tau}_{k,i-1}\min\{d\log T,L^2\}d\log^6 T}{T^4(1-\widehat{\tau}_{k,i-1})}.
\end{align}
where $\widetilde{\varepsilon}_{k,i}(x_k)=
\|s_{\tau_{k,i}}(y_{\tau_{k,i}}(x_k))-s_{\tau_{k,i}}^{\star}(y_{\tau_{k,i}}(x_k))\|_2$.
\end{lemma}

Then according to the definition of $\xi_{k,n}$ (cf. \eqref{eq:def-xi}) and $\varepsilon_{k,i}^2$ (cf. \eqref{eq:score-error}), we could prove Lemma \ref{lem:discrete}.
The remaining proof focuses on establishing Lemma \ref{lem:discrete-pre}.
According to definitions of $y_{\tau_{k,n}}$ and $y_{\tau_{k,n}}^{\star}$,
we have
\begin{align}\label{eq:proof-temp-4}
\frac{\|y_{\tau_{k,n}}(x_k)-y_{\tau_{k,n}}^{\star}(x_k)\|_2}{\sqrt{1-\tau_{k,n}}}
&\le\frac{\widetilde{\varepsilon}_{k,0}(x_k)({\tau}_{k,0}-\widehat{\tau}_{k,0})}{2(1-\tau_{k,0})^{3/2}} + \sum_{i=1}^{n-1}\frac{\widetilde{\varepsilon}_{k,i}(x_k)(\widehat{\tau}_{k,i-1}-\widehat{\tau}_{k,i})}{2(1-\tau_{k,i})^{3/2}}\nonumber\\
&\quad + \frac{\widetilde{\varepsilon}_{k,n-1}(x_k)(\widehat{\tau}_{k,n-1}-{\tau}_{k,n})}{2(1-\tau_{k,n-1})^{3/2}}.
\end{align}
Recalling Lemma \ref{lem:learning},
we have
\begin{align*}
\frac{{\tau}_{k,0}-\widehat{\tau}_{k,0}}{2(1-\tau_{k,0})^{3/2}}
&\le \frac{\widehat{\tau}_{k,-1}-\widehat{\tau}_{k,0}}{2(1-\tau_{k,0})^{3/2}}\le 
\frac{\widehat{\tau}_{k,-1}\log T}{2T(1-\tau_{k,0})^{1/2}}\lesssim \frac{\widehat{\tau}_{k,-1}\log T}{T(1-\widehat{\tau}_{k,-1})^{1/2}}\\
\frac{\widehat{\tau}_{k,i-1}-\widehat{\tau}_{k,i}}{2(1-\tau_{k,i})^{3/2}} &\le \frac{\widehat{\tau}_{k,i-1}\log T}{2T(1-\tau_{k,i})^{1/2}}\lesssim \frac{\widehat{\tau}_{k,i-1}\log T}{T(1-\widehat{\tau}_{k,i-1})^{1/2}},\\
\frac{\widehat{\tau}_{k,n-1}-{\tau}_{k,n}}{2(1-\tau_{k,n-1})^{3/2}}&\le \frac{\widehat{\tau}_{k,n-1}-\widehat{\tau}_{k,n}}{2(1-\tau_{k,n-1})^{3/2}}\le \frac{\widehat{\tau}_{k,n-1}\log T}{2T(1-\tau_{k,n-1})^{1/2}}\notag\\
&\lesssim \frac{\widehat{\tau}_{k,n-1}\log T}{T(1-\widehat{\tau}_{k,n-1})^{1/2}}\lesssim \frac{\widehat{\tau}_{k,n-2}\log T}{T(1-\widehat{\tau}_{k,n-2})^{1/2}}.
\end{align*}
Inserting into \eqref{eq:proof-temp-4}, we have that
\begin{align*}
\frac{\|y_{\tau_{k,n}}(x_k)-y_{\tau_{k,n}}^{\star}(x_k)\|_2}{\sqrt{1-\tau_{k,n}}} 
&\lesssim \sum_{i=0}^{n-1}\frac{\widetilde{\varepsilon}_{k,i}(x_k)\widehat{\tau}_{k,i-1}\log T}{T(1-\widehat{\tau}_{k,i-1})^{1/2}}.
\end{align*}
Thus we have
\begin{align*}
\frac{\|y_{\tau_{k,n}}^{\star}(x_k) - y_{\tau_{k,n}}(x_k)\|_2^2}{1-\tau_{k,n}}\lesssim \frac{N\log^2 T}{T^2}\sum_{i=0}^{n-1}\frac{\widetilde{\varepsilon}_{k,i}^2(x_k)\widehat{\tau}_{k,i-1}}{1-\widehat{\tau}_{k,i-1}}\lesssim \frac{N\log^2 T}{T^2}\sum_{i=0}^{n-1}\frac{\widetilde{\varepsilon}_{k,i}^2(x_k)\widehat{\tau}_{k,i}}{1-\widehat{\tau}_{k,i}}
\end{align*}
and complete the proof of \eqref{eq:lem-discrete-pre-1}, where the last inequality uses \eqref{eq:diff-tau-2}.

Now we are ready to prove \eqref{eq:lem-discrete-pre-2}.
The definition of $x_{{\tau}_{k,n}}$(cf. \eqref{eq:def-xtau})  can be decomposed as
\begin{align*}
\frac{x_{\tau_{k, n}}}{\sqrt{1-\tau_{k, n}}} 
&= \frac{x_{\tau_{k, 0}}}{\sqrt{1-\tau_{k, 0}}} + \int_{\widehat{\tau}_{k,0}}^{\tau_{k,0}} \frac{s_\tau^{\star}(x_\tau)}{2(1-\tau)^{3/2}} \mathrm d \tau + \sum_{i=1}^{n-1}\int_{\widehat{\tau}_{k,i}}^{\widehat{\tau}_{k,i-1}} \frac{s_\tau^{\star}(x_\tau)}{2(1-\tau)^{3/2}} \mathrm d \tau\nonumber\\
&\quad + \int_{{\tau}_{k,n}}^{\widehat{\tau}_{k,n-1}} \frac{s_\tau^{\star}(x_\tau)}{2(1-\tau)^{3/2}} \mathrm d \tau.
\end{align*}
Based on the definitions of $x_{{\tau}_{k,n}}$ and $z_{\tau_{k,n}}^{\star}$, we first make the following decomposition as 
\begin{align*}
\frac{x_{\tau_{k,n}}(x_k) - z_{\tau_{k,n}}^{\star}(x_k)}{\sqrt{1-\tau_{k,n}}}
&= \mathcal{E}_{1, 1}(x_{\tau_{k,0}}) + \mathcal{E}_{1, 0}(x_{\tau_{k,0}}) - \mathcal{E}_2(x_{\tau_{k,0}}), 
\end{align*}
where
\begin{align*}
\mathcal{E}_{1, 1}(x_{\tau_{k,0}}) &:= \int_{{\tau}_{k,n}}^{\widehat{\tau}_{k,n-1}} \frac{s_\tau^{\star}(x_\tau)}{2(1-\tau)^{3/2}} \mathrm d \tau
- \frac{s_{\tau_{k,n-1}}^{\star}(x_{\tau_{k, n-1}})}{2(1-\tau_{k,n-1})^{3/2}}(\widehat{\tau}_{k,n-1} - \tau_{k, n}); \\
\mathcal{E}_{1, 0}(x_{\tau_{k,0}}) &:= \int_{\widehat{\tau}_{k,0}}^{\tau_{k,0}} \frac{s_\tau^{\star}(x_\tau)}{2(1-\tau)^{3/2}} \mathrm d \tau
- \frac{s_{\tau_{k,0}}^{\star}(x_{\tau_{k, 0}})}{2(1-\tau_{k,0})^{3/2}}(\tau_{k, 0} - \widehat{\tau}_{k,0}); \\
\mathcal{E}_2(x_{\tau_{k,0}}) &:= \sum_{i = 1}^{n-1} \bigg[\frac{s_{\tau_{k,i}}^{\star}(x_{\tau_{k, i}})}{2(1-\tau_{k,i})^{3/2}}(\widehat{\tau}_{k,i-1} - \widehat{\tau}_{k,i}) - \int_{\widehat{\tau}_{k,i}}^{\widehat{\tau}_{k,i-1}}\frac{s_{\tau}^{\star}(x_{\tau})}{2(1-\tau)^{3/2}} \mathrm{d} \tau\bigg].
\end{align*}

In the following, we will control these three terms separately, and then combine them.
The remaining proof is divided into four steps.
%\begin{itemize}

\emph{1. Analysis of $\mathcal{E}_{1,1}(x_{\tau_{k,0}})$.}
This term can be calculated as
\begin{align*}
\mathcal{E}_{1,1}(x_{\tau_{k,0}}) &= \int_{\tau_{k, n}}^{\widehat{\tau}_{k,n-1}}\bigg(\frac{s_{\tau}^{\star}(x_{\tau})}{2(1-\tau)^{3/2}} - \frac{s_{\tau_{k,n-1}}^{\star}(x_{\tau_{k, n-1}})}{2(1-\tau_{k,n-1})^{3/2}}\bigg) \mathrm{d} \tau \\
&\overset{\text{(a)}}{=} -\int_{\tau_{k, n}}^{\widehat{\tau}_{k,n-1}}\bigg(\int_{\tau'}^{\tau_{k,n-1}} \frac{\partial}{\partial \tau}\frac{s_{\tau}^{\star}(x_{\tau})}{2(1-\tau)^{3/2}} \mathrm{d} \tau\bigg) \mathrm{d} \tau' \\
&\overset{\text{(b)}}{=} -\int_{\tau_{k,n}}^{{\tau}_{k,n-1}} \frac{\partial}{\partial \tau}\frac{s_{\tau}^{\star}(x_{\tau})}{2(1-\tau)^{3/2}}\bigg(\int_{\tau_{k, n}}^{\widehat{\tau}_{k,n-1}}\ind(\tau'<\tau)\mathrm{d} \tau'\bigg) \mathrm{d} \tau \\
&= -\int_{\tau_{k, n}}^{\widehat{\tau}_{k,n-1}} (\tau-{\tau}_{k,n}) \frac{\partial}{\partial \tau}\frac{s_{\tau}^{\star}(x_{\tau})}{2(1-\tau)^{3/2}} \mathrm{d} \tau  -\int_{\widehat{\tau}_{k, n-1}}^{{\tau}_{k,n-1}} (\widehat{\tau}_{k,n-1}\notag\\
&\quad -{\tau}_{k,n}) \frac{\partial}{\partial \tau}\frac{s_{\tau}^{\star}(x_{\tau})}{2(1-\tau)^{3/2}} \mathrm{d} \tau , 
\end{align*}
where (a) comes from the fact that
\begin{align*}
  \frac{s_{\tau'}^{\star}(x_{\tau'})}{2(1-\tau')^{3/2}} - \frac{s_{\tau_{k,n-1}}^{\star}(x_{\tau_{k, n-1}})}{2(1-\tau_{k,n-1})^{3/2}}
  = \int_{\tau_{k,n-1}}^{\tau'} \frac{\partial}{\partial \tau}\frac{s_{\tau}^{\star}(x_{\tau})}{2(1-\tau)^{3/2}} \mathrm{d} \tau
\end{align*}
and the $\ind(\tau'<\tau)$ in (b) denotes an indicator function.
We further have
\begin{align*}
\|\mathcal{E}_{1,1}(x_{\tau_{k,0}})\|_2^2
&\le (\widehat{\tau}_{k,n-1}-{\tau}_{k,n})^2\left(\int_{\tau_{k, n}}^{{\tau}_{k,n-1}}  \left\|\frac{\partial}{\partial \tau}\frac{s_{\tau}^{\star}(x_{\tau})}{2(1-\tau)^{3/2}}\right\|_2 \mathrm{d} \tau \right)^2\nonumber\\
&\le(\widehat{\tau}_{k,n-1}-\widehat{\tau}_{k,n})^2(\widehat{\tau}_{k,n-2}-\widehat{\tau}_{k,n})\int_{{\tau}_{k, n}}^{\widehat{\tau}_{k,n-2}}  \left\|\frac{\partial}{\partial \tau}\frac{s_{\tau}^{\star}(x_{\tau})}{2(1-\tau)^{3/2}}\right\|_2^2 \mathrm{d} \tau.
\end{align*}
By using \eqref{eq:diff-tau-1} and \eqref{eq:diff-tau-2}, we have
\begin{align*}
\|\mathcal{E}_{1,1}(x_{\tau_{k,0}})\|_2^2 
&\le
\left(1+\frac{\widehat{\tau}_{k,n-2}-\widehat{\tau}_{k,n-1}}{\widehat{\tau}_{k,n-1}-\widehat{\tau}_{k,n}}\right)(\widehat{\tau}_{k,n-1}-\widehat{\tau}_{k,n})^3\int_{{\tau}_{k, n}}^{\widehat{\tau}_{k,n-1}}  \left\|\frac{\partial}{\partial \tau}\frac{s_{\tau}^{\star}(x_{\tau})}{2(1-\tau)^{3/2}}\right\|_2^2 \mathrm{d} \tau\\
&\quad+\left(\frac{(\widehat{\tau}_{k,n-1}-\widehat{\tau}_{k,n})^3}{(\widehat{\tau}_{k,n-2}-\widehat{\tau}_{k,n-1})^3}+\frac{(\widehat{\tau}_{k,n-1}-\widehat{\tau}_{k,n})^2}{(\widehat{\tau}_{k,n-2}-\widehat{\tau}_{k,n-1})^2}\right)(\widehat{\tau}_{k,n-2}-\widehat{\tau}_{k,n-1})^3\notag\\
&\qquad\cdot\int_{\widehat{\tau}_{k, n-1}}^{\widehat{\tau}_{k,n-2}}   \left\|\frac{\partial}{\partial \tau}\frac{s_{\tau}^{\star}(x_{\tau})}{2(1-\tau)^{3/2}}\right\|_2^2 \mathrm{d} \tau\\
&\overset{\text{(a)}}{\lesssim} (\widehat{\tau}_{k,n-1}-\widehat{\tau}_{k,n})^3\int_{{\tau}_{k, n}}^{\widehat{\tau}_{k,n-1}}  \left\|\frac{\partial}{\partial \tau}\frac{s_{\tau}^{\star}(x_{\tau})}{2(1-\tau)^{3/2}}\right\|_2^2 \mathrm{d} \tau \notag\\
&\quad + (\widehat{\tau}_{k,n-2}-\widehat{\tau}_{k,n-1})^3\int_{\widehat{\tau}_{k, n-1}}^{\widehat{\tau}_{k,n-2}}  \left\|\frac{\partial}{\partial \tau}\frac{s_{\tau}^{\star}(x_{\tau})}{2(1-\tau)^{3/2}}\right\|_2^2 \mathrm{d} \tau,
\end{align*}
where (a) uses \eqref{eq:diff-tau-2}.
By taking expectation, we have,
\begin{align*}
\mathbb{E}_{x_k \sim {\widehat{X}_k}}\left[\|\mathcal{E}_{1,1}(x_k)\|_2^2\right] 
&\lesssim 
(\widehat{\tau}_{k,n-1}-\widehat{\tau}_{k,n})^3\int_{{\tau}_{k, n}}^{\widehat{\tau}_{k,n-1}}  \mathbb{E}_{x_{\tau} \sim X_\tau}\bigg[\left\|\frac{\partial}{\partial \tau}\frac{s_{\tau}^{\star}(x_{\tau})}{2(1-\tau)^{3/2}}\right\|_2^2\bigg] \mathrm{d} \tau \nonumber\\
&\quad + (\widehat{\tau}_{k,n-2}-\widehat{\tau}_{k,n-1})^3\int_{\widehat{\tau}_{k, n-1}}^{\widehat{\tau}_{k,n-2}}  \mathbb{E}_{x_{\tau} \sim X_\tau}\bigg[\left\|\frac{\partial}{\partial \tau}\frac{s_{\tau}^{\star}(x_{\tau})}{2(1-\tau)^{3/2}}\right\|_2^2\bigg] \mathrm{d} \tau.
% \sum_{i=n-1}^{n}(\widehat{\tau}_{k,i-1}-\widehat{\tau}_{k,i})^3\int_{\widehat{\tau}_{k, i}}^{\widehat{\tau}_{k,i-1}}  \mathbb{E}_{x_{\tau} \sim X_\tau}\bigg[\left\|\frac{\partial}{\partial \tau}\frac{s_{\tau}^{\star}(x_{\tau})}{2(1-\tau)^{3/2}}\right\|^2\bigg] \mathrm{d} \tau.
%(\widehat{\tau}_{k,n-1} - \widehat{\tau}_{k,n})^2(\widehat{\tau}_{k,n-2}-\widehat{\tau}_{k,n})\int_{\tau_{k, n}}^{{\tau}_{k,n-1}} \mathbb{E}_{x_{\tau} \sim X_\tau}\bigg[\Big\|\frac{\partial}{\partial \tau}\frac{s_{\tau}^{\star}(x_{\tau})}{(1-\tau)^{3/2}}\Big\|_2^2\bigg]  \mathrm{d} \tau.
\end{align*}
%\item
\emph{2. Analysis of $\mathcal{E}_{1,0}(x_{\tau_{k,0}})$.}
This term can be calculated in a similar way as
\begin{align*}
\mathcal{E}_{1,0}(x_{\tau_{k,0}}) &= \int_{\widehat{\tau}_{k, 0}}^{{\tau}_{k,0}}\bigg(\frac{s_{\tau}^{\star}(x_{\tau})}{2(1-\tau)^{3/2}} - \frac{s_{\tau_{k,0}}^{\star}(x_{\tau_{k, 0}})}{2(1-\tau_{k,0})^{3/2}}\bigg) \mathrm{d} \tau \\
&= -\int_{\widehat{\tau}_{k, 0}}^{{\tau}_{k,0}}\bigg(\int_{\tau'}^{\tau_{k,0}} \frac{\partial}{\partial \tau}\frac{s_{\tau}^{\star}(x_{\tau})}{2(1-\tau)^{3/2}} \mathrm{d} \tau\bigg) \mathrm{d} \tau' \\
&= -\int_{\widehat{\tau}_{k,0}}^{{\tau}_{k,0}} \frac{\partial}{\partial \tau}\frac{s_{\tau}^{\star}(x_{\tau})}{2(1-\tau)^{3/2}}\bigg(\int_{\widehat{\tau}_{k,0}}^{{\tau}_{k,0}}\ind(\tau'<\tau)\mathrm{d} \tau'\bigg) \mathrm{d} \tau \\
&= -\int_{\widehat{\tau}_{k, 0}}^{{\tau}_{k,0}} (\tau-\widehat{\tau}_{k,0}) \frac{\partial}{\partial \tau}\frac{s_{\tau}^{\star}(x_{\tau})}{2(1-\tau)^{3/2}} \mathrm{d} \tau.
\end{align*}
We further have
\begin{align*}
\|\mathcal{E}_{1,0}(x_{\tau_{k,0}})\|_2^2\le (\tau_{k,0}-\widehat{\tau}_{k,0})^3\int_{\widehat{\tau}_{k, 0}}^{{\tau}_{k,0}}  \left\|\frac{\partial}{\partial \tau}\frac{s_{\tau}^{\star}(x_{\tau})}{2(1-\tau)^{3/2}}\right\|_2^2 \mathrm{d} \tau.
\end{align*}
By taking expectation, we have
\begin{align*}
% \mathbb{E}_{x_{\tau_{k,0}} \sim p_{\widehat{X}_k}}\left[\|\mathcal{E}_{1,1}(x_k)\|_2^2 \mymid \tau\right] 
% \le&  (\widehat{\tau}_{k,n-1} - \widehat{\tau}_{k,n})^2(\widehat{\tau}_{k,n-2}-\widehat{\tau}_{k,n})\int_{\tau_{k, n}}^{{\tau}_{k,n-1}} \mathbb{E}_{x_{\tau} \sim X_\tau}\bigg[\Big\|\frac{\partial}{\partial \tau}\frac{s_{\tau}^{\star}(x_{\tau})}{(1-\tau)^{3/2}}\Big\|_2^2\bigg]  \mathrm{d} \tau,\nonumber\\
% \overset{\text{(a)}}{\lesssim}&  (\widehat{\tau}_{k,n-1} - \widehat{\tau}_{k,n})^3\int_{\tau_{k, n}}^{{\tau}_{k,n-1}} \mathbb{E}_{x_{\tau} \sim X_\tau}\bigg[\Big\|\frac{\partial}{\partial \tau}\frac{s_{\tau}^{\star}(x_{\tau})}{(1-\tau)^{3/2}}\Big\|_2^2\bigg]  \mathrm{d} \tau,\nonumber\\
\mathbb{E}_{x_{\tau_{k,0}} \sim {\widehat{X}_k}}\left[\|\mathcal{E}_{1,0}(x_{\tau_{k,0}})\|_2^2\right] 
\lesssim& ({\tau}_{k,0} - \widehat{\tau}_{k,0})^3\int_{\widehat{\tau}_{k, 0}}^{{\tau}_{k,0}} \mathbb{E}_{x_{\tau} \sim X_\tau}\bigg[\Big\|\frac{\partial}{\partial \tau}\frac{s_{\tau}^{\star}(x_{\tau})}{(1-\tau)^{3/2}}\Big\|_2^2\bigg]  \mathrm{d} \tau\nonumber\\
\lesssim& (\widehat{\tau}_{k,-1} - \widehat{\tau}_{k,0})^3\int_{\widehat{\tau}_{k, 0}}^{{\tau}_{k,0}} \mathbb{E}_{x_{\tau} \sim X_\tau}\bigg[\Big\|\frac{\partial}{\partial \tau}\frac{s_{\tau}^{\star}(x_{\tau})}{(1-\tau)^{3/2}}\Big\|_2^2\bigg]  \mathrm{d} \tau.
\end{align*}
%where $\widehat{\tau}_{k,-1} = 1-\widehat{\alpha}_{T-\frac{kN}{2}+1}$, and 
% $$
% \widehat{\tau}_{k,-1}-\widehat{\tau}_{k,0} = \frac{c_1\widehat{\tau}_{k,-1}(1-\widehat{\tau}_{k,-1})\log T}{T}.
% $$
%\item Next, we intend to bound $\mathcal{E}_2(x_{\tau_{k,0}})$, which is a random variable about $\tau=\{\tau_{k,n}\}_{n=1}^{N-1}$. 
\emph{3. Analysis of $\mathcal{E}_2(x_{\tau_{k,0}})$.}
This term is a random variable about $\tau=\{\tau_{k,n}\}_{n=1}^{N-1}$. 
To bound it with high probability, we intend to calculate its $r$-th order moment $\mathbb{E}\big[\big(\mathbb{E}_{x_{\tau_{k,0}} \sim p_{\widehat{X}_k}}\big[\|\mathcal{E}_2(x_{\tau_{k,0}})\|_2^2 \mymid \tau\big]\big)^r\big] $.
Towards this, we introduce  $r$  independent random variables $x_k^{(j)}\sim p_{\widehat{X}_k}$, $j=1,\cdots,r$.
For any integer $r > 0$, considering that $\mathbb{E}_{x_{\tau_{k,0}} \sim p_{\widehat{X}_k}}\big[\|\mathcal{E}_2(x_{\tau_{k,0}})\|_2^2 \mymid \tau\big]$ is a random variable independent on $x_{\tau_{k,0}}$, we get 
\begin{align}\label{eq:r-moment}
\mathbb{E}\big[\big(\mathbb{E}_{x_{\tau_{k,0}} \sim p_{\widehat{X}_k}}\big[\|\mathcal{E}_2(x_{\tau_{k,0}})\|_2^2 \mymid \tau\big]\big)^r\big] 
&= \mathbb{E}\Big[\mathbb{E}_{x_k^{(j)} \sim p_{\widehat{X}_k}}\Big[\prod_{1\le j \le r}\|\mathcal{E}_2(x_k^{(j)})\|_2^2 \mymid \tau\Big]\Big] \nonumber\\
&= \mathbb{E}_{x_k^{(j)} \sim p_{\widehat{X}_k}}\Big[\mathbb{E}\Big[\prod_{1\le j \le r}\|\mathcal{E}_2(x_k^{(j)})\|_2^2 \mymid x_k^{(j)}\Big]\Big].
\end{align}
To control the above term, we first make the following observations about the expectation and bias of random variable $\frac{s_{\tau_{k,i}}^{\star}(x_{\tau_{k,i}})}{2(1-\tau_{k,i})^{3/2}}$ conditioned on $x_{\tau_{k,0}} = x_k^{(j)}$:
\begin{align*}
\mathbb{E}\bigg[\frac{s_{\tau_{k,i}}^{\star}(x_{\tau_{k,i}})}{2(1-\tau_{k,i})^{3/2}}(\widehat{\tau}_{k,i-1} - \widehat{\tau}_{k,i})|x_k^{(j)}\bigg] &\overset{\text{(a)}}{=} 
\int_{\widehat{\tau}_{k,i}}^{\widehat{\tau}_{k,i-1}}\frac{s_{\tau}^{\star}(x_{\tau})}{2(1-\tau)^{3/2}}(\widehat{\tau}_{k,i-1} - \widehat{\tau}_{k,i})\frac{1}{\widehat{\tau}_{k,i-1} - \widehat{\tau}_{k,i}} \mathrm{d} \tau\\
&=\int_{\widehat{\tau}_{k,i}}^{\widehat{\tau}_{k,i-1}}\frac{s_{\tau}^{\star}(x_{\tau})}{2(1-\tau)^{3/2}} \mathrm{d} \tau,
\end{align*}
where (a) comes from the fact that $\tau_{k,i}$ is uniformly distributed within $[\widehat{\tau}_{k,i},\widehat{\tau}_{k,i-1}]$,
and
\begin{align*}
&\quad\int_{\widehat{\tau}_{k,i}}^{\widehat{\tau}_{k,i-1}}\frac{s_{\tau}^{\star}(x_{\tau})}{2(1-\tau)^{3/2}} \mathrm{d} \tau - \frac{s_{\tau_{k,i}}^{\star}(x_{\tau_{k,i}})}{2(1-\tau_{k,i})^{3/2}}(\widehat{\tau}_{k,i-1} - \widehat{\tau}_{k,i})\\
&=\int_{\widehat{\tau}_{k,i}}^{\widehat{\tau}_{k,i-1}}\left[\frac{s_{\tau}^{\star}(x_{\tau})}{2(1-\tau)^{3/2}}  - \frac{s_{\tau_{k,i}}^{\star}(x_{\tau_{k,i}})}{2(1-\tau_{k,i})^{3/2}}\right]\mathrm{d} \tau\\
&=\int_{\widehat{\tau}_{k,i}}^{\widehat{\tau}_{k,i-1}}\int_{\tau_{k,i}}^{\tau'}\frac{\partial}{\partial \tau}\frac{s_{\tau}^{\star}(x_{\tau})}{2(1-\tau)^{3/2}} \mathrm{d} \tau \mathrm{d} \tau'\\
&=\left(\int_{{\tau}_{k,i}}^{\widehat{\tau}_{k,i-1}}\int_{\tau}^{\widehat{\tau}_{k,i-1}}+\int_{\widehat{\tau}_{k,i}}^{{\tau}_{k,i}}\int_{\widehat{\tau}_{k,i}}^{\tau}\right)\frac{\partial}{\partial \tau}\frac{s_{\tau}^{\star}(x_{\tau})}{2(1-\tau)^{3/2}} \mathrm{d} \tau' \mathrm{d} \tau\\
% &=\int_{{\tau}_{k,i}}^{\widehat{\tau}_{k,i-1}}\int_{\tau}^{\widehat{\tau}_{k,i-1}}\frac{\partial}{\partial \tau}\frac{s_{\tau}^{\star}(x_{\tau})}{2(1-\tau)^{3/2}} \mathrm{d} \tau' \mathrm{d} \tau\\
&=\int_{{\tau}_{k,i}}^{\widehat{\tau}_{k,i-1}}(\widehat{\tau}_{k,i-1}-\tau)\frac{\partial}{\partial \tau}\frac{s_{\tau}^{\star}(x_{\tau})}{2(1-\tau)^{3/2}}  \mathrm{d} \tau+\int_{\widehat{\tau}_{k,i}}^{{\tau}_{k,i}}(\tau-\widehat{\tau}_{k,i})\frac{\partial}{\partial \tau}\frac{s_{\tau}^{\star}(x_{\tau})}{2(1-\tau)^{3/2}}  \mathrm{d} \tau
%&= \int_{\widehat{\tau}_{k,i}}^{\widehat{\tau}_{k,i-1}} (\tau-{\tau}_{k,i}) \frac{\partial}{\partial \tau}\frac{s_{\tau}^{\star}(x_{\tau})}{2(1-\tau)^{3/2}} \mathrm{d} \tau.
\end{align*}
Furthermore,  we have
\begin{align*}
&\quad \left\|\int_{\widehat{\tau}_{k,i}}^{\widehat{\tau}_{k,i-1}}\frac{s_{\tau}^{\star}(x_{\tau})}{2(1-\tau)^{3/2}} \mathrm{d} \tau - \frac{s_{\tau_{k,i}}^{\star}(x_{\tau_{k,i}})}{2(1-\tau_{k,i})^{3/2}}(\widehat{\tau}_{k,i-1} - \widehat{\tau}_{k,i})\right\|_2\\
&\le (\widehat{\tau}_{k,i-1}-\widehat{\tau}_{k,i})\int_{{\tau}_{k,i}}^{\widehat{\tau}_{k,i-1}}\left\|\frac{\partial}{\partial \tau}\frac{s_{\tau}^{\star}(x_{\tau})}{2(1-\tau)^{3/2}}\right\|_2  \mathrm{d} \tau+(\widehat{\tau}_{k,i-1}-\widehat{\tau}_{k,i})\int_{\widehat{\tau}_{k,i}}^{{\tau}_{k,i}}\left\|\frac{\partial}{\partial \tau}\frac{s_{\tau}^{\star}(x_{\tau})}{2(1-\tau)^{3/2}}\right\|_2  \mathrm{d} \tau\\
%&\le \int_{\widehat{\tau}_{k,i}}^{\widehat{\tau}_{k,i-1}} (\widehat{\tau}_{k,i-1}-\widehat{\tau}_{k,i}) \left\|\frac{\partial}{\partial \tau}\frac{s_{\tau}^{\star}(x_{\tau})}{2(1-\tau)^{3/2}}\right\|_2 \mathrm{d} \tau\\
&\le (\widehat{\tau}_{k,i-1}-\widehat{\tau}_{k,i})\int_{\widehat{\tau}_{k,i}}^{\widehat{\tau}_{k,i-1}}  \left\|\frac{\partial}{\partial \tau}\frac{s_{\tau}^{\star}(x_{\tau})}{2(1-\tau)^{3/2}}\right\|_2 \mathrm{d} \tau.
\end{align*}
Then given $x_k^{(j)}$, according to Bernstein inequality, we have with probability at least $1-\delta$, for all $1\le j\le r$,
\begin{align}\label{eq:bound-E2-bernstein}
\|\mathcal{E}_2(x_k^{(j)})\|_2^2 
&\lesssim \log\frac{r}{\delta}\sum_{i = 1}^{n-1} \bigg[(\widehat{\tau}_{k,i-1} - \widehat{\tau}_{k,i})\int_{\widehat{\tau}_{k,i}}^{\widehat{\tau}_{k,i-1}} \Big\|\frac{\partial}{\partial \tau}\frac{s_{\tau}^{\star}(x_{\tau})}{(1-\tau)^{3/2}}\Big\|_2 \mathrm{d} \tau\bigg]^2\nonumber\\
&\overset{\text{(a)}}{\lesssim} \log\frac{r}{\delta}\sum_{i = 1}^{n-1} (\widehat{\tau}_{k,i-1} - \widehat{\tau}_{k,i})^3\int_{\widehat{\tau}_{k,i}}^{\widehat{\tau}_{k,i-1}} \Big\|\frac{\partial}{\partial \tau}\frac{s_{\tau}^{\star}(x_{\tau})}{(1-\tau)^{3/2}}\Big\|_2^2 \mathrm{d} \tau,
\end{align}
where (a) uses the cauchy-schwarz inequality.
% For convenience, we introduce a temporary variable as below.
% \begin{align*}
% A_{k,i}:=  \bigg[(\widehat{\tau}_{k,i-1} - \widehat{\tau}_{k,i})\int_{\widehat{\tau}_{k,i}}^{\widehat{\tau}_{k,i-1}} \Big\|\frac{\partial}{\partial \tau}\frac{s_{\tau}^{\star}(x_{\tau})}{(1-\tau)^{3/2}}\Big\|_2 \mathrm{d} \tau\bigg]^2.
% \end{align*}
% We bound $A_{k,i}$ as 
% \begin{align}\label{eq:bound-Aki}
% A_{k,i} &= (\widehat{\tau}_{k,i-1} - \widehat{\tau}_{k,i})^2 \bigg[\int_{\widehat{\tau}_{k,i}}^{\widehat{\tau}_{k,i-1}} \Big\|\frac{\partial}{\partial \tau}\frac{s_{\tau}^{\star}(x_{\tau})}{(1-\tau)^{3/2}}\Big\|_2 \mathrm{d} \tau\bigg]^2\nonumber\\
% &\overset{(a)}{\le} (\widehat{\tau}_{k,i-1} - \widehat{\tau}_{k,i})^2(\widehat{\tau}_{k,i-1} - \widehat{\tau}_{k,i})\int_{\widehat{\tau}_{k,i}}^{\widehat{\tau}_{k,i-1}} \Big\|\frac{\partial}{\partial \tau}\frac{s_{\tau}^{\star}(x_{\tau})}{(1-\tau)^{3/2}}\Big\|_2^2 \mathrm{d} \tau\nonumber\\
% &=(\widehat{\tau}_{k,i-1} - \widehat{\tau}_{k,i})^3\int_{\widehat{\tau}_{k,i}}^{\widehat{\tau}_{k,i-1}} \Big\|\frac{\partial}{\partial \tau}\frac{s_{\tau}^{\star}(x_{\tau})}{(1-\tau)^{3/2}}\Big\|_2^2 \mathrm{d} \tau,
% \end{align}
% where (a) uses the cauchy-schwarz inequality.

In addition, notice the relation that for $\mathbb{P}(X > \log\frac{r}{\delta}) < \frac{\delta}{r}$ and $Y\sim \mathsf{Exp}(1)$, the following inequality holds.
\begin{align*}
\mathbb{E}[X^r] \le \mathbb{E}[Y^r] = r!.
\end{align*}
Then the \eqref{eq:bound-E2-bernstein} implies that
\begin{align}\label{eq:bound-E2-prob}
\mathbb{E}\Big[\prod_{1\le j \le r}\|\mathcal{E}_2(x_k^{(j)})\|_2^2 \mymid x_k^{(j)}\Big]
&\lesssim r! \left(\sum_{i = 1}^{n-1} (\widehat{\tau}_{k,i-1} - \widehat{\tau}_{k,i})^3\int_{\widehat{\tau}_{k,i}}^{\widehat{\tau}_{k,i-1}} \Big\|\frac{\partial}{\partial \tau}\frac{s_{\tau}^{\star}(x_{\tau})}{(1-\tau)^{3/2}}\Big\|_2^2 \mathrm{d} \tau\right)^r.
\end{align}
% Combining \eqref{eq:bound-Aki} and \eqref{eq:bound-E2-prob}, we have that
% \begin{align*}
% \mathbb{E}\Big[\prod_{1\le j \le r}\|\mathcal{E}_2(x_k^{(j)})\|_2^2 \mymid x_k^{(j)}\Big]
% &\lesssim r!\prod_{1\le j \le r}\sum_{i = 1}^{n-1} (\widehat{\tau}_{k,i-1} - \widehat{\tau}_{k,i})^3\int_{\widehat{\tau}_{k,i}}^{\widehat{\tau}_{k,i-1}} \Big\|\frac{\partial}{\partial \tau}\frac{s_{\tau}^{\star}(x_{\tau}^{(j)})}{(1-\tau)^{3/2}}\Big\|_2^2 \mathrm{d} \tau.
% \end{align*}
Inserting \eqref{eq:bound-E2-prob} into \eqref{eq:r-moment}, we have
\begin{align*}
&\quad \mathbb{E}\big[\big(\mathbb{E}_{x_k \sim p_{\widehat{X}_k}}\big[\|\mathcal{E}_2(x_k)\|_2^2 \mymid \tau\big]\big)^r\big]\nonumber\\
&\lesssim r!\left(\sum_{i = 1}^{n-1} (\widehat{\tau}_{k,i-1} - \widehat{\tau}_{k,i})^3\int_{\widehat{\tau}_{k,i}}^{\widehat{\tau}_{k,i-1}} \mathbb{E}_{x_{\tau} \sim X_\tau}\left[\Big\|\frac{\partial}{\partial \tau}\frac{s_{\tau}^{\star}(x_{\tau}^{(j)})}{(1-\tau)^{3/2}}\Big\|_2^2\right] \mathrm{d} \tau\right)^r,
\end{align*}
which tells us that with probability at least $1 - T^{-100}$,
\begin{align*}
\mathbb{E}_{x_k \sim p_{\widehat{X}_k}}\big[\|\mathcal{E}_2(x_k)\|_2^2\big] 
&\lesssim \log T\sum_{i = 1}^{n-1} (\widehat{\tau}_{k,i-1} - \widehat{\tau}_{k,i})^3\int_{\widehat{\tau}_{k,i}}^{\widehat{\tau}_{k,i-1}} \mathbb{E}_{x_{\tau} \sim X_\tau}\bigg[\Big\|\frac{\partial}{\partial \tau}\frac{s_{\tau}^{\star}(x_{\tau})}{(1-\tau)^{3/2}}\Big\|_2^2\bigg]  \mathrm{d} \tau.
\end{align*}
%It is obvious that $\mathbb{E}_{x_k \sim p_{\widehat{X}_k}}\big[\|\mathcal{E}_2(x_k)\|_2^2 \big] $ obeys the same bound. 
%\end{itemize}
\emph{4. Combining the above results.} Combining these together, we have
\begin{align}\label{eq:proof-temp-1}
&\quad \frac{\mathbb{E}_{x_k \sim p_{\widehat{X}_{k}}}\big[\|x_{\tau_{k,n}}(x_k)-z_{\tau_{k,n}}^{\star}(x_k)\|_2^2\big]}{1-\tau_{k,n}} \nonumber\\
&\lesssim \log T\sum_{i = 1}^{n-1} (\widehat{\tau}_{k,i-1} - \widehat{\tau}_{k,i})^3\int_{\widehat{\tau}_{k,i}}^{\widehat{\tau}_{k,i-1}} \mathbb{E}_{x_{\tau} \sim X_\tau}\bigg[\Big\|\frac{\partial}{\partial \tau}\frac{s_{\tau}^{\star}(x_{\tau})}{(1-\tau)^{3/2}}\Big\|_2^2\bigg]  \mathrm{d} \tau\nonumber\\
&\quad +(\widehat{\tau}_{k,-1} - \widehat{\tau}_{k,0})^3\int_{\widehat{\tau}_{k, 0}}^{{\tau}_{k,0}} \mathbb{E}_{x_{\tau} \sim X_\tau}\bigg[\Big\|\frac{\partial}{\partial \tau}\frac{s_{\tau}^{\star}(x_{\tau})}{(1-\tau)^{3/2}}\Big\|_2^2\bigg]  \mathrm{d} \tau\nonumber\\
&\quad +(\widehat{\tau}_{k,n-1}-\widehat{\tau}_{k,n})^3\int_{{\tau}_{k, n}}^{\widehat{\tau}_{k,n-1}}  \mathbb{E}_{x_{\tau} \sim X_\tau}\bigg[\left\|\frac{\partial}{\partial \tau}\frac{s_{\tau}^{\star}(x_{\tau})}{2(1-\tau)^{3/2}}\right\|^2\bigg] \mathrm{d} \tau.
\end{align}

We claim that for any $0<\tau<1$, 
\begin{align}\label{eq:bound-deriv}
\mathbb{E}_{x_{\tau} \sim X_\tau}\bigg[\Big\|\frac{\partial}{\partial \tau}\frac{s_{\tau}^{\star}(x_{\tau})}{(1-\tau)^{3/2}}\Big\|_2^2\bigg]
&\lesssim \frac{d\log T}{(1-\tau)^5\tau^3}\min\left\{d + \mathbb{E}[\mathsf{Tr}(\Sigma_\tau^2(x_\tau))], L^2\right\}.
\end{align}
The proof is deferred to the end of this section.
Inserting into \eqref{eq:proof-temp-1}, we have
\begin{align*}
&\quad(\widehat{\tau}_{k,i-1} - \widehat{\tau}_{k,i})^3\int_{\widehat{\tau}_{k,i}}^{\widehat{\tau}_{k,i-1}} \mathbb{E}_{x_{\tau} \sim X_\tau}\bigg[\Big\|\frac{\partial}{\partial \tau}\frac{s_{\tau}^{\star}(x_{\tau})}{(1-\tau)^{3/2}}\Big\|_2^2\bigg]  \mathrm{d} \tau\\
&\lesssim \frac{(\widehat{\tau}_{k,i-1} - \widehat{\tau}_{k,i})^3d\log T}{(1-\widehat{\tau}_{k,i-1})^3\widehat{\tau}_{k,i}^3}\min\left\{\frac{d(\widehat{\tau}_{k,i-1}-\widehat{\tau}_{k,i})}{(1-\widehat{\tau}_{k,i-1})^2} + \int_{\widehat{\tau}_{k,i}}^{\widehat{\tau}_{k,i-1}} \frac{\mathbb{E}[\mathsf{Tr}(\Sigma_\tau^2(x_\tau))]}{(1-\tau)^2}\mathrm d \tau,\frac{L^2(\widehat{\tau}_{k,i-1}-\widehat{\tau}_{k,i})}{(1-\widehat{\tau}_{k,i-1})^2}\right\}\nonumber\\
&\overset{\text{(a)}}{\lesssim} \frac{d\log^4 T}{T^3}\min\left\{\frac{d\widehat{\tau}_{k,i-1}\log T}{(1-\widehat{\tau}_{k,i-1})T} + \int_{\widehat{\tau}_{k,i}}^{\widehat{\tau}_{k,i-1}} \frac{\mathbb{E}[\mathsf{Tr}(\Sigma_\tau^2(x_\tau))]}{(1-\tau)^2}\mathrm d \tau,\frac{L^2\widehat{\tau}_{k,i-1}\log T}{(1-\widehat{\tau}_{k,i-1})T}\right\},
\end{align*}
where (a) uses \eqref{eq:diff-tau-1} and \eqref{eq:diff-tau-2}.
Similarly, we have
\begin{align*}
&\quad (\widehat{\tau}_{k,-1} - \widehat{\tau}_{k,0})^3\int_{\widehat{\tau}_{k, 0}}^{{\tau}_{k,0}} \mathbb{E}_{x_{\tau} \sim X_\tau}\bigg[\Big\|\frac{\partial}{\partial \tau}\frac{s_{\tau}^{\star}(x_{\tau})}{(1-\tau)^{3/2}}\Big\|_2^2\bigg]  \mathrm{d} \tau\nonumber\\
&\lesssim \frac{d\log^4 T}{T^3}\min\left\{\frac{d\widehat{\tau}_{k,-1}\log T}{(1-\widehat{\tau}_{k,-1})T} + \int_{\widehat{\tau}_{k,0}}^{{\tau}_{k,0}} \frac{\mathbb{E}[\mathsf{Tr}(\Sigma_\tau^2(x_\tau))]}{(1-\tau)^2}\mathrm d \tau,\frac{L^2\widehat{\tau}_{k,-1}\log T}{(1-\widehat{\tau}_{k,-1})T}\right\},\nonumber\\
&\quad (\widehat{\tau}_{k,n-1}-\widehat{\tau}_{k,n})^3\int_{{\tau}_{k, n}}^{\widehat{\tau}_{k,n-1}}  \mathbb{E}_{x_{\tau} \sim X_\tau}\bigg[\left\|\frac{\partial}{\partial \tau}\frac{s_{\tau}^{\star}(x_{\tau})}{2(1-\tau)^{3/2}}\right\|_2^2\bigg] \mathrm{d} \tau\nonumber\\
&\lesssim \frac{d\log^4 T}{T^3}\min\left\{\frac{d\widehat{\tau}_{k,n-1}\log T}{(1-\widehat{\tau}_{k,n-1})T} + \int_{{\tau}_{k,n}}^{\widehat{\tau}_{k,n-1}} \frac{\mathbb{E}[\mathsf{Tr}(\Sigma_\tau^2(x_\tau))]}{(1-\tau)^2}\mathrm d \tau,\frac{L^2\widehat{\tau}_{k,n-1}\log T}{(1-\widehat{\tau}_{k,n-1})T}\right\}.
\end{align*}
Thus we have
\begin{align*}
&\quad\frac{\mathbb{E}_{x_k \sim p_{\widehat{X}_{k}}}\big[\|x_{\tau_{k,n}}(x_k)-z_{\tau_{k,n}}^{\star}(x_k)\|_2^2\big]}{1-\tau_{k,n}}\notag\\
&\lesssim \min\Big\{\frac{Nd^2\widehat{\tau}_{k,-1}\log^6 T}{T^4(1-\widehat{\tau}_{k,-1})}+\frac{d\log^5 T}{T^3}\int_{{\tau}_{k,n}}^{{\tau}_{k,0}} \frac{\mathbb{E}[\mathsf{Tr}(\Sigma_\tau^2(x_\tau))]}{(1-\tau)^2}\mathrm d \tau,\frac{NdL^2\widehat{\tau}_{k,-1}\log^6 T}{(1-\widehat{\tau}_{k,-1})T^4}\Big\}
\end{align*}
and complete the proof.

% We assume that
% \begin{align*}
% \left\|y_{\tau_{k,i}} - x_{\tau_{k,i}}\right\| \lesssim \sqrt{\frac{\tau_{k,i}}{d\log T}},\quad i\le n-1,
% \end{align*}
% for any $x_{\tau}\in\mathcal{S}_{\tau}$, $y_{\tau_{k,i}}\in\mathcal{S}_{\tau_{k,i}}$.
% In the end of the proof, we shall show that this holds for $i=n$.
% Denote
% $$
% \mathcal{D} = \{x_{\tau_{k,0}}: x_{\tau}\in\mathcal{S}_\tau, y_{\tau_{k,i}}\in\mathcal{S}_{\tau_{k,i}}\}.
% $$
% We have
% \begin{align*}
% &\quad \mathbb{E}_{x_{\tau_{k,0}}\sim X_{\tau_{k,0}}}\big\|s_{\tau_{k,i}}(y_{\tau_{k,i}}) - s_{\tau_{k,i}}^{\star}(y_{\tau_{k,i}})\big\|_2^2 \nonumber\\
% &= \int_{\mathcal{D}} \big\|s_{\tau_{k,i}}(y_{\tau_{k,i}}) - s_{\tau_{k,i}}^{\star}(y_{\tau_{k,i}})\big\|_2^2 p_{\tau_{k,i}}(x_{\tau_{k,i}}) \mathrm{d} x_{\tau_{k,i}}
% +\int_{\mathcal{D}^c} \big\|s_{\tau_{k,i}}(y_{\tau_{k,i}}) - s_{\tau_{k,i}}^{\star}(y_{\tau_{k,i}})\big\|_2^2 p_{\tau_{k,i}}(x_{\tau_{k,i}}) \mathrm{d} x_{\tau_{k,i}}
% \nonumber\\
% &\le 2\int_{\tau_{k,0}} \big\|s_{\tau_{k,i}}(y_{\tau_{k,i}}) - s_{\tau_{k,i}}^{\star}(y_{\tau_{k,i}})\big\|_2^2 p_{\tau_{k,i}}(y_{\tau_{k,i}}) \mathrm{d} y_{\tau_{k,0}}\nonumber\\
% &\le2\varepsilon_{k,i}^2
% \end{align*}
% and complete the proof.

\noindent \textbf{Proof of \eqref{eq:bound-deriv}.}
Notice that
\begin{align}\label{eq:proof-cal-deriv}
\frac{\partial}{\partial \tau}\frac{s_{\tau}^{\star}(x_{\tau})}{(1-\tau)^{3/2}} = \frac{\partial}{\partial y}\frac{ s_{\tau}^{\star}(\sqrt{1-\tau}y)}{(1-\tau)^{3/2}}\Big|_{y = \frac{x_{\tau}}{\sqrt{1-\tau}}}\frac{\partial }{\partial \tau}\frac{x_\tau}{\sqrt{1-\tau}} + \frac{\partial}{\partial \tau}\frac{s_{\tau}^{\star}(\sqrt{1-\tau}y)}{(1-\tau)^{3/2}}\Big|_{y = \frac{x_{\tau}}{\sqrt{1-\tau}}}.
\end{align}
For the first term, we have
\begin{align*}
\frac{\partial}{\partial y}\frac{ s_{\tau}^{\star}(\sqrt{1-\tau}y)}{(1-\tau)^{3/2}}\Big|_{y = \frac{x_{\tau}}{\sqrt{1-\tau}}}\frac{\partial }{\partial \tau}\frac{x_\tau}{\sqrt{1-\tau}} &= -\frac{\sqrt{1-\tau}}{(1-\tau)^{3/2}}J_\tau(x_{\tau})\frac{s_\tau^{\star}(x_\tau)}{(1-\tau)^{3/2}} \notag\\
&= -J_\tau(x_{\tau})\frac{s_\tau^{\star}(x_\tau)}{(1-\tau)^{5/2}},
\end{align*}
where 
$$
J_\tau(x):=\frac{\partial s_{\tau}^{\star}(x)}{\partial x}.
$$
According to Definition \ref{def:score-lipschitz}, we have $\tau\|J_\tau(x)\|\le L$ for $x\in\mathcal{L}_\tau$.
Moreover, we use the following lemma, the proof of which is postponed to the end of this section.
\begin{lemma}\label{lem:bound-s-J}
Suppose that $x_\tau\in\mathcal{S}_\tau$.
Then we have
\begin{align}\label{eq:bound-s}
\left\|s_{\tau}^{\star}(x_\tau)\right\|_2^2\le\frac{1}{\tau^2}\int\left\|x_\tau - \sqrt{1-\tau}x_0\right\|_2^2p_{X_0\mymid X_\tau}(x_0\mymid x_\tau)\mathrm{d} x_0 \le \frac{25(\theta+c_0)d\log T}{\tau}.
\end{align}
Moreover, we have
\begin{align}\label{eq:bound-J}
&\int_{x_\tau}\left\|J_\tau(x_{\tau})s_\tau^{\star}(x_\tau)\right\|_2^2p_\tau(x_\tau)\mathrm{d}x_{\tau}\lesssim \frac{d\log T}{\tau^3}\min\left\{d+\mathbb{E}[\mathsf{Tr}(\Sigma_\tau^2(x_\tau))],L^2\right\},
\end{align}
where $C_0$ is a universal constant.
\end{lemma}

By using Lemma \ref{lem:bound-s-J}, we have
\begin{align}\label{eq:proof-cal-deriv-1}
&\quad \mathbb{E}_{x_{\tau} \sim X_\tau}\left\|\frac{\partial}{\partial y}\frac{ s_{\tau}^{\star}(\sqrt{1-\tau}y)}{(1-\tau)^{3/2}}\Big|_{y = \frac{x_{\tau}}{\sqrt{1-\tau}}}\frac{\partial }{\partial \tau}\frac{x_\tau}{\sqrt{1-\tau}} \right\|_2^2\nonumber\\
&\le
\mathbb{E}_{x_{\tau} \sim X_\tau}\Big\|J_\tau(x_{\tau})\frac{s_\tau^{\star}(x_\tau)}{(1-\tau)^{5/2}}\Big\|_2^2 \lesssim \frac{d\log T}{\tau^3(1-\tau)^5}\min\left\{d+\mathbb{E}[\mathsf{Tr}(\Sigma_\tau^2(x_\tau))],L^2\right\}.
\end{align}

For the second term, we have
\begin{align*}
\frac{\partial}{\partial \tau}\frac{s_{\tau}^{\star}(\sqrt{1-\tau}y)}{(1-\tau)^{3/2}} 
&= -\frac{\partial}{\partial \tau}\frac{1}{(1-\tau)\tau}\int_{x_0}(y-x_0)p_{X_0|X_\tau}(x_0|\sqrt{1-\tau}y)\mathrm d x_0\nonumber\\
&=\frac{1-2\tau}{(1-\tau)^2\tau^2}\int_{x_0}(y-x_0)p_{X_0|X_\tau}(x_0|\sqrt{1-\tau}y)\mathrm d x_0\nonumber\\
&\quad- \frac{1}{(1-\tau)\tau}\frac{\partial}{\partial \tau}\int_{x_0}(y-x_0)p_{X_0|X_\tau}(x_0|\sqrt{1-\tau}y)\mathrm d x_0.
\end{align*}
Thus we have
\begin{align}\label{eq:proof-cal-deriv-2}
&\quad\mathbb{E}_{x_{\tau} \sim X_\tau}\left\|\frac{\partial}{\partial \tau}\frac{s_{\tau}^{\star}(\sqrt{1-\tau}y)}{(1-\tau)^{3/2}}\right\|_2^2 \notag\\
&\le \frac{2(1-2\tau)^2}{(1-\tau)^4\tau^4}\int_{x_0}\mathbb{E}_{y\sim X_\tau/\sqrt{1-\tau}}\left\|y-x_0\right\|_2^2p_{X_0|X_\tau}(x_0|\sqrt{1-\tau}y)\mathrm d x_0 \nonumber\\
&\quad+ \frac{2}{(1-\tau)^2\tau^2}\mathbb{E}\left\|\frac{\partial}{\partial \tau}\int_{x_0}(y-x_0)p_{X_0|X_\tau}(x_0|\sqrt{1-\tau}y)\mathrm d x_0\right\|_2^2\nonumber\\
&\overset{(i)}{\le} \frac{2(1-2\tau)^2}{(1-\tau)^4\tau^4}\frac{\tau d}{1-\tau}
+ \frac{2}{(1-\tau)^2\tau^2} \frac{d\min\{d\log T,L^2\}}{\tau(1-\tau)^3}\nonumber\\
&\lesssim \frac{(1-2\tau)^2d}{(1-\tau)^5\tau^3}
+ \frac{d\min\{d\log T,L^2\}}{(1-\tau)^5\tau^3}\lesssim \frac{d\min\{d\log T,L^2\}}{(1-\tau)^5\tau^3},
\end{align}
where (i) comes from the following Lemma.
\begin{lemma}\label{lem:bound-der-tau}
For $y\sim p_{X_\tau/\sqrt{1-\tau}}$, the following inequality holds.
\begin{align*}
\mathbb{E}\left\|\frac{\partial}{\partial \tau}\int_{x_0}(y-x_0)p_{X_0|X_\tau}(x_0|\sqrt{1-\tau}y)\mathrm d x_0\right\|_2^2
\le \frac{d\min\{d\log T,L\}}{\tau(1-\tau)^3}.
\end{align*}
\end{lemma}
%
% fact that
% \begin{align*}
% &\quad\mathbb{E}\left\|\frac{\partial}{\partial \tau}\int_{x_0}(y-x_0)p_{X_0|X_\tau}(x_0|\sqrt{1-\tau}y)\mathrm d x_0\right\|_2^2\nonumber\\
% &\le\mathbb{E}\left\|\frac{\partial}{\partial \frac{\tau}{1-\tau}}\int_{x_0}(y-x_0)p_{X_0|X_\tau}(x_0|\sqrt{1-\tau}y)\mathrm d x_0\right\|_2^2\left|\frac{\partial}{\partial \tau}\frac{\tau}{1-\tau}\right|^2\nonumber\\
% &\le \mathbb{E}\lim\inf_{\Delta\to0}\frac{1}{\Delta^2}W_1^2\left(\widetilde{P}_{0|t+\Delta}(\cdot|x_\tau),\widetilde{P}_{0|t}(\cdot|x_\tau)\right)\frac{1}{(1-\tau)^4} \nonumber\\
% &\overset{\text{(a)}}{\le} \frac{d(1-\tau)}{\tau}\frac{1}{(1-\tau)^4},
% \end{align*}
% where $\widetilde{P}_{0|t}$ denotes the distribution of $X_0$ conditioned on $X_0 + \sqrt{t}Z$, $t=\frac{\tau}{1-\tau}$, and (a) is derived in a similar way as \citet{Chen2023The}.
% Moreover, we have
% \begin{align*}
% \mathbb{E}_{x_{\tau} \sim X_\tau}\left\|\frac{\partial}{\partial \tau}\frac{s_{\tau}^{\star}(\sqrt{1-\tau}y)}{(1-\tau)^{3/2}}\right\|^2 \lesssim \frac{L^2d}{(1-\tau)^5}\left(\frac{1}{\tau}\vee L\right).
% \end{align*}
Inserting \eqref{eq:proof-cal-deriv-1} and \eqref{eq:proof-cal-deriv-2} into \eqref{eq:proof-cal-deriv}, we have
\begin{align*}
\mathbb{E}_{x_\tau\sim X_\tau}\left\|\frac{\partial}{\partial \tau}\frac{s_{\tau}^{\star}(x_{\tau})}{(1-\tau)^{3/2}}\right\|^2\lesssim \frac{d\log T}{\tau^3(1-\tau)^5}\min\left\{d+\mathbb{E}[\mathsf{Tr}(\Sigma_\tau^2(x_\tau))],L^2\right\}.
\end{align*}

\noindent \textbf{Proof of Lemma \ref{lem:bound-s-J}.}
Inequality \eqref{eq:bound-s} can be proved in a similar way as \citet{li2024d} (cf. Lemma 1).
Specifically, we have
\begin{align*}
&\quad \int\left\|x_\tau - \sqrt{1-\tau}x_0\right\|_2^2p_{X_0\mymid X_\tau}(x_0\mymid x_\tau)\mathrm{d} x_0\nonumber\\
&\le \int\left\|x_\tau - \sqrt{1-\tau}x_0\right\|_2^2p_{X_0\mymid X_\tau}(x_0\mymid x_\tau)\mathds{1}(\left\|x_\tau - \sqrt{1-\tau}x_0\right\|_2\le R)\mathrm{d} x_0\nonumber\\
&\quad+ \int\left\|x_\tau - \sqrt{1-\tau}x_0\right\|_2^2p_{X_0\mymid X_\tau}(x_0\mymid x_\tau)\mathds{1}(\left\|x_\tau - \sqrt{1-\tau}x_0\right\|_2> R)\mathrm{d} x_0\nonumber\\
&\overset{\text{(a)}}{\le} R^2 + \int p_{X_0}(x_0)\exp\left(-\frac{\|x_\tau-\sqrt{1-\tau}x_0\|_2^2}{3\tau}\right)\left\|x_\tau - \sqrt{1-\tau}x_0\right\|_2^2\notag\\
&\qquad \cdot\mathds{1}(\left\|x_\tau - \sqrt{1-\tau}x_0\right\|_2> R)\mathrm{d} x_0\nonumber\\
&\overset{\text{(b)}}{\le} R^2 + 3\tau\int p_{X_0}(x_0)\frac{R^2}{3\tau}\exp\left(-\frac{R^2}{3\tau}\right)\mathrm{d} x_0 \le 2R^2,
\end{align*}
where $R^2 = (6\theta+3c_0)d\tau\log T$,
(a) uses the fact that $p_{X_0\mymid X_\tau}(x_0\mymid x_\tau)\le p_{X_0}(x_0)\exp(-\frac{\|x_\tau - \sqrt{1-\tau}x_0\|_2^2}{3\tau})$ for $\left\|x_\tau - \sqrt{1-\tau}x_0\right\|_2> R$, and $z\exp(-z)\le \frac{R^2}{3\tau}\exp(-\frac{R^2}{3\tau})$ for $z\ge \frac{R^2}{3\tau}\ge 1$, which is satisfied by $(2\theta + c_0)d\log T \ge 1$. 
The remaining proof focuses on \eqref{eq:bound-J}.
For convenience, we denote $p_\tau$ as the probability density function of $X_\tau$ in this section. 
For $d <L^2$, we have
\begin{align}\label{eq:proof-Js}
&\quad \mathbb{E}_{x_{\tau} \sim X_\tau}\Big\|J_\tau(x_{\tau})s_\tau^{\star}(x_\tau)\Big\|_2^2 \notag\\
&\le 
\int_{\mathcal{S}_\tau} \Big\|J_\tau(x_{\tau})s_\tau^{\star}(x_\tau)\Big\|_2^2 p_\tau(x_\tau) \mathrm d x_\tau
+\int_{\mathcal{S}_\tau^{\rm c}} \Big\|J_\tau(x_{\tau})s_\tau^{\star}(x_\tau)\Big\|_2^2 p_\tau(x_\tau) \mathrm d x_\tau\nonumber\\
&\le 
\frac{25(\theta+c_0)d\log T}{\tau}\int\Big\|J_\tau(x_{\tau})\Big\|_2^2 p_\tau(x_\tau) \mathrm d x_\tau  +\int_{\mathcal{S}_\tau^{\rm c}} \Big\|J_\tau(x_{\tau})s_\tau^{\star}(x_\tau)\Big\|_2^2 p_\tau(x_\tau) \mathrm d x_\tau.
%&\lesssim \frac{(\theta+c_0)\min\{d\log T,L^2\}d\log T}{\tau^3} +\int_{\mathcal{S}_\tau^{\rm c}} \Big\|J_\tau(x_{\tau})s_\tau^{\star}(x_\tau)\Big\|^2 p_\tau(x_\tau) \mathrm d x_\tau,
\end{align}
Noticing that
\begin{align*}
J_\tau(x_\tau) = -\frac{1}{\tau}I_d + \frac{1}{\tau}\Sigma_\tau(x_\tau),
\end{align*}
we have
\begin{align*}
\left\|J_\tau(x_\tau)\right\|_2^2
&\le\left\|J_\tau(x_\tau)\right\|_F^2 = \mathsf{Tr}(J_\tau^2(x_\tau))
=\frac{1}{\tau^2}\mathsf{Tr}(I_d-2\Sigma_\tau(x_\tau) + \Sigma_\tau^2(x_\tau))\nonumber\\
&\le \frac{1}{\tau^2}\mathsf{Tr}(I_d+ \Sigma_\tau^2(x_\tau))\le \frac{d+\mathsf{Tr}( \Sigma_\tau^2(x_\tau))}{\tau^2}.
\end{align*}
Thus we have
\begin{align}\label{eq:proof-Js-temp-1}
\int \Big\|J_\tau(x_{\tau})\Big\|_2^2 p_\tau(x_\tau) \mathrm d x_\tau \le \frac{d+\mathbb{E}[\mathsf{Tr}( \Sigma_\tau^2(x_\tau))]}{\tau^2}.
\end{align}

To bound the second term, we notice that 
% \begin{align*}
% J_\tau(x_\tau) = -\frac{1}{\tau}I_d + \mathsf{Cov}\left[\frac{1}{\tau}(x-\sqrt{1-\tau}x_0)\right],
% \end{align*}
% where $\mathsf{Cov}[\cdot]$ denotes the covariance matrix under the conditional distribution $p_{X_0|X_\tau}(\cdot|x_\tau)$.
% Thus we have
\begin{align*}
\Big\|J_\tau(x_{\tau})s_\tau^{\star}(x_\tau)\Big\|_2 &\le \frac{1}{\tau^{3/2}}\left(\Big\|\sqrt{\tau}s_\tau^{\star}(x_\tau)\Big\|_2 + \Big\|\Sigma_{\tau}(x_\tau)\sqrt{\tau}s_\tau^{\star}(x_\tau)\Big\|_2 \right)\nonumber\\
& \overset{}{\le}\frac{1}{\tau^{3/2}}\left(\Big\|\sqrt{\tau}s_\tau^{\star}(x_\tau)\Big\|_2 + \|\Sigma_{\tau}(x_\tau)\|_2\Big\|\sqrt{\tau}s_\tau^{\star}(x_\tau)\Big\|_2 \right)\nonumber\\
&\le\frac{\Big\|\sqrt{\tau}s_\tau^{\star}(x_\tau)\Big\|_2}{\tau^{3/2}}\left(1 + \|\Sigma_\tau(x_\tau)\|_2 \right).
\end{align*}
%where (i) uses the fact that $\|\Sigma_{\tau}(x_\tau)\| \le \mathbb{E}_{x_0}\left[\|\frac{1}{\sqrt{\tau}}(x-\sqrt{1-\tau}x_0)\|^2\mymid x\right]$.
Then according to Jensen's inequality, we have
\begin{align*}
&\quad \int_{\mathcal{S}_\tau^{\rm c}} \Big\|J_\tau(x_{\tau})s_\tau^{\star}(x_\tau)\Big\|_2^2 p_\tau(x_\tau) \mathrm d x_\tau 
\le \frac{1}{\tau^3}\int_{\mathcal{S}_\tau^{\rm c}} \Big\|\sqrt{\tau}s_\tau^{\star}(x_\tau)\Big\|_2^2\left(1 + \|\Sigma_\tau(x_\tau)\|_2\right)^2 p_\tau(x_\tau) \mathrm d x_\tau\nonumber\\
&\lesssim \frac{1}{\tau^3}\left(\int \|\sqrt{\tau}s_\tau^{\star}(x_\tau)\|_2^4\left(1 + \|\Sigma_{\tau}(x_\tau)\|_2\right)^4 p_\tau(x_\tau) \mathrm d x_\tau\right)^{1/2}\left(\int_{\mathcal{S}_\tau^{\rm c}}p_\tau(x_\tau) \mathrm d x_\tau\right)^{1/2}. 
\end{align*}

Noticing that
\begin{align*}
\left\|s_\tau^{\star}(x_\tau)\right\|_2 &\le \frac{1}{\tau}\mathbb{E}_{x_0}\left[\left\|x_\tau-\sqrt{1-\tau}x_0\right\|_2\mymid x_\tau\right],\nonumber\\
\|\Sigma_{\tau}(x_\tau)\|_2 &\le \frac{1}{\tau}\mathbb{E}_{x_0}\left[\left\|x_\tau-\sqrt{1-\tau}x_0\right\|_2^2\mymid x_\tau\right],
\end{align*}
we have
\begin{align*}
&\quad \int \|\sqrt{\tau}s_\tau^{\star}(x_\tau)\|_2^4\left(1 + \|\Sigma_\tau(x_\tau)\|_2\right)^4 p_\tau(x_\tau) \mathrm d x_\tau\nonumber\\
&\lesssim \int \frac{1}{\tau^2}\mathbb{E}_{x_0}\left[\left\|x_\tau-\sqrt{1-\tau}x_0\right\|_2^4\mymid x_\tau\right] \left(1 + \frac{1}{\tau^4}\mathbb{E}_{x_0}\left[\left\|x_\tau-\sqrt{1-\tau}x_0\right\|_2^8\mymid x_\tau\right]\right) p_\tau(x_\tau) \mathrm d x_\tau\nonumber\\
&\lesssim \int \frac{1}{\tau^2}\left\|x_\tau-\sqrt{1-\tau}x_0\right\|_2^4p_{X_0|X_\tau}(x_0\mymid x_\tau) p_\tau(x_\tau) \mathrm d x_0 \mathrm d x_\tau\nonumber\\
&\quad + \int\frac{1}{\tau^6}\|x_\tau-\sqrt{1-\tau}x_0\|_2^{12} p_{X_0|X_\tau}(x_0\mymid x_\tau)p_\tau(x_\tau) \mathrm d x_0 \mathrm d x_\tau\lesssim d^6.
\end{align*}

Moreover, we have
\begin{align*}
\int_{\mathcal{S}_\tau^{\rm c}}p_\tau(x_\tau) \mathrm d x_\tau 
&\le \int_{\mathcal{S}_\tau^{\rm c}\cap\{x_\tau:\|x_\tau\|\le \sqrt{1-\tau}T^{c_R+4} + \sqrt{\tau d\log T}\}} p_\tau \mathrm d x_\tau \notag\\
&\quad+ \int_{\{x_\tau:\|x_\tau\|> \sqrt{1-\tau}T^{c_R+4} + \sqrt{\tau d\log T}\}} p_\tau \mathrm d x_\tau \nonumber\\
&\le\left(2\sqrt{1-\tau}T^{c_R+4} + 2\sqrt{\tau d\log T}\right)^d \exp(-\theta d\log T)\nonumber\\
&\quad + \mathbb{P}\left(\|x_\tau\|> \sqrt{1-\tau}T^{c_R+4} + \sqrt{\tau d\log T}\right)\nonumber\\
&\le \exp(-(\theta-c_R-6) d\log T) +  \mathbb{P}\left(\|X_0\|> T^{c_R+4}\right) \notag\\
&\quad + P\left(\left\|W_\tau\right\| \ge \sqrt{d\log T}\right)\nonumber\\
&\le \frac{1}{T^4} + \frac{\mathbb{E}\|X_0\|}{T^{c_R+4}} + \frac{1}{T^4} \lesssim \frac{1}{T^4},
\end{align*}
as long as $\theta\ge c_R+10$.
Thus we have
\begin{align}\label{eq:proof-Js-temp-2}
\int_{\mathcal{S}_\tau^{\rm c}} \Big\|J_\tau(x_{\tau})s_\tau^{\star}(x_\tau)\Big\|^2 p_\tau(x_\tau) \mathrm d x_\tau 
\le  \frac{d^3}{\tau^3T^2}\lesssim \frac{d^2}{\tau^3}
\end{align}
for $T\ge K\gtrsim \sqrt{d}$.
Inserting \eqref{eq:proof-Js-temp-1} and \eqref{eq:proof-Js-temp-2} into \eqref{eq:proof-Js}, we have
\begin{align}\label{eq:proof-temp-9}
\mathbb{E}_{x_{\tau} \sim X_\tau}\Big\|J_\tau(x_{\tau})s_\tau^{\star}(x_\tau)\Big\|^2 
&\lesssim \frac{d\log T}{\tau^3}\left(d+ \mathbb{E}[\mathsf{Tr}(\Sigma_\tau(x_\tau))]\right) + \frac{d^2}{\tau^3}\nonumber\\
&\lesssim\frac{d\log T}{\tau^3}\left(d+ \mathbb{E}[\mathsf{Tr}(\Sigma_\tau(x_\tau))] \right).
\end{align}

Otherwise, for $d\ge L^2$, we have
\begin{align*}
\mathbb{E}_{x_{\tau} \sim X_\tau}\Big\|J_\tau(x_{\tau})s_\tau^{\star}(x_\tau)\Big\|_2^2 
&\le \frac{L^2}{\tau^2}\int_{\mathcal{L}_\tau} \Big\|s_\tau^{\star}(x_\tau)\Big\|_2^2 p_\tau(x_\tau) \mathrm d x_\tau\notag\\
&\qquad +  \int_{\mathcal{L}_\tau^{\rm c}} \Big\|J_\tau(x_\tau)s_\tau^{\star}(x_\tau)\Big\|_2^2 p_\tau(x_\tau) \mathrm d x_\tau.
\end{align*}
For the first term, we have
\begin{align*}
\int_{\mathcal{L}_\tau} \Big\|s_\tau^{\star}(x_\tau)\Big\|_2^2 p_\tau(x_\tau) \mathrm d x_\tau&\le \frac{1}{\tau^2}\int\left\|x_\tau-\sqrt{1-\tau}x_0\right\|_2^2 p_{X_0\mymid X_{\tau}}(x_0\mymid x_\tau) p_\tau(x_\tau) \mathrm d x_0 \mathrm d x_\tau \le \frac{d}{\tau}.
\end{align*}
For the second term, we have
\begin{align*}
&\quad\int_{\mathcal{L}_\tau^{\rm c}} \Big\|J_\tau(x_\tau)s_\tau^{\star}(x_\tau)\Big\|_2^2 p_\tau(x_\tau) \mathrm d x_\tau\notag\\
&\le \left(\int \Big\|J_\tau(x_\tau)s_\tau^{\star}(x_\tau)\Big\|_2^4 p_\tau(x_\tau) \mathrm d x_\tau\right)^{1/2}
\left(\int_{\mathcal{L}_\tau^{\rm c}}p_\tau(x_\tau) \mathrm d x_\tau\right)^{1/2}
\lesssim \frac{d^3}{\tau^3}\frac{1}{d^2}
\lesssim \frac{d}{\tau^3}.
\end{align*}
Thus we have
\begin{align}\label{eq:proof-temp-10}
\mathbb{E}_{x_{\tau} \sim X_\tau}\Big\|J_\tau(x_{\tau})s_\tau^{\star}(x_\tau)\Big\|_2^2 
&\lesssim \frac{dL^2}{\tau^3} + \frac{d}{\tau^3}\lesssim \frac{dL^2}{\tau^3}.
\end{align}
Combining \eqref{eq:proof-temp-9} and \eqref{eq:proof-temp-10}, we could complete the proof. 

\noindent \textbf{Proof of Lemma \ref{lem:bound-der-tau}.}
According to the definition, we have
\begin{align*}
&\quad \int_{x_0}(y-x_0)p_{X_0|X_\tau}(x_0|\sqrt{1-\tau}y)\mathrm d x_0\notag\\
&= \frac{\int(y-x_0)\phi(\sqrt{1-\tau}y|\sqrt{1-\tau}x_0,\tau I_d)p_{X_0}(x_0)\mathrm d x_0}{\int\phi(\sqrt{1-\tau}y|\sqrt{1-\tau}x_0,\tau I_d)p_{X_0}(x_0)\mathrm d x_0}\nonumber\\
&=\frac{\int(y-x_0)\phi(y|x_0,\frac{\tau}{1-\tau} I_d)p_{X_0}(x_0)\mathrm d x_0}{\int\phi(y|x_0,\frac{\tau}{1-\tau} I_d)p_{X_0}(x_0)\mathrm d x_0},
\end{align*}
where $\phi(\cdot|\mu,\Sigma)$ denotes the probability density function of Gaussian distribution with mean vector $\mu$ and covariance matrix $\Sigma$.
Define $t=\frac{\tau}{1-\tau}$.
We have
\begin{align*}
&\quad \frac{\partial}{\partial \tau}\int_{x_0}(y-x_0)p_{X_0|X_\tau}(x_0|\sqrt{1-\tau}y)\mathrm d x_0 \notag\\
&=\frac{\partial}{\partial t}\frac{\int(y-x_0)\phi(y|x_0,t I_d)p_{X_0}(x_0)\mathrm d x_0}{\int\phi(y|x_0, t I_d)p_{X_0}(x_0)\mathrm d x_0} \frac{\partial \frac{\tau}{1-\tau}}{\partial \tau}\nonumber\\
&= -\frac{1}{(1-\tau)^2}\Big(\frac{\int(y-x_0)\frac{\|y-x\|_2^2}{2t^2}\phi(y|x_0, t I_d)p_{X_0}(x_0)\mathrm d x_0}{\int\phi(y|x_0, t I_d)p_{X_0}(x_0)\mathrm d x_0}\nonumber\\
&\quad - \frac{\int(y-x_0)\phi(y|x_0, t I_d)p_{X_0}(x_0)\mathrm d x_0}{\int\phi(y|x_0, t I_d)p_{X_0}(x_0)\mathrm d x_0}\frac{\int\frac{\|y-x\|_2^2}{2t^2}\phi(y|x_0, t I_d)p_{X_0}(x_0)\mathrm d x_0}{\int\phi(y|x_0, t I_d)p_{X_0}(x_0)\mathrm d x_0}\Big)\nonumber\\
&=-\frac{1}{(1-\tau)^2}\left(\frac{\tau}{1-\tau}\right)^{3/2}\frac{1}{2t^2}\left(\mathbb{E}_{Z|y}\left[\|Z\|_2^2Z-\mathbb{E}_{Z|y}Z\mathbb{E}_{Z|y}\|Z\|_2^2\right] \right),
\end{align*}
where $\mathbb{E}_{Z|y}[\cdot]$ denotes the expectation conditioned on $X_0 + \sqrt{\frac{\tau}{1-\tau}}Z = y$.
Notice that
\begin{align*}
\mathbb{E}_{Z|y}\left[\|Z\|_2^2Z-\mathbb{E}_{Z|y}Z\mathbb{E}_{Z|y}\|Z\|_2^2\right] = \mathbb{E}_{Z|y}\left[\left(Z-\mathbb{E}_{Z|y}Z\right)\left(\|Z\|_2^2 - \mathbb{E}_{Z|y}\|Z\|_2^2\right)\right].
\end{align*}
According to Cauchy-Switch inequality, we have
\begin{align}
&\quad \left\|\mathbb{E}_{Z|y}\left[\left(Z-\mathbb{E}_{Z|y}Z\right)\left(\|Z\|_2^2 - \mathbb{E}_{Z|y}\|Z\|_2^2\right)\right]\right\|^2\notag\\
&=\max_{u\in\mathbb{R}^d,\|u\|=1} \left\|\mathbb{E}_{Z|y}\left[u^{\top}\left(Z-\mathbb{E}_{Z|y}Z\right)\left(\|Z\|_2^2 - \mathbb{E}_{Z|y}\|Z\|_2^2\right)\right]\right\|^2\nonumber\\
&\le\max_{u\in\mathbb{R}^d,\|u\|=1} \mathbb{E}_{Z|y}\left|u^{\top}\left(Z-\mathbb{E}_{Z|y}Z\right)\right|^2\mathbb{E}_{Z|y}\left[\left(\|Z\|_2^2 - \mathbb{E}_{Z|y}\|Z\|_2^2\right)^2\right].
\end{align}
Notice that according to the definition of Jacobian matrix $J_\tau$, the first factor is bounded by
\begin{align*}
\mathbb{E}_{Z|y}\left|u^{\top}\left(Z-\mathbb{E}_{Z|y}Z\right)\right|^2
= u^{\top}\mathbb{E}_{Z|y}\left(Z-\mathbb{E}_{Z|y}Z\right)\left(Z-\mathbb{E}_{Z|y}Z\right)^{\top}u\nonumber\\
=u^{\top}\left(I_d + \tau J_\tau(X_\tau)\right)u\le \left\|I_d + \tau J_\tau(X_\tau)\right\|_2.
\end{align*}
For the second factor, we have
\begin{align}\label{eq:proof-temp-11}
&\quad\int\mathbb{E}_{Z|\frac{x}{\sqrt{1-\tau}}}\left(\|Z\|_2^2 - \mathbb{E}_{Z|\frac{x}{\sqrt{1-\tau}}}\|Z\|_2^2\right)^2p_{X_\tau}(x)\mathrm{d}x\nonumber\\
%&\le \int\mathbb{E}_{Z|\frac{x}{\sqrt{1-\tau}}}\left(\|Z\|_2^2 - \mathbb{E}_{Z|\frac{x}{\sqrt{1-\tau}}}\|Z\|_2^2\right)^2p_{X_\tau}(x)\mathrm{d}x\nonumber\\
&\le \int\left(\mathbb{E}_{Z|\frac{x}{\sqrt{1-\tau}}}\|Z\|_2^4 - \left(\mathbb{E}_{Z|\frac{x}{\sqrt{1-\tau}}}\|Z\|_2^2\right)^2\right)p_{X_\tau}(x)\mathrm{d}x\nonumber\\
&=\int\|z\|_2^4p_Z(z)\mathrm{d} z - \frac{1}{\tau^2}\int\left(\int\|x-\sqrt{1-\tau}x_0\|_2^2p_{X_0|X_\tau}(x_0|x)\mathrm{d}z\right)^2p_{X_\tau}(x)\mathrm{d}x\nonumber\\
&\overset{\text{(a)}}{\le}
\int\|z\|_2^4p_Z(z)\mathrm{d} z - \left(\int\|z\|_2^2p_Z(z)\mathrm{d} z\right)^2\nonumber\\
&=\left(d^2+2d-d^2\right)\lesssim d,
\end{align}
where (a) uses the fact that $\mathbb{E}\left(\mathbb{E}_{Z|\frac{x}{\sqrt{1-\tau}}}\|Z\|_2^2\right)^2\ge \left(\mathbb{E}\mathbb{E}_{Z|\frac{x}{\sqrt{1-\tau}}}\|Z\|_2^2\right)^2$.

In addition, for each set $\mathcal{A}_\tau$ holding with probability at least $1-O(1/d^4)$, we have
\begin{align}\label{eq:proof-temp-12}
&\quad\int_{\mathcal{A}_\tau^{\rm c}}\left(\left\|I_d+\tau J_\tau(x)\right\|_2\right)\mathbb{E}_{Z|\frac{x}{\sqrt{1-\tau}}}\left(\|Z\|_2^2 - \mathbb{E}_{Z|\frac{x}{\sqrt{1-\tau}}}\|Z\|_2^2\right)^2p_{X_\tau}(x)\mathrm{d}x\nonumber\\
&\lesssim \int_{\mathcal{A}_\tau^{\rm c}}\mathbb{E}_{Z|\frac{x}{\sqrt{1-\tau}}}\left\|Z-\mathbb{E}_{Z|\frac{x}{\sqrt{1-\tau}}}Z\right\|_2^2\mathbb{E}_{Z|\frac{x}{\sqrt{1-\tau}}}\left(\|Z\|_2^2 - \mathbb{E}_{Z|\frac{x}{\sqrt{1-\tau}}}\|Z\|_2^2\right)^2p_{X_\tau}(x)\mathrm{d}x\nonumber\\
&\lesssim \int_{\mathcal{A}_\tau^{\rm c}}\mathbb{E}_{Z|\frac{x}{\sqrt{1-\tau}}}\left\|Z\right\|_2^2\mathbb{E}_{Z|\frac{x}{\sqrt{1-\tau}}}\|Z\|_2^4p_{X_\tau}(x)\mathrm{d}x\nonumber\\
&\lesssim \left(\int\mathbb{E}_{Z|\frac{x}{\sqrt{1-\tau}}}\left\|Z\right\|_2^4\mathbb{E}_{Z|\frac{x}{\sqrt{1-\tau}}}\|Z\|_2^8p_{X_\tau}(x)\mathrm{d}x\right)^{1/2}\left(\int_{\mathcal{A}_\tau^{\rm c}}p_{X_\tau}(x)\mathrm{d}x\right)^{1/2}\nonumber\\
&\lesssim \frac{1}{d^2}\left(\mathbb{E}\|Z\|_2^8 \mathbb{E}\|Z\|_2^{16}\right)^{1/4}
\lesssim d.
\end{align}

Armed with the above observations, we decompose the expectation as 
\begin{align}\label{eq:bound-der-tau-1}
&\quad\mathbb{E}\left\|\frac{\partial}{\partial \tau}\int_{x_0}(y-x_0)p_{X_0|X_\tau}(x_0|\sqrt{1-\tau}y)\mathrm d x_0\right\|_2^2\notag\\
&\le \frac{1}{\tau(1-\tau)^3}\mathbb{E}\left(\left\|I_d+\tau J_\tau(X_\tau)\right\|_2\mathbb{E}_{Z|y}\left(\|Z\|_2^2 - \mathbb{E}_{Z|y}\|Z\|_2^2\right)^2\right)\nonumber\\
&=\frac{1}{\tau(1-\tau)^3}\int_{\mathcal{L}_\tau}\left\|I_d+\tau J_\tau(x)\right\|_2\mathbb{E}_{Z|\frac{x}{\sqrt{1-\tau}}}\left(\|Z\|_2^2 - \mathbb{E}_{Z|\frac{x}{\sqrt{1-\tau}}}\|Z\|_2^2\right)^2p_{X_\tau}(x)\mathrm{d}x\nonumber\\
&\quad +\frac{1}{\tau(1-\tau)^3}\int_{\mathcal{L}_\tau^{\rm c}}\left\|I_d+\tau J_\tau(x)\right\|_2\mathbb{E}_{Z|\frac{x}{\sqrt{1-\tau}}}\left(\|Z\|_2^2 - \mathbb{E}_{Z|\frac{x}{\sqrt{1-\tau}}}\|Z\|_2^2\right)^2p_{X_\tau}(x)\mathrm{d}x\nonumber\\
&\overset{\text{(a)}}{\le} \frac{1+L}{\tau(1-\tau)^3}\int_{\mathcal{L}_\tau}\mathbb{E}_{Z|\frac{x}{\sqrt{1-\tau}}}\left(\|Z\|_2^2 - \mathbb{E}_{Z|\frac{x}{\sqrt{1-\tau}}}\|Z\|_2^2\right)^2p_{X_\tau}(x)\mathrm{d}x+\frac{d}{\tau(1-\tau)^3}\nonumber\\
&\overset{\text{(b)}}{\le} \frac{dL}{\tau(1-\tau)^3},
\end{align}
where (a) uses the fact that $\mathbb{P}(X_\tau\in\mathcal{L}_\tau^{\rm c})\lesssim 1/d^4$ and \eqref{eq:proof-temp-12},
and (b) uses \eqref{eq:proof-temp-11}.

In addition, recalling that for $x_\tau\in\mathcal{S}_\tau$, we have
\begin{align*}
\|I_d + \tau J_\tau(x_\tau)\|_2 
&\le \mathbb{E}_{Z|\frac{x}{\sqrt{1-\tau}}}\left\|Z-\mathbb{E}_{Z|\frac{x}{\sqrt{1-\tau}}}Z\right\|_2^2
\le \mathbb{E}_{Z|\frac{x}{\sqrt{1-\tau}}}\left\|Z\right\|_2^2\nonumber\\
&=\frac{1}{\tau}\int \|x_\tau-\sqrt{1-\tau}x_0\|_2^2p_{X_0|X_\tau}(x_0|x_\tau)\mathrm{d} x_0
\overset{\text{(a)}}{\le} 25(\theta+c_0)d\log T,
\end{align*}
where (a) uses Lemma \ref{lem:bound-s-J}.
Thus for $d<L$, we have
\begin{align}\label{eq:bound-der-tau-2}
&\quad\mathbb{E}\left\|\frac{\partial}{\partial \tau}\int_{x_0}(y-x_0)p_{X_0|X_\tau}(x_0|\sqrt{1-\tau}y)\mathrm d x_0\right\|_2^2\nonumber\\
%\le \frac{1}{\tau(1-\tau)^3}\mathbb{E}\left(\left\|I_d + \tau J_\tau(X_\tau)\right\|_2\mathbb{E}_{Z|y}\left(\|Z\|_2^2 - \mathbb{E}_{Z|y}\|Z\|_2^2\right)^2\right)\nonumber\\
&=\frac{1}{\tau(1-\tau)^3}\int_{\mathcal{S}_\tau}\left\|I_d + \tau J_\tau(X_\tau)\right\|_2\mathbb{E}_{Z|\frac{x}{\sqrt{1-\tau}}}\left(\|Z\|_2^2 - \mathbb{E}_{Z|\frac{x}{\sqrt{1-\tau}}}\|Z\|_2^2\right)^2p_{X_\tau}(x)\mathrm{d}x\nonumber\\
&\quad +\frac{1}{\tau(1-\tau)^3}\int_{\mathcal{S}_\tau^{\rm c}}\left\|I_d + \tau J_\tau(X_\tau)\right\|_2\mathbb{E}_{Z|\frac{x}{\sqrt{1-\tau}}}\left(\|Z\|_2^2 - \mathbb{E}_{Z|\frac{x}{\sqrt{1-\tau}}}\|Z\|_2^2\right)^2p_{X_\tau}(x)\mathrm{d}x\nonumber\\
&\lesssim \frac{d\log T}{\tau(1-\tau)^3} \int_{\mathcal{S}_\tau}\mathbb{E}_{Z|\frac{x}{\sqrt{1-\tau}}}\left(\|Z\|_2^2 - \mathbb{E}_{Z|\frac{x}{\sqrt{1-\tau}}}\|Z\|_2^2\right)^2p_{X_\tau}(x)\mathrm{d}x + \frac{d}{\tau(1-\tau)^3} \nonumber\\
&\lesssim \frac{d^2\log T}{\tau(1-\tau)^3} + \frac{d}{\tau(1-\tau)^3} \lesssim \frac{d^2\log T}{\tau(1-\tau)^3},
\end{align}
where (a) uses \eqref{eq:proof-temp-12} and the fact the $\mathbb{P}(X_\tau\in\mathcal{S}_\tau^{\rm c})\lesssim 1/T^4\lesssim 1/d^4$ for $T\gtrsim K \gtrsim\min\{d\log T,L\}\log T\gtrsim d$, and (b) uses \eqref{eq:proof-temp-11}.
Combining \eqref{eq:bound-der-tau-1} and \eqref{eq:bound-der-tau-2}, we have
\begin{align*}
\mathbb{E}\left\|\frac{\partial}{\partial \tau}\int_{x_0}(y-x_0)p_{X_0|X_\tau}(x_0|\sqrt{1-\tau}y)\mathrm d x_0\right\|_2^2
\le \frac{d\min\{d\log T,L\}}{\tau(1-\tau)^3}.
\end{align*}
and complete the proof.

\subsection{Proof of Lemma~\ref{lem:KL}}
\label{sec:proof-lem:KL}

We introduce a more preliminary lemma which leads to Lemma \ref{lem:KL} immediately.
\begin{lemma}
\label{lem:KL-pre}
According to Lemma~\ref{lem:discrete}, it can be shown that for any $0\le n\le N$,
\begin{align} %\label{eq:KL-pre}
&\quad\|y_{\tau_{k,n}}(x_k)-x_{\tau_{k,n}}(x_k)\|_2^2
\le \zeta_{k,n}(x_k) + \frac{N\log^2 T}{T^2}\sum_{i=0}^{n-1}\widehat{\tau}_{k,i-1}\widetilde{\varepsilon}_{{k,i}}^2(x_k), x_k\in\mathcal{E}_k,\label{eq:lem-KL-pre-1}\\
&\quad\int_{{\mathcal{E}}_{k,n}} \zeta_{k,n}(x_k) p_{\widehat{X}_{k}}(x_k) \mathrm d x_k\lesssim \frac{d\log^5 T}{T^3}\min\Big\{\frac{Nd\widehat{\tau}_{k,-1}\log T}{T}\notag\\
&\qquad \qquad\qquad \qquad \qquad +(1-\widehat{\tau}_{k,n})\int_{{\tau}_{k,n}}^{{\tau}_{k,0}} \frac{\mathbb{E}[\mathsf{Tr}(\Sigma_\tau^2(x_\tau))]}{(1-\tau)^2}\mathrm d \tau,\frac{NL^2\widehat{\tau}_{k,-1}\log T}{T}\Big\},\label{eq:lem-KL-pre-2}
% \le
% &\quad\frac{1-\tau_{k+1, 0}}{2(\tau_{k+1, 0}-\tau_{k, N})}\int_{\mathcal{E}_k}\|y_{\tau_{k, N}}(x_k) - x_{\tau_{k, N}}(x_k)\|_2^2 p_{\widehat{X}_k}(x_k)\mathrm d x_k
% \nonumber\\
% &\lesssim \frac{\min\left\{d\log T, L^2\right\}d\log^5 T}{T^3}
% + \frac{\log T}{T}\sum_{i = 0}^{N-1}\varepsilon_{k,i}^2,
\end{align}
where $\widetilde{\varepsilon}_{k,i}^2(x_k)$ is defined in Lemma \ref{lem:discrete-pre}, and 
\begin{small}
\begin{align*}
\mathcal{E}_{k,n}: = \left\{
\begin{array}{ll}
   \{x_{\tau_{k,0}}: x_{\tau_{k,i}}(x_{\tau_{k,0}})\in\widetilde{\mathcal{S}}_{\tau_{k,i}}\cap \mathcal{L}_{\tau_{k,i}}, y_{\tau_{k,i}}(x_{\tau_{k,0}})\in\mathcal{S}_{\tau_{k,i}}, \forall 0\le i \le n-1\},  & {\rm if}~L> d\log T, \\
   \emptyset  &  {\rm if}~L\le d\log T.
\end{array}
\right.
\end{align*}
\end{small}
\end{lemma}
According to the definition of $\varepsilon_{k,i}^2$ in~\eqref{eq:score-error}, we have
\begin{align}\label{eq:proof-temp-8}
&\quad \mathbb{E}_{x_{\tau_{k,0}}\sim p_{\widetilde{X}_k}}\left\|y_{\tau_{k, N}} - x_{\tau_{k, N}}\right\|_2^2\notag\\
&\le \int_{{\mathcal{E}}_{k}} \zeta_{k,N}(x_k) p_{\widetilde{X}_{k}}(x_k) \mathrm d x_k + \frac{N\log^2 T}{T^2}\sum_{i=0}^{N-1}\widehat{\tau}_{k,i-1}{\varepsilon}_{{k,i}}^2(x_k)\nonumber\\
&\lesssim\frac{d\log^5 T}{T^3}\min\Big\{\frac{Nd\widehat{\tau}_{k,-1}\log T}{T}+(1-\widehat{\tau}_{k,N})\int_{{\tau}_{k,n}}^{{\tau}_{k,0}} \frac{\mathbb{E}[\mathsf{Tr}(\Sigma_\tau^2(x_\tau))]}{(1-\tau)^2}\mathrm d \tau,\nonumber\\
&\qquad\qquad \qquad\frac{NL^2\widehat{\tau}_{k,-1}\log T}{T}\Big\}+\frac{\widehat{\tau}_{k,-1}N\log^2 T}{T^2}\sum_{i = 0}^{N-1} \varepsilon_{k,i}^2.
\end{align}
because $\mathcal{E}_{k}\subset\mathcal{E}_{k,n}$, $p_{\widetilde{X}_{k}}(x_k)\le p_{\widehat{X}_{k}}(x_k)$ ($x\neq\infty$), and 
$$
\int_{{\mathcal{E}}_{k}} \zeta_{k,n}(x_k) p_{\widetilde{X}_{k}}(x_k) \mathrm d x_k\le \int_{{\mathcal{E}}_{k,n}} \zeta_{k,n}(x_k) p_{\widehat{X}_{k}}(x_k) \mathrm d x_k.
$$
According to \eqref{eq:KL-condition}, we have that
%According to definitions, we have
\begin{align}\label{eq:proof-temp-3}
\mathsf{KL}\big(p_{\widehat{X}_{k+1}|\widehat{X}_{k}}\left(\,\cdot\mymid x_{\tau_{k,0}}\right)\,\Vert\,p_{Y_{k+1}|Y_{k}}\left(\,\cdot\mymid x_{\tau_{k,0}}\right)\big)
&= \frac{1-\tau_{k+1, 0}}{2(\tau_{k+1, 0}-\tau_{k, N})}\|y_{\tau_{k, N}} - x_{\tau_{k, N}}\|_2^2 \nonumber\\
&\lesssim \frac{T}{N\widehat{\tau}_{k, N}\log T}\|y_{\tau_{k, N}} - x_{\tau_{k, N}}\|_2^2.
\end{align}
Inserting \eqref{eq:proof-temp-8} into \eqref{eq:proof-temp-3},
we have
\begin{align}
&\quad \mathbb{E}_{x_{\tau_{k,0}}\sim p_{\widetilde{X}_{k}}}\left[\mathsf{KL}\left(p_{\widehat{X}_{k+1}|\widehat{X}_{k}}\left(\,\cdot\mymid x_{\tau_{k,0}}\right)\,\Vert\,p_{Y_{k+1}|Y_{k}}\left(\,\cdot\mymid x_{\tau_{k,0}}\right)\right)\right]\nonumber\\
&\lesssim \frac{T}{N\widehat{\tau}_{k, N}\log T}\mathbb{E}_{x_{\tau_{k,0}}\sim p_{\widetilde{X}_{k}}}\|y_{\tau_{k, N}} - x_{\tau_{k, N}}\|_2^2\nonumber\\
&\overset{\text{(a)}}{\lesssim} \frac{dK\log^4 T}{T^3}\min\Big\{\frac{Nd\log T}{T}+\frac{1-\widehat{\tau}_{k,N}}{\widehat{\tau}_{k,N}}\int_{{\tau}_{k,N}}^{{\tau}_{k,0}} \frac{\mathbb{E}[\mathsf{Tr}(\Sigma_\tau^2(x_\tau))]}{(1-\tau)^2}\mathrm d \tau,\frac{NL^2\log T}{T}\Big\}\notag\\
&\quad+\frac{KN\log T}{T^2}\sum_{i = 0}^{N-1} \varepsilon_{k,i}^2
\nonumber\\
&\overset{\text{(b)}}{\lesssim} \frac{d\log^4 T}{T^3}\min\Big\{d\log T+\frac{K(1-\widehat{\tau}_{k,N})}{\widehat{\tau}_{k,N}}\int_{{\tau}_{k,N}}^{{\tau}_{k,0}} \frac{\mathbb{E}[\mathsf{Tr}(\Sigma_\tau^2(x_\tau))]}{(1-\tau)^2}\mathrm d \tau,L^2\log T\Big\}
+\frac{\log T}{T}\sum_{i = 0}^{N-1} \varepsilon_{k,i}^2,\label{eq:proof-lemKL-temp-3}
\end{align}
where (a) uses \eqref{eq:diff-tau-2}, and (b) uses the fact that $KN\lesssim T$.
Notice that for any $0<\tau_1<\tau_2<1$,
\begin{align*}
\int_{\tau_1}^{\tau_2}\frac{\mathbb{E}[\mathsf{Tr}(\Sigma_\tau^2(x_\tau))]}{(1-\tau)^2}\mathrm{d}\tau
&=\frac{\tau_2}{1-\tau_2}\mathbb{E}[\mathsf{Tr}(\Sigma_{\tau_2}(x_{\tau_2}))]-\frac{\tau_1}{1-\tau_1}\mathbb{E}[\mathsf{Tr}(\Sigma_{\tau_1}(x_{\tau_1}))],
\end{align*}
which has been proved in \citet{li2024sharp} (cf. (90)).
We have
\begin{align*}
&\quad\frac{1-\widehat{\tau}_{k,N}}{\widehat{\tau}_{k,N}}\int_{{\tau}_{k,N}}^{{\tau}_{k,0}} \frac{\mathbb{E}[\mathsf{Tr}(\Sigma_\tau^2(x_\tau))]}{(1-\tau)^2}\mathrm d \tau\notag\\
&=\frac{\tau_{k,0}(1-\widehat{\tau}_{k,N})}{\widehat{\tau}_{k,N}(1-\tau_{k,0})}\mathbb{E}[\mathsf{Tr}(\Sigma_{\tau_{k,0}}(x_{\tau_{k,0}}))] -\frac{\tau_{k,N}(1-\widehat{\tau}_{k,N})}{\widehat{\tau}_{k,N}(1-\tau_{k,N})}\mathbb{E}[\mathsf{Tr}(\Sigma_{\tau_{k,N}}(x_{\tau_{k,N}}))].
\end{align*}
Furthermore, we have
\begin{align}
&\quad \sum_{k=0}^{K-1}\frac{1-\widehat{\tau}_{k,N}}{\widehat{\tau}_{k,N}}\int_{{\tau}_{k,N}}^{{\tau}_{k,0}} \frac{\mathbb{E}[\mathsf{Tr}(\Sigma_\tau^2(x_\tau))]}{(1-\tau)^2}\mathrm d \tau\nonumber\\
&\le \sum_{k=0}^{K-3}\frac{\tau_{k+2,0}(1-\widehat{\tau}_{k+2,N})}{\widehat{\tau}_{k+2,N}(1-\tau_{k+2,0})}\mathbb{E}[\mathsf{Tr}(\Sigma_{\tau_{k+2,0}}(x_{\tau_{k+2,0}}))]\nonumber\\
&\quad-\sum_{k=0}^{K-3}\frac{\tau_{k,N}(1-\widehat{\tau}_{k,N})}{\widehat{\tau}_{k,N}(1-\tau_{k,N})}\mathbb{E}[\mathsf{Tr}(\Sigma_{\tau_{k,N}}(x_{\tau_{k,N}}))]\nonumber\\
&\quad+\frac{\tau_{0,0}(1-\widehat{\tau}_{0,N})}{\widehat{\tau}_{0,N}(1-\tau_{0,0})}\mathbb{E}[\mathsf{Tr}(\Sigma_{\tau_{0,0}}(x_{\tau_{0,0}}))]+\frac{\tau_{1,0}(1-\widehat{\tau}_{1,N})}{\widehat{\tau}_{1,N}(1-\tau_{1,0})}\mathbb{E}[\mathsf{Tr}(\Sigma_{\tau_{1,0}}(x_{\tau_{1,0}}))]\nonumber\\
&=\sum_{k=0}^{K-3}\left(\frac{\tau_{k+2,0}(1-\widehat{\tau}_{k+2,N})}{\widehat{\tau}_{k+2,N}(1-\tau_{k+2,0})}-\frac{\tau_{k,N}(1-\widehat{\tau}_{k,N})}{\widehat{\tau}_{k,N}(1-\tau_{k,N})}\right)\mathbb{E}[\mathsf{Tr}(\Sigma_{\tau_{k+2,0}}(x_{\tau_{k+2,0}}))]\nonumber\\
&\quad+\frac{\tau_{0,0}(1-\widehat{\tau}_{0,N})}{\widehat{\tau}_{0,N}(1-\tau_{0,0})}\mathbb{E}[\mathsf{Tr}(\Sigma_{\tau_{0,0}}(x_{\tau_{0,0}}))]+\frac{\tau_{1,0}(1-\widehat{\tau}_{1,N})}{\widehat{\tau}_{1,N}(1-\tau_{1,0})}\mathbb{E}[\mathsf{Tr}(\Sigma_{\tau_{1,0}}(x_{\tau_{1,0}}))],\label{eq:proof-lemKL-temp-1}
\end{align}
where the last inequality uses the fact that $\tau_{k+2,0} = \tau_{k,N}$.
Notice that 
$$
\mathbb{E}[\Sigma_\tau(x_\tau)] = \mathbb{E}[\mathsf{Cov}(Z\mymid \sqrt{1-\tau}X_0 + \sqrt{\tau}Z = x_\tau)]\preceq \mathsf{Cov}[Z] = I_d,
$$
and by using \eqref{eq:diff-tau-2},
we have
$$
\frac{\tau_{0,0}(1-\widehat{\tau}_{0,N})}{\widehat{\tau}_{0,N}(1-\tau_{0,0})}\mathbb{E}[\mathsf{Tr}(\Sigma_{\tau_{0,0}}(x_{\tau_{0,0}}))] \lesssim d,\quad \frac{\tau_{1,0}(1-\widehat{\tau}_{1,N})}{\widehat{\tau}_{1,N}(1-\tau_{1,0})}\mathbb{E}[\mathsf{Tr}(\Sigma_{\tau_{1,0}}(x_{\tau_{1,0}}))]\lesssim d.
$$
Moreover, we have
\begin{align*}
&\quad\frac{\tau_{k+2,0}(1-\widehat{\tau}_{k+2,N})}{\widehat{\tau}_{k+2,N}(1-\tau_{k+2,0})}-\frac{\tau_{k,N}(1-\widehat{\tau}_{k,N})}{\widehat{\tau}_{k,N}(1-\tau_{k,N})}\notag\\
&=\frac{\tau_{k+2,0}(1-\widehat{\tau}_{k+2,N})}{\widehat{\tau}_{k+2,N}(1-\tau_{k+2,0})}-\frac{\tau_{k+2,0}(1-\widehat{\tau}_{k+2,0})}{\widehat{\tau}_{k+2,0}(1-\tau_{k+2,0})}\nonumber\\
&= \frac{\tau_{k+2,0}(\widehat{\tau}_{k+2,0}-\widehat{\tau}_{k+2,N})}{\widehat{\tau}_{k+2,0}\widehat{\tau}_{k+2,N}(1-\tau_{k+2,0})}\lesssim \frac{N\tau_{k+2,0}\widehat{\tau}_{k+2,0}(1-\widehat{\tau}_{k+2,N})\log T}{T\widehat{\tau}_{k+2,0}\widehat{\tau}_{k+2,N}(1-\tau_{k+2,0})} \nonumber\\
&\lesssim \frac{\log T}{K}.
\end{align*}
Inserting into \eqref{eq:proof-lemKL-temp-1}, we have
\begin{align}
\sum_{k=0}^{K-1}\frac{1-\widehat{\tau}_{k,N}}{\widehat{\tau}_{k,N}}\int_{{\tau}_{k,N}}^{{\tau}_{k,0}} \frac{\mathbb{E}[\mathsf{Tr}(\Sigma_\tau^2(x_\tau))]}{(1-\tau)^2}\mathrm d \tau
&\lesssim\sum_{k=0}^{K-3}\frac{\log T}{K}\mathbb{E}[\mathsf{Tr}(\Sigma_{\tau_{k+2,0}}(x_{\tau_{k+2,0}}))]+d\nonumber\\
&\lesssim d\log T.\label{eq:proof-lemKL-temp-2}
\end{align}
Inserting \eqref{eq:proof-lemKL-temp-2} into \eqref{eq:proof-lemKL-temp-3}, we have
\begin{align}
&\quad \sum_{k=0}^{K-1}\mathbb{E}_{x_{\tau_{k,0}}\sim p_{\widetilde{X}_{k}}}\left[\mathsf{KL}\left(p_{\widehat{X}_{k+1}|\widehat{X}_{k}}\left(\,\cdot\mymid x_{\tau_{k,0}}\right)\,\Vert\,p_{Y_{k+1}|Y_{k}}\left(\,\cdot\mymid x_{\tau_{k,0}}\right)\right)\right]\nonumber\\
&\lesssim\frac{Kd\log^4 T}{T^3}\min\Big\{d\log T+d\log T,L^2\log T\Big\}
+\frac{\log T}{T}\sum_{k=0}^{K-1}\sum_{i = 0}^{N-1} \varepsilon_{k,i}^2\nonumber\\
&\lesssim \frac{Kd\log^5 T}{T^3}\min\Big\{d,L^2\Big\}
+\varepsilon_{\mathsf{score}}^2\log T,
\end{align}
and we complete the proof of Lemma \ref{lem:KL}.

The remaining of this section shall use the following lemma to prove Lemma \ref{lem:KL-pre}.
Its proof is postponed to the end of this section.
\begin{lemma}\label{lem:lipschitz}
For any $x\in\mathcal{S}_\tau\cap\mathcal{L}_\tau$, $y\in\mathcal{S}_\tau$, and any $0<\tau<1$, we have
\begin{align*}
\left\|s_\tau^\star(x)-s_\tau^{\star}(y)\right\|_2\le 
\frac{C\min\{d\log T,L\}}{\tau}\left\|x-y\right\|_2,
\end{align*}
where $C$ is a sufficiently large constant dependent on $\theta+c_0$.
\end{lemma}

Recalling the definition of $\xi_{k,n}$, we have
\begin{align}\label{eq:proof-temp-2}
\frac{\left\|y_{\tau_{k, n}} - x_{\tau_{k, n}}\right\|_2}{\sqrt{1-\tau_{k,n}}} \le \frac{\left\|y^{\star}_{\tau_{k, n}} - z^{\star}_{\tau_{k, n}}\right\|_2}{\sqrt{1-\tau_{k,n}}} + \left\|\xi_{k,n}\right\|_2.
\end{align}
According to the definitions of $y^{\star}_{\tau_{k, n}}$ and $z^{\star}_{\tau_{k, n}}$, for $x_{\tau_{k,i}},y_{\tau_{k,i}}\in\mathcal{S}_{\tau_{k,i}}$, we have
\begin{align*}
\frac{\left\|y^{\star}_{\tau_{k, n}} - z^{\star}_{\tau_{k, n}}\right\|_2}{\sqrt{1-\tau_{k,n}}}  
&\le \sum_{i=1}^{n-1}\frac{\left\|s_{\tau_{k,i}}^{\star}(y_{\tau_{k,i}})-s_{\tau_{k,i}}^{\star}(x_{\tau_{k,i}})\right\|_2}{2(1-\tau_{k,i})^{3/2}}(\widehat{\tau}_{k,i-1}-\widehat{\tau}_{k,i})\nonumber\\
&\quad + \frac{\left\|s_{\tau_{k,n-1}}^{\star}(y_{\tau_{k,n-1}})-s_{\tau_{k,n-1}}^{\star}(x_{\tau_{k,n-1}})\right\|_2}{2(1-\tau_{k,n-1})^{3/2}}(\widehat{\tau}_{k,n-1}-{\tau}_{k,n})\nonumber\\
&\le \sum_{i=1}^{n-1}\frac{C\min\{d\log T,L\}\left\|y_{\tau_{k,i}}-x_{\tau_{k,i}}\right\|_2}{2\tau_{k,i}(1-\tau_{k,i})^{3/2}}(\widehat{\tau}_{k,i-1}-\widehat{\tau}_{k,i}) \nonumber\\
&\quad + \frac{C\min\{d\log T,L\}\left\|y_{\tau_{k,n-1}}-x_{\tau_{k,n-1}}\right\|_2}{2\tau_{k,n-1}(1-\tau_{k,n-1})^{3/2}}(\widehat{\tau}_{k,n-1}-{\tau}_{k,n})\nonumber\\
&\overset{\text{(a)}}{\lesssim} \frac{\min\{d\log T,L\}\log T}{T}\sum_{i=1}^{n-1}\frac{\left\|y_{\tau_{k,i}}-x_{\tau_{k,i}}\right\|_2}{\sqrt{1-\tau_{k,i}}}\notag\\
&\quad+ \frac{\min\{d\log T,L\}\log T\left\|y_{\tau_{k,n-1}}-x_{\tau_{k,n-1}}\right\|_2}{T\sqrt{1-\tau_{k,n-1}}}\nonumber\\
&\lesssim \frac{\min\{d\log T,L\}\log T}{T}\sum_{i=1}^{n-1}\frac{\left\|y_{\tau_{k,i}}-x_{\tau_{k,i}}\right\|_2}{\sqrt{1-\tau_{k,i}}},
\end{align*}
where (a) uses \eqref{eq:diff-tau-1} and \eqref{eq:diff-tau-2}.
Inserting into \eqref{eq:proof-temp-2}, we have
\begin{align*}
\frac{\left\|y_{\tau_{k, n}} - x_{\tau_{k, n}}\right\|_2}{\sqrt{1-\tau_{k,n}}} \lesssim  \frac{\min\{d\log T,L\}\log T}{T}\sum_{i=1}^{n-1}\frac{\left\|y_{\tau_{k,i}}-x_{\tau_{k,i}}\right\|_2}{\sqrt{1-\tau_{k,i}}} + \left\|\xi_{k,n}(x_{\tau_{k,0}})\right\|_2.
\end{align*}
By applying the above relation recursively, for $T\gtrsim \min\{d\log T,L\}N\log T$, we have
\begin{align*}
\frac{\left\|y_{\tau_{k, n}} - x_{\tau_{k, n}}\right\|_2}{\sqrt{1-\tau_{k,n}}} \lesssim \frac{\min\{d\log T,L\}\log T}{T}\sum_{i=1}^{n-1} \left\|\xi_{k,i}(x_{\tau_{k,0}})\right\|_2 + \left\|\xi_{k,n}(x_{\tau_{k,0}})\right\|_2.
\end{align*}
Thus 
\begin{align*}
\frac{\left\|y_{\tau_{k, n}} - x_{\tau_{k, n}}\right\|_2^2}{1-\tau_{k,n}} \lesssim \frac{N\min\{d\log T,L\}^2\log^2 T}{T^2}\sum_{i=1}^{n-1} \left\|\xi_{k,i}(x_{\tau_{k,0}})\right\|_2^2 + \left\|\xi_{k,n}(x_{\tau_{k,0}})\right\|_2^2.
\end{align*}
Define 
\begin{align*}
    \frac{\zeta_{k,n}}{1-\tau_{k,n}} = \frac{N\min\{d\log T,L\}^2\log^2 T}{T^2}\sum_{i=1}^{n-1}\frac{\|x_{\tau_{k, i}}-z^{\star}_{\tau_{k, i}}\|_2^2}{1-\tau_{k,i}} + \frac{\|x_{\tau_{k, n}}-z^{\star}_{\tau_{k, n}}\|_2^2}{1-\tau_{k,n}}.
\end{align*}
We have
\begin{align*}
\frac{\left\|y_{\tau_{k, n}} - x_{\tau_{k, n}}\right\|_2^2}{1-\tau_{k,n}} 
&\lesssim \frac{\zeta_{k,n}}{1-\tau_{k,n}} + \frac{N\min\{d\log T,L\}^2\log^2 T}{T^2}\sum_{i=1}^{n-1}\frac{\|y^{\star}_{\tau_{k, i}} - y_{\tau_{k, i}}\|_2^2}{1-\tau_{k,i}}\notag\\
&\quad + \frac{\|y^{\star}_{\tau_{k, n}} - y_{\tau_{k, n}}\|_2^2}{1-\tau_{k,n}}.
%&\lesssim \zeta_{k,n} + \frac{N\min\{d\log T,L\}^2\log^4 T}{T^3}\sum_{i=1}^n\frac{\tau_{k,i}\widetilde{\varepsilon}_{k,i}^2}{1-\tau_{k,i}} + \frac{\|y^{\star}_{\tau_{k, n}} - y_{\tau_{k, n}}\|^2}{1-\tau_{k,n}},
\end{align*}
According to Lemma \ref{lem:discrete-pre}, we have
\begin{align*}
\frac{\|y^{\star}_{\tau_{k, n}} - y_{\tau_{k, n}}\|_2^2}{1-\tau_{k,n}}\lesssim \frac{N\log^2 T}{T^2}\sum_{i=0}^{n-1}\frac{\tau_{k,i}\widetilde{\varepsilon}_{k,i}^2}{1-\tau_{k,i}}.
\end{align*}
Considering that $N^2\min\{d\log T,L\}^2\log^2 T/T^2\lesssim 1$, we have
\begin{align*}
\frac{N\min\{d\log T,L\}^2\log^2 T}{T^2}\sum_{i=1}^n\frac{\|y^{\star}_{\tau_{k, i}} - y_{\tau_{k, i}}\|_2^2}{1-\tau_{k,i}} + \frac{\|y^{\star}_{\tau_{k, n}} - y_{\tau_{k, n}}\|_2^2}{1-\tau_{k,n}} \lesssim \frac{N\log^2 T}{T^2}\sum_{i=0}^{n-1}\frac{\tau_{k,i}\widetilde{\varepsilon}_{k,i}^2}{1-\tau_{k,i}}.
\end{align*}
Thus we have
\begin{align*} %\label{eq:KL-pre}
\frac{\|y_{\tau_{k,n}}(x_k)-x_{\tau_{k,n}}(x_k)\|_2^2}{1-\tau_{k,n}}
&\le \frac{\zeta_{k,n}(x_k)}{1-\tau_{k,n}} + \frac{N\log^2 T}{T^2}\sum_{i=1}^n\frac{\tau_{k,i}\widetilde{\varepsilon}_{k,i}^2(x_k)}{1-\tau_{k,i}},
\end{align*}
which establishes \eqref{eq:lem-KL-pre-1}.
Furthermore, we have
\begin{align*}
&\quad \int_{\mathcal{E}_{k,n}} \frac{\zeta_{k,n}(x_k)}{1-\tau_{k,n}} p_{\widehat{X}_{k}}(x_k) \mathrm d x_k\notag\\
&\lesssim \int_{\mathcal{E}_{k,n}} \frac{N\min\{d\log T,L\}^2\log^2 T}{T^2}\sum_{i=1}^{n-1}\frac{\|x_{\tau_{k, i}}(x_k)-z^{\star}_{\tau_{k, i}}(x_k)\|_2^2}{1-\tau_{k,i}}
p_{\widehat{X}_{k}}(x_k) \mathrm d x_k\nonumber\\
&\quad + \int_{\mathcal{E}_{k,n}} \frac{\|x_{\tau_{k, n}}(x_k)-z^{\star}_{\tau_{k, n}}(x_k)\|_2^2}{1-\tau_{k,n}} p_{\widehat{X}_{k}}(x_k)\mathrm{d}x_k.
\end{align*}
Recalling \eqref{eq:lem-discrete-pre-2} in Lemma \ref{lem:discrete-pre} and the fact that $N^2\min\{d\log T,L\}^2\log^2 T/T^2\lesssim 1$, we have
\begin{align*}
&\quad\int_{\mathcal{E}_{k,n}} \frac{\zeta_{k,n}(x_k)}{1-\tau_{k,n}} p_{\widehat{X}_{k}}(x_k) \mathrm d x_k \notag\\
& \lesssim \frac{d\log^5 T}{T^3}\min\Big\{\frac{Nd\widehat{\tau}_{k,-1}\log T}{T(1-\widehat{\tau}_{k,-1})}+\int_{{\tau}_{k,n}}^{{\tau}_{k,0}} \frac{\mathbb{E}[\mathsf{Tr}(\Sigma_\tau^2(x_\tau))]}{(1-\tau)^2}\mathrm d \tau,\frac{NL^2\widehat{\tau}_{k,-1}\log T}{T(1-\widehat{\tau}_{k,-1})}\Big\},
% &\lesssim \int_{\mathcal{E}_{k,n}} \frac{\|x_{\tau_{k, n}}(x_k)-z^{\star}_{\tau_{k, n}}(x_k)\|_2^2}{1-\tau_{k,n}} p_{\widehat{X}_{k}}(x_k)\mathrm{d}x_k\nonumber\\
% &\lesssim \sum_{i = 0}^{n} \frac{\widehat{\tau}_{k,i-1}\min\{d\log T,L^2\}d\log^6 T}{T^4(1-\widehat{\tau}_{k,i-1})}\nonumber\\
% &\lesssim \frac{\widehat{\tau}_{k,-1}N\min\{d\log T,L^2\}d\log^6 T}{T^4(1-\widehat{\tau}_{k,-1})}
\end{align*}
and thus
\begin{align*}
&\quad \int_{\mathcal{E}_{k,n}} \zeta_{k,n}(x_k) p_{\widehat{X}_{k}}(x_k) \mathrm d x_k \notag\\
& \lesssim \frac{d\log^5 T}{T^3}\min\Big\{\frac{Nd\widehat{\tau}_{k,-1}\log T}{T}+(1-\widehat{\tau}_{k,n})\int_{{\tau}_{k,n}}^{{\tau}_{k,0}} \frac{\mathbb{E}[\mathsf{Tr}(\Sigma_\tau^2(x_\tau))]}{(1-\tau)^2}\mathrm d \tau,\frac{NL^2\widehat{\tau}_{k,-1}\log T}{T}\Big\}
\end{align*}
and establish \eqref{eq:lem-KL-pre-2},
where we use \eqref{eq:diff-tau-2}.

\noindent \textbf{Proof of Lemma \ref{lem:lipschitz}.}
%Now we are ready to prove Lemma \ref{lem:lipschitz}.
Before diving into the proof details, we first present the following lemma.
Its proof is postponed to Appendix \ref{subsec:proof-lem-ratio-pxy}.
\begin{lemma}\label{lem:ratio-pxy}
For any $x,y$ satisfying 
$$
\left\|x-y\right\|_2\le c\sqrt{\frac{\tau}{d\log T}}
$$
with $c\le \sqrt{\frac{1}{\theta+c_0}}/12$,
if $x\in\mathcal{S}_{\tau}$, we have
\begin{align*}
p_{X_\tau}(x) - p_{X_\tau}(y) \le  6\|y-x\|_2\sqrt{\frac{(\theta+c_0)d\log T}{\tau}}p_{X_{\tau}}(x).
\end{align*}
Moreover, if $x,y\in\mathcal{S}_\tau$, then we have
\begin{align*}
\frac{1}{2}\le 1-6\left\|x-y\right\|_2\sqrt{\frac{(\theta+c_0)d\log T}{\tau}} \le \frac{p_{X_\tau}(x)}{p_{X_\tau}(y)}\le\frac{1}{1-6\left\|x-y\right\|_2\sqrt{\frac{(\theta+c_0)d\log T}{\tau}}} \le 2.
\end{align*}
\end{lemma}

We first prove that for $x\in\mathcal{S}_\tau$ and $y\in\mathcal{S}_\tau$, we have
\begin{align}\label{eq:proof-lipschitz-d}
\|s_\tau^{\star}(x) - s_\tau^{\star}(y)\|_2\lesssim \frac{d\log T}{\tau}\|x-y\|.
\end{align}
To this end,
without loss of generality, we assume that $p_{X_{\tau}}(y)\ge p_{X_{\tau}}(x)$.
Then according to the definition of score function, we have
\begin{align*}
s_\tau^{\star}(y) - s_\tau^{\star}(x) 
&= \frac{1}{\tau}(x-y) - \frac{1}{\tau}\int_{x_0}(x-\sqrt{1-\tau}x_0)p_{X_0|X_\tau}(x_0|y)\notag\\
&\quad + (x-\sqrt{1-\tau}x_0)p_{X_0|X_\tau}(x_0|x)\mathrm{d}x_0.
\end{align*}

Thus we have
\begin{align}\label{eq:proof-lipschitz-0}
&\quad\left\|s_\tau^{\star}(y) - s_\tau^{\star}(x)\right\|_2 \notag\\
&\le \frac{1}{\tau}\left\|x-y\right\|_2 + \frac{1}{\tau}\left\|\int_{x_0}(x-\sqrt{1-\tau}x_0)\frac{p_{X_\tau|X_0}(y|x_0)-p_{X_\tau|X_0}(x|x_0)}{p_{X_\tau}(y)}p_{X_0}(x_0)\mathrm{d}x_0\right\|_2\nonumber\\
&\quad+ \frac{1}{\tau}\left\|\left(\frac{p_{X_\tau}(x)}{p_{X_\tau}(y)}-1\right)\int_{x_0}(x-\sqrt{1-\tau}x_0)p_{X_0|X_\tau}(x_0|x)\mathrm{d}x_0\right\|_2.
\end{align}
For the second and the third term, we define $\mathcal{R} = \{x_0\in\mathbb{R}^d: \left\|x-\sqrt{1-\tau}x_0\right\|_2\le\left\|y-\sqrt{1-\tau}x_0\right\|_2\}$, and have
\begin{small}
\begin{align}
&\quad\left\|\int_{x_0}(x-\sqrt{1-\tau}x_0)\frac{p_{X_\tau|X_0}(y|x_0)-p_{X_\tau|X_0}(x|x_0)}{p_{X_\tau}(y)}p_{X_0}(x_0)\mathrm{d}x_0\right\|_2\nonumber\\
&\le \left\|\int_{\mathcal{R}}(x-\sqrt{1-\tau}x_0)\frac{p_{X_\tau|X_0}(x|x_0)}{p_{X_\tau}(y)}p_{X_0}(x_0)\left(\exp\left(\frac{(x-y)^{\top}(x+y-2\sqrt{1-\tau}x_0}{2\tau}\right)-1\right)\mathrm{d}x_0\right\|_2\nonumber\\
&\quad +\left\|\int_{\mathcal{R}^{\rm c}}(x-\sqrt{1-\tau}x_0)\frac{p_{X_\tau|X_0}(y|x_0)}{p_{X_\tau}(y)}p_{X_0}(x_0)\left(1-\exp\left(\frac{(y-x)^{\top}(x+y-2\sqrt{1-\tau}x_0}{2\tau}\right)\right)\mathrm{d}x_0\right\|_2.\nonumber
\end{align}
\end{small}
The first term is smaller than
$$
\int_{\mathcal{R}}\|x-\sqrt{1-\tau}x_0\|_2\frac{\|x-y\|\|x+y-2\sqrt{1-\tau}x_0\|_2}{2\tau}p_{X_0|X_\tau}(x_0|x)\mathrm{d}x_0\frac{p_{X_\tau}(x)}{p_{X_\tau}(y)},
$$
and the second term is smaller than
\begin{align*}
&\quad\int_{\mathcal{R}^{\rm c}}\|y-\sqrt{1-\tau}x_0\|_2\frac{\|y-x\|_2\|x+y-2\sqrt{1-\tau}x_0\|_2}{2\tau}p_{X_0|X_\tau}(x_0|y)\mathrm{d}x_0\nonumber\\
&\quad+ \left\|(x-y)\left(1-\frac{p_{X_\tau}(x)}{p_{X_\tau}(y)}\right)\right\|_2.
\end{align*}
Combining these, we have
\begin{align}
&\quad \left\|\int_{x_0}(x-\sqrt{1-\tau}x_0)\frac{p_{X_\tau|X_0}(y|x_0)-p_{X_\tau|X_0}(x|x_0)}{p_{X_\tau}(y)}p_{X_0}(x_0)\mathrm{d}x_0\right\|_2\notag\\
&\le \frac{\left\|y-x\right\|_2}{\tau}\int_{\mathcal{R}}\|x-\sqrt{1-\tau}x_0\|_2^2p_{X_0|X_\tau}(x_0|x)\mathrm{d}x_0 \notag\\
&\quad +\frac{\left\|y-x\right\|_2^2}{2\tau}\int_{\mathcal{R}}\|x-\sqrt{1-\tau}x_0\|_2p_{X_0|X_\tau}(x_0|x)\mathrm{d}x_0\nonumber\\
&\quad +\frac{\left\|y-x\right\|_2}{\tau}\int_{\mathcal{R}^{\rm c}}\left\|y-\sqrt{1-\tau}x_0\right\|_2^2p_{X_0|X_\tau}(x_0|y)\mathrm{d}x_0 \nonumber\\
&\quad+\frac{\left\|y-x\right\|_2^2}{2\tau}\int_{\mathcal{R}^{\rm c}}\|y-\sqrt{1-\tau}x_0\|_2p_{X_0|X_\tau}(x_0|y)\mathrm{d}x_0 + \left\|x-y\right\|_2\nonumber\\
&\overset{\text{(a)}}{\lesssim} d\log T\left\|x-y\right\|_2 + \left\|x-y\right\|_2^2\sqrt{\frac{d\log T}{\tau}}\lesssim d\log T\left\|x-y\right\|_2,\label{eq:proof-lipschitz-1}
\end{align}
for 
$$
\left\|x-y\right\|_2\le c\sqrt{\frac{\tau}{d\log T}},
$$
where (a) uses Lemma \ref{lem:bound-s-J}.
For the third term, by using Lemma \ref{lem:ratio-pxy}, we have
\begin{align}\label{eq:proof-lipschitz-2}
&\quad \left\|\left(\frac{p_{X_\tau}(x)}{p_{X_\tau}(y)}-1\right)\int_{x_0}(x-\sqrt{1-\tau}x_0)p_{X_0|X_\tau}(x_0|x)\mathrm{d}x_0\right\|_2\nonumber\\
&\le6\left\|x-y\right\|_2\sqrt{\frac{(\theta+c_0)d\log T}{\tau}}\int_{x_0}\left\|x-\sqrt{1-\tau}x_0\right\|_2p_{X_0|X_\tau}(x_0|x)\mathrm{d}x_0\nonumber\\
&\lesssim d\log T\left\|x-y\right\|_2.
\end{align}
Inserting \eqref{eq:proof-lipschitz-1} and \eqref{eq:proof-lipschitz-2} into \eqref{eq:proof-lipschitz-0}, we have for $\left\|x-y\right\|_2\le c\sqrt{\frac{\tau}{d\log T}}$,
\begin{align*}
\left\|s_\tau^{\star}(y) - s_\tau^{\star}(x)\right\|_2 \lesssim \frac{d\log T}{\tau}\left\|x-y\right\|_2.
\end{align*}
For $\left\|x-y\right\|_2\ge c\sqrt{\frac{\tau}{d\log T}}$, we have
\begin{align*}
\left\|s_\tau^{\star}(y) - s_\tau^{\star}(x)\right\|_2 \le \left\|s_\tau^{\star}(y)\right\|_2+ \left\|s_\tau^{\star}(x)\right\|_2\le 10\sqrt{\frac{(\theta+c_0)d\log T}{\tau}}
\lesssim \frac{d\log T}{\tau}\left\|x-y\right\|_2.
\end{align*}
Thus we prove \eqref{eq:proof-lipschitz-d}.

In addition, for $x\in\mathcal{L}_\tau$ and $\|x-y\|\le \frac{C\sqrt{d\tau}\log T}{L}$, according to Definition \ref{def:score-lipschitz}, we have
\begin{align*}
\left\|s_\tau^{\star}(y) - s_\tau^{\star}(x)\right\|_2 \le \frac{L}{\tau}\|x-y\|_2.
\end{align*}
Otherwise, for $x\in\mathcal{S}_\tau$, $y\in\mathcal{S}_\tau$, and $\|x-y\|> \frac{C\sqrt{d\tau}\log T}{L}$, according to Lemma \ref{lem:bound-s-J},
we have
\begin{align*}
\left\|s_\tau^{\star}(y) - s_\tau^{\star}(x)\right\|_2 \le 2\sqrt{\frac{25(\theta+c_0)d\log T}{\tau}} \le \frac{L}{\tau}\|x-y\|_2,
\end{align*}
for $C\ge 10\sqrt{\theta+c_0}$.
Thus we complete the proof.

\subsection{Proof of Lemma \ref{lem:endpoint}}
\label{sec:proof-lem-endpoint}

According to Lemma \ref{lem:ODE}, we know that $\widehat{X}_0 \overset{\mathrm{d}}{=} X_{\tau_{0,0}} = \sqrt{1-\tau_{0,0}}X_0 + \sqrt{\tau_{0,0}}Z$ for some standard Gaussian variable $Z$. 
Then we have
\begin{align*}
\mathsf{KL}\big(p_{\widehat{X}_{0}}\Vert p_{Y_{0}}\big) \le& \mathsf{KL}\big(p_{X_{\tau_{0,0}},X_0}\Vert p_{Y_{0}}p_{X_0}\big) = \mathbb{E}_{X_0\sim p_{\mathsf{data}}}[\mathsf{KL}\big(p_{\widehat{X}_{0}|X_0}\Vert p_{Y_{0}}\big)]\nonumber\\
=&\frac12\left[(1-\tau_{0,0})\mathbb{E}_{X_0\sim p_{\mathsf{data}}}\left\|X_0\right\|^2-d\log(\tau_{0,0})+d\tau_{0,0}-d\right]\\
\le& \frac12T^{c_R-c_0}+d\frac{(1-\tau_{0,0})^2}{2\tau_{0,0}}
\le \frac12T^{c_R-c_0}+\frac{d}{T^{2c_0}} \le \frac{1}{T^{10}},
\end{align*}
as long as $c_0\ge \max\{c_R+10,10\}$, and $T\gtrsim d^{1/10}$.
% In addition, the analysis of $\mathsf{TV}(p_{\mathsf{data}},p_{\widehat{X}_K})$ can be found in the proof of Theorem 2 in \citep{Chen2023The}.

The remaining of this section focuses on the proof of \eqref{eq:prob-setE}.
% Before diving into the proof details, we first present the following lemma.
% Now we are ready to prove \eqref{eq:prob-setE}.
For $d\log T \ge L$, we have $\mathcal{E}_k = \emptyset$ and \eqref{eq:prob-setE} holds trivially.
Below we only consider the case of $d\log T\le L$.
We decompose the set $\mathcal{E}_k^{\rm c}$ as 
\begin{align*}
\mathcal{E}_k^{\rm c} = \cup_{n=0}^{N-1} \widehat{\mathcal{E}}_{k,n},
\end{align*}
where for $1\le n\le N-1$,
\begin{small}
\begin{align*}
\widehat{\mathcal{E}}_{k,n} = \{x_k: &x_{\tau_{k,i}}(x_k)\in\widetilde{\mathcal{S}}_{\tau_{k,i}}\cap\mathcal{L}_{\tau_{k,i}}, \forall 0\le i\le N-1, \notag\\
&y_{\tau_{k,i}}(x_k)\in\mathcal{S}_{\tau_{k,i}}, \forall 0\le i \le n-1, y_{\tau_{k,n}}(x_k)\notin\mathcal{S}_{\tau_{k,n}}\},
\end{align*}
\end{small}
and 
$$
\widehat{\mathcal{E}}_{k,0} = \{x_k: \exists~0\le i\le N-1, x_{\tau_{k,i}}(x_k)\notin\widetilde{\mathcal{S}}_{\tau_{k,i}}\cap\mathcal{L}_{\tau_{k,i}}\}.
$$
Furthermore, we introduce another auxiliary set
$$
\mathcal{B}_{k,n} = \left\{x_k:\frac{N\widehat{\tau}_{k,-1}\log^2 T}{T^2}\sum_{i=0}^{n-1}\|s_{\tau_{k,i}}(y_{\tau_{k,i}}(x_k))-s_{\tau_{k,i}}^{\star}(y_{\tau_{k,i}}(x_k))\|^2 \le \frac{C\tau_{k,n}}{d\log T}\right\},
$$
where $C$ is a sufficiently small constant.
Then we have 
$$
\mathcal{E}_k^{\rm c} \subset \widehat{\mathcal{E}}_{k,0}\cup \left(\cup_{n=1}^{N-1} \left(\widehat{\mathcal{E}}_{k,n}\cap\mathcal{B}_{k,n}\right)\right) \cup \left(\cup_{n=1}^{N-1}\mathcal{B}_{k,n}^{\rm c}\right).
$$
Thus we have
\begin{align}\label{eq:proof-lem-prob-setE-temp-1}
P(\widehat{X}_k\in\mathcal{E}_k^{\rm c}) \le P(\widehat{X}_k\in \widehat{\mathcal{E}}_{k,0}) + \sum_{n=1}^{N-1}P(\widehat{X}_k\in \widehat{\mathcal{E}}_{k,n}\cap\mathcal{B}_{k,n}) + P(\widehat{X}_k\in\cup_{n=1}^{N-1}\mathcal{B}_{k,n}^{\rm c}).
\end{align}
Below we shall calculate these terms separately.
We start from considering $P(\widehat{X}_k\in \widehat{\mathcal{E}}_{k,0})$.
Noticing that $x_{\tau_{k,n}}$ has the identical distribution with $X_{\tau_{k,n}}$, we have
\begin{align}\label{eq:proof-lem-prob-setE-temp-2}
P(\widehat{X}_k\in \widehat{\mathcal{E}}_{k,0}) \le \sum_{n=0}^{N-1}P({X}_{\tau_{k,n}}\in \widetilde{\mathcal{S}}_{\tau_{k,n}}^{\rm c}) + \sum_{n=0}^N P({X}_{\tau_{k,n}}\in \mathcal{L}_{\tau_{k,n}}^{\rm c})\lesssim \frac{N}{T^4},
%\le \sum_{n=0}^{N-1}\int_{\widetilde{\mathcal{S}}_{\tau_{k,n}}^{\rm c}\cap\mathcal{A}} p_{X_{\tau_{k,n}}}(x) \mathrm d x + \sum_{n=0}^{N-1}\int_{\mathcal{A}^{\rm c}} p_{X_{\tau_{k,n}}}(x) \mathrm d x.
\end{align}
where the last inequality comes from the fact that
$$
P({X}_{\tau_{k,n}}\in \widetilde{\mathcal{S}}_{\tau_{k,n}}^{\rm c})\lesssim \frac{1}{T^4},\qquad P({X}_{\tau_{k,n}}\in \mathcal{L}_{\tau_{k,n}}^{\rm c})\lesssim \frac{1}{T^4},
$$
where the first inequality is proved in \citet{li2024d} (cf. (A.18)), and the second inequality comes from Definition \ref{def:score-lipschitz}.

Next, we analyze $P(\widehat{X}_k\in \widehat{\mathcal{E}}_{k,n}\cap\mathcal{B}_{k,n})$.
According to Lemma \ref{lem:ratio-pxy}, we have
\begin{align*}
p_{X_\tau}(y) \ge \left(1- 6\|y-x\|\sqrt{\frac{(\theta+c_0)d\log T}{\tau}}\right)p_{X_{\tau}}(x),\quad {\rm if}~ \|x-y\|^2\le \frac{c\tau}{d\log T},
\end{align*}
where $c = 1/(144(\theta+c_0))$.
Recalling that $\log p_{X_{\tau}}(x) \ge \log 2 - \theta d\log T$, we have
\begin{align*}
&\quad P(\widehat{X}_k\in \widehat{\mathcal{E}}_{k,n}\cap\mathcal{B}_{k,n}) \nonumber\\
&\le P\Big(\frac{N\widehat{\tau}_{k,-1}\log^2 T}{T^2}\sum_{i=0}^{n-1}\widetilde{\varepsilon}_{k,i}^2(\widehat{X}_k) \le \frac{C\tau_{k,n}}{d\log T},\|x_{\tau_{k,n}}(\widehat{X}_k)-y_{\tau_{k,n}}(\widehat{X}_k)\|_2^2\ge \frac{c\tau_{k,n}}{d\log T},\nonumber\\
&\quad x_{\tau_{k,i}}(\widehat{X}_k)\in\widetilde{\mathcal{S}}_{\tau_{k,i}}\cap\mathcal{L}_{\tau_{k,i}}, y_{\tau_{k,i}}(\widehat{X}_k)\in\mathcal{S}_{\tau_{k,i}},\forall~0\le i< n\Big).
\end{align*}
According to \eqref{eq:lem-KL-pre-1}, for 
\begin{align*}
x\in\mathcal{A}_{k,n}
&:=\Big\{x:\frac{N\widehat{\tau}_{k,-1}\log^2 T}{T^2}\sum_{i=0}^{n-1}\widetilde{\varepsilon}_{k,i}^2(x) \le \frac{C\tau_{k,n}}{d\log T},\notag\\
&\qquad\qquad x_{\tau_{k,i}}(x)\in\widetilde{\mathcal{S}}_{\tau_{k,i}}\cap\mathcal{L}_{\tau_{k,i}}, y_{\tau_{k,i}}(x)\in\mathcal{S}_{\tau_{k,i}},\forall~0\le i< n\Big\},
\end{align*}
which satisfies $\mathcal{A}_{k,n}\subset \mathcal{E}_{k,n}$,
we have
\begin{align*}
\|x_{\tau_{k,n}}(x)-y_{\tau_{k,n}}(x)\|^2 
&\lesssim  \zeta_{k,n}(x) + \frac{N\log^2 T}{T^2}\sum_{i=0}^{n-1}\widehat{\tau}_{k,i-1}\widetilde{\varepsilon}_{k,i}^2(x)\nonumber\\
&\lesssim\zeta_{k,n}(x) + \frac{\widehat{\tau}_{k,-1}N\log^2 T}{T^2}\sum_{i=0}^{n-1}\widetilde{\varepsilon}_{k,i}^2(x)
\overset{\text{(a)}}{\lesssim}\zeta_{k,n}(x) + \frac{C\tau_{k,n}}{d\log T},
\end{align*}
where (a) comes from the condition that $x\in\mathcal{A}_{k,n}$.
Assuming $C\le c/2$, we have
\begin{align*}%\label{eq:proof-lem-prob-setE-temp-3}
&\quad P(\widehat{X}_k\in \widehat{\mathcal{E}}_{k,n}\cap\mathcal{B}_{k,n}) \notag\\
&\le P\left(\widehat{X}_k\in \mathcal{A}_{k,n}, \zeta_{k,n}(\widehat{X}_k)\ge \frac{c\tau_{k,n}}{2d\log T}\right)\nonumber\\
&\overset{\text{(a)}}{\le} \frac{2d\log T}{c\tau_{k,n}}\int_{\mathcal{A}_{k,n}} \zeta_{k,n}(x) p_{\widehat{X}_k}(x) \mathrm d x \le \frac{2d\log T}{c\tau_{k,n}}\int_{{\mathcal{E}}_{k,n}} \zeta_{k,n}(x) p_{\widehat{X}_k}(x) \mathrm d x\nonumber\\
&\overset{\text{(b)}}{\lesssim} \frac{Kd\log^4 T}{T^3}\min\Big\{\frac{Nd\log T}{T}+\frac{1-\widehat{\tau}_{k,n}}{\widehat{\tau}_{k,n}}\int_{{\tau}_{k,n}}^{{\tau}_{k,0}} \frac{\mathbb{E}[\mathsf{Tr}(\Sigma_\tau^2(x_\tau))]}{(1-\tau)^2}\mathrm d \tau,\frac{NL^2\log T}{T}\Big\}\nonumber\\
&\lesssim \frac{d\log^5 T}{T^3}\min\Big\{d+\frac{K(1-\widehat{\tau}_{k,N})}{\widehat{\tau}_{k,N}\log T}\int_{{\tau}_{k,N}}^{{\tau}_{k,0}} \frac{\mathbb{E}[\mathsf{Tr}(\Sigma_\tau^2(x_\tau))]}{(1-\tau)^2}\mathrm d \tau,L^2\Big\}
\end{align*}
where (a) uses the Markov inequality, and (b) uses \eqref{eq:lem-KL-pre-2}  and the fact that $K\gtrsim \min\{d\log T,L\}\log T = d\log^2 T$ for $d\log T\lesssim L$.
Summing from $n=0$ to $N-1$, we have
\begin{align}\label{eq:proof-lem-prob-setE-temp-3}
\sum_{n=1}^{N-1}P(\widehat{X}_k\in \widehat{\mathcal{E}}_{k,n}\cap\mathcal{B}_{k,n}) 
&\lesssim \frac{\log^3 T}{T^2}\min\Big\{d+\frac{K(1-\widehat{\tau}_{k,N})}{\widehat{\tau}_{k,N}\log T}\int_{{\tau}_{k,N}}^{{\tau}_{k,0}} \frac{\mathbb{E}[\mathsf{Tr}(\Sigma_\tau^2(x_\tau))]}{(1-\tau)^2}\mathrm d \tau,L^2\Big\}.
\end{align}

Finally, we analyze $P(\widehat{X}_k\in\cup_{n=1}^{N-1}\mathcal{B}_{k,n}^{\rm c})$. Noticing that $\mathcal{B}_{k,n}\subset \mathcal{B}_{k,n-1}$,
we have
\begin{align}\label{eq:proof-lem-prob-setE-temp-4}
&\quad P(\widehat{X}_k\in\cup_{n=0}^{N-1}\mathcal{B}_{k,n}^{\rm c}) 
\le P(\widehat{X}_k\in\mathcal{B}_{k,N-1}^{\rm c})\nonumber\\
&\le P\left(\frac{N\widehat{\tau}_{k,0}\log^2 T}{T^2}\sum_{i=1}^{N-1}\|s_{\tau_{k,i}}(y_{\tau_{k,i}}(\widehat{X}_k))-s_{\tau_{k,i}}^{\star}(y_{\tau_{k,i}}(\widehat{X}_k))\|^2 > \frac{C\tau_{k,N}}{d\log T}\right)\nonumber\\
&\overset{\text{(a)}}{\lesssim} \frac{N d\log^3 T}{T^2}\sum_{i=1}^{N-1}\mathbb{E}_{x_k\sim\widehat{X}_k}\|s_{\tau_{k,i}}(y_{\tau_{k,i}}(x_k))-s_{\tau_{k,i}}^{\star}(y_{\tau_{k,i}}(x_k))\|^2 \nonumber\\
&\lesssim \frac{N d\log^3 T}{T^2}\sum_{i=0}^{N-1}\varepsilon_{k,i}^2,
\end{align}
where (a) uses Markov inequality.
Inserting \eqref{eq:proof-lem-prob-setE-temp-2}, \eqref{eq:proof-lem-prob-setE-temp-3}, and \eqref{eq:proof-lem-prob-setE-temp-4} into \eqref{eq:proof-lem-prob-setE-temp-1}, we have
\begin{align}
P(\widehat{X}_k\in\mathcal{E}_k) 
&\lesssim \frac{N}{T^4} + \frac{\log T}{T}\sum_{i=0}^{N-1}\varepsilon_{k,i}^2 \notag\\
&\qquad +\frac{\log^3 T}{T^2}\min\Big\{d+\frac{K(1-\widehat{\tau}_{k,N})}{\widehat{\tau}_{k,N}\log T}\int_{{\tau}_{k,N}}^{{\tau}_{k,0}} \frac{\mathbb{E}[\mathsf{Tr}(\Sigma_\tau^2(x_\tau))]}{(1-\tau)^2}\mathrm d \tau,L^2\Big\}.\label{eq:lem-setE}
\end{align}
Moreover, we have
\begin{align*}
\sum_{k=0}^{K-1}P(\widehat{X}_k\in\mathcal{E}_k) &\overset{\text{(a)}}{\lesssim} \frac{1}{T^3} + \frac{\log T}{T}\sum_{k=0}^{K-1}\sum_{i=0}^{N-1}\varepsilon_{k,i}^2 +\frac{K\log^3 T}{T^2}\min\Big\{d+d,L^2\Big\}\nonumber\\
&\overset{\text{(b)}}{\lesssim}  \frac{d^2\log^5 T}{T^2} + \frac{\log T}{T}\sum_{k=0}^{K-1}\sum_{i=0}^{N-1}\varepsilon_{k,i}^2 
\end{align*}
where (a) uses \eqref{eq:proof-lemKL-temp-2} and (b) uses the fact that $d\log T\lesssim L$.

\subsection{Proof of \eqref{eq:KL-infty}}
\label{subsec:proof-eq-KL-infty}

We first show a basic inequality: for any two probability density functions $f$, $g$, and any set $\mathcal{R}$, we have
\begin{align}\label{eq:KL-int}
\int_{\mathcal{R}} f(x)\log \frac{f(x)}{g(x)} \mathrm d x\ge  \log \left(\frac{\int_\mathcal{R} f(x) \mathrm d x}{\int_\mathcal{R} g(x) \mathrm d x }\right)\int_\mathcal{R} f(x) \mathrm d x
\end{align}
which is proved in Lemma 6 in \citet{li2024d}.
Now we are ready to prove \eqref{eq:KL-infty}.
To this end, let us prove a more general conclusion.
For any two probability density functions $f$ and $g$, and any set $\mathcal{R}$, assume that $\tilde{f}$ and $\tilde{g}$ is defined as
\begin{align*}
\tilde{f}(x) = f(x)\mathds{1}\{x\in\mathcal{R}\} + \int_{x\notin\mathcal{R}} f(x)\mathrm{d} x \delta_\infty,\quad \tilde{g}(x) = g(x)\mathds{1}\{x\in\mathcal{R}\} + \int_{x\notin\mathcal{R}} g(x)\mathrm{d} x \delta_\infty.
\end{align*}
Then
\begin{align*}
\mathsf{KL}(\tilde{f}\Vert \tilde{g}) \le \mathsf{KL}(f\Vert g).
\end{align*}
Towards this, according to definitions, we have
\begin{align*}
\mathsf{KL}(\tilde{f}\Vert \tilde{g}) - \mathsf{KL}(f\Vert g) = \int_{x\notin\mathcal{R}} f(x)\mathrm d x\log\left(\frac{\int_{x\notin\mathcal{R}} f(x)\mathrm d x}{\int_{x\notin\mathcal{R}} g(x)\mathrm d x}\right) - \int_{x\notin\mathcal{R}} f(x)\log\left(\frac{f(x)}{g(x)}\right)\mathrm d x \le 0
\end{align*}
and complete the proof.

\section{Parallel sampling}

\subsection{Parallel algorithm}
\label{subsec:parallel-alg}

The parallel sampling procedure follows the same structure as the original sampler described in \eqref{eq:sampler}: it consists of $K$ rounds, with each round comprising several iterations. The key difference is that, in each round, we use $N$ processors and perform $M\ll N$ iterations. In each iteration, each processor updates the sample $Y_{m,k,n}$ using the outputs of other processors from the previous iteration $\{Y_{m-1,k,i}\}_{i< n}$.
The implementation details for the $k$-th round are as follows.
%with the key difference being that updates to $Y_{k,n}$ for $n=1,\cdots,N$ are performed simultaneously. The process in each round $k$ is as follows:
\begin{enumerate}
\item Initialization: for the $n$-th parallel processor, the sample is initialized as
\begin{subequations}
\label{eq:sampler-parallel}
\begin{align}
\frac{Y_{0, k, n}}{\sqrt{1-\tau_{k, n}}} = \frac{Y_k}{\sqrt{1-\tau_{k,0}}}, \quad n=1,\cdots,N.
\end{align}
\item Parallel updates:
we use $N$ processors to update $Y_{m,k,n}$ ($n=1,\cdots,N$) simultaneously for $M$ iterations.
% For $M$ parallel rounds, all $Y_{m,k,n}$ ($n=1,\cdots,N$)
% are updated simultaneously.
In the $m$-th iteration, the update rule is:
\begin{align}
\frac{Y_{m, k, n}}{\sqrt{1-\tau_{k, n}}} &= \frac{Y_k}{\sqrt{1-\tau_{k,0}}} 
+ \frac{s_{T-\frac{kN}{2}+1}(Y_{k})}{2(1-\tau_{k,0})^{3/2}}(\tau_{k, 0} - \widehat{\tau}_{k,0})\notag\\
&\quad
+ \sum_{i = 1}^{n-1} \frac{s_{T-\frac{kN}{2}-i+1}(Y_{m-1, k,i})}{2(1-\tau_{k,i})^{3/2}}(\widehat{\tau}_{k,i-1} - \widehat{\tau}_{k,i})\notag\\
&\quad + \frac{s_{T-\frac{kN}{2}-n+2}(Y_{m-1,k,n-1})}{2(1-\tau_{k,n-1})^{3/2}}(\widehat{\tau}_{k,n-1} - \tau_{k, n}).
\end{align}
\item Noise injection:
Once $Y_{M, k, N}$ is obtained, we update $Y_{k+1}$ as follows.
\begin{align}
Y_{k+1} &= \sqrt{\frac{1-\tau_{k+1, 0}}{1-\tau_{k, N}}}Y_{M, k, N} + \sqrt{\frac{\tau_{k+1, 0}-\tau_{k, N}}{1-\tau_{k, N}}}Z_k,
\end{align}
\end{subequations}
where $Z_k\sim\mathcal{N}(0,I_d)$.
\end{enumerate}

In this parallel framework, the total number of parallel rounds required to generate the final sample $Y_K$ is $MK$, and $N$ parallel processors are needed.
The convergence rate of this procedure is established in Theorem~\ref{thm:parallel}.
We remark that the implementation of this parallel algorithm assumes that the GPU memory is capable of supporting score estimations for a large batch of data simultaneously. Parallelizing across multiple GPUs introduces additional communication overhead, which may impact efficiency.

\subsection{Analysis for parallelization (Theorem~\ref{thm:parallel})}
\label{sec:proof-thm-parallel}

%\begin{align*}
%\frac{y_{0, \tau_{k, n}}}{\sqrt{1-\tau_{k, n}}} = \frac{x_{\widehat{\tau}_{k, 0}}}{\sqrt{1-\widehat{\tau}_{k,0}}} 
%\end{align*}
By comparing the update rules for $Y_{k, n}$ and $Y_{m, k, n}$,
it is natural to control the difference of the following two sequences:
\begin{align*}
\frac{y_{\tau_{k, n}}(y_{\tau_{k,0}})}{\sqrt{1-\tau_{k, n}}} &= \frac{y_{\tau_{k, 0}}}{\sqrt{1-\tau_{k,0}}} 
+ \frac{s_{\tau_{k,0}}(y_{\tau_{k, 0}})}{2(1-\tau_{k,0})^{3/2}}(\tau_{k, 0} - \widehat{\tau}_{k,0})
+ \sum_{i = 1}^{n-1} \frac{s_{\tau_{k,i}}(y_{\tau_{k,i}})}{2(1-\tau_{k,i})^{3/2}}(\widehat{\tau}_{k,i-1} - \widehat{\tau}_{k,i}) \\
&\qquad\qquad+ \frac{s_{\tau_{k,n-1}}(y_{\tau_{k,n-1}})}{2(1-\tau_{k,n-1})^{3/2}}(\widehat{\tau}_{k,n-1} - \tau_{k, n}),
\end{align*}
and
\begin{align*}
\frac{y_{m, \tau_{k, n}}(y_{\tau_{k,0}})}{\sqrt{1-\tau_{k, n}}} &= \frac{y_{\tau_{k, 0}}}{\sqrt{1-\tau_{k,0}}} 
+ \frac{s_{\tau_{k,0}}(y_{\tau_{k, 0}})}{2(1-\tau_{k,0})^{3/2}}(\tau_{k, 0} - \widehat{\tau}_{k,0})
+ \sum_{i = 1}^{n-1} \frac{s_{\tau_{k,i}}(y_{m-1,\tau_{k,i}})}{2(1-\tau_{k,i})^{3/2}}(\widehat{\tau}_{k,i-1} - \widehat{\tau}_{k,i}) \\
&\qquad\qquad+ \frac{s_{\tau_{k,n-1}}(y_{m-1,\tau_{k,n-1}})}{2(1-\tau_{k,n-1})^{3/2}}(\widehat{\tau}_{k,n-1} - \tau_{k, n}).
\end{align*}
We construct the typical set
\begin{align*}
\mathcal{E}_k: = \left\{
\begin{array}{ll}
   \{x_{\tau_{k,0}}: x_{\tau_{k,n}}(x_{\tau_{k,0}})\in\widetilde{\mathcal{S}}_{\tau_{k,n}}\cap\mathcal{L}_{\tau_{k,n}}, y_{m,\tau_{k,n}}(x_{\tau_{k,0}})\in\mathcal{S}_{\tau_{k,n}}, &\nonumber\\
   \quad\qquad y_{\tau_{k,n}}(x_{\tau_{k,0}})\in\mathcal{S}_{\tau_{k,n}}\cap\mathcal{L}_{\tau_{k,n}}, \forall~0\le n < N,\forall~0\le m< M\},  & {\rm if}~L> d\log T, \\
   \emptyset  &  {\rm if}~L\le d\log T,
\end{array}
\right.
\end{align*}
and the auxiliary sequences $\widetilde{X}_k$, $\widetilde{Y}_k$ for $k=0\cdots,K$ as \eqref{eq:def-Xtilde} and \eqref{eq:def-Ytilde}, with
\begin{align*}
P_{Y_{k+1}|Y_k}(y|{y}_k) = \phi\left(y|y_{M,\tau_{k,N}}({y}_k),\sigma_k^2\right).
\end{align*}

Similar with the proof of Theorem \ref{thm:main}, we have
\begin{align*}
&\mathsf{TV}\big(q_K, p_{Y_{K}}\big)
\le 2\sum_{k=0}^{K-1} P(\widehat{X}_{k}\in\mathcal{E}_{k}^{\rm c})\nonumber\\
&\quad+ \sqrt{\frac12\mathsf{KL}\big(p_{\widehat{X}_{0}}\Vert p_{{Y}_{0}}\big)+\frac12\sum_{k=0}^{K-1}\mathbb{E}_{x_{k}\sim p_{\widetilde{X}_{k}}}\left[\mathsf{KL}\big(p_{\widehat{X}_{k+1}|\widehat{X}_{k}}\left(\,\cdot\mymid x_{k}\right)\,\Vert\,p_{{Y}_{k+1}|{Y}_{k}}\left(\,\cdot\mymid x_{k}\right)\big)\right]}.
\end{align*}
According to the Lipschitz condition of score estimates and Lemma \ref{lem:lipschitz}, we have
\begin{align*}
\frac{\big\|y_{m, \tau_{k, n}} - y_{\tau_{k, n}}\big\|_2}{\sqrt{1-\tau_{k, n}}} 
&\lesssim \frac{\min\{d\log T,L\}\log T}{T} \sum_{i = 1}^{n-1} \frac{\big\|y_{m-1, \tau_{k,i}} - y_{\tau_{k,i}}\big\|_2}{\sqrt{1-\tau_{k,i}}},\qquad  y_{\tau_{k, 0}}\in\mathcal{E}_k.
\end{align*}
Applying the above relation recursively gives
\begin{align*}
\max_n\frac{\big\|y_{M, \tau_{k, n}} - y_{\tau_{k, n}}\big\|_2}{\sqrt{1-\tau_{k, n}}} 
\le \bigg(\frac{N\min\{d\log T,L\}\log T}{T}\bigg)^M\max_n\frac{\big\|y_{0, \tau_{k, n}} - y_{\tau_{k, n}}\big\|_2}{\sqrt{1-\tau_{k, n}}} 
\le \frac{1}{\mathsf{poly}(T)},
\end{align*}
provided that $M \gtrsim \log T$ and $T \gtrsim N\min\{d\log T,L\}\log T$.
Thus as long as 
$$
N \gtrsim \frac{(\min\{d^{2/3}L^{-2/3},d^{1/3}\}+1)\log^{5/3} T}{\varepsilon^{2/3}},
$$
which guarantees that $T \gtrsim \frac{\min\{d,d^{2/3}L^{1/3},d^{1/3}L\}\log^{8/3} T}{\varepsilon^{2/3}}$,
we can get the desired result immediately through just inserting the above error bound into Lemma~\ref{lem:KL} and Lemma \ref{lem:endpoint}.
The probability of $P(\widehat{X}_k\in\mathcal{E}_k^{\rm c})$ is bounded by using Lemma \ref{lem:ratio-pxy} and the fact that $P({x}_{\tau_{k,n}}\in\mathcal{L}_{\tau_{k,n}}^{\rm c})\lesssim 1/T^4$. 
We omit the details here due to the similarity.

\section{Proof of auxiliary lemmas in Theorem~\ref{thm:main}}

\noindent \textbf{Proof of \eqref{eq:diff-tau-2}.}
According to the definition of $\widehat{\tau}_{k,i}$, we have that
\begin{align*}
\frac{1-\widehat{\tau}_{k, N}}{1-\widehat{\tau}_{k,-1}}
=\frac{\widehat{\alpha}_{T-\frac{kN}{2}-N}}{\widehat{\alpha}_{T-\frac{kN}{2}+1}}
\le \left(1+\frac{c_1\log T}{T}\right)^{N+1}\le \exp\left(\frac{c_1(N+1)\log T}{T}\right) \le {\rm e},
\end{align*}
and
\begin{align*}
\frac{\widehat{\tau}_{k, -1}}{\widehat{\tau}_{k,N}}
= \frac{1-\widehat{\alpha}_{T-\frac{kN}{2}+1}}{1-\widehat{\alpha}_{T-\frac{kN}{2}-N}}
\le \left(1-\frac{c_1\log T}{T}\right)^{-N-1} \le \exp\left(\frac{2c_1(N+1)\log T}{T}\right) \le {\rm e},
\end{align*}
as long as $T\ge 4c_1N\log T$.
Thus we complete the proof.

\subsection{Proof of Lemma \ref{lem:ODE}}
\label{sec:proof-lem:ODE}

Notice that the map $\Phi$ in~\eqref{eq:lem-ODE} is just the integral form of~\eqref{eq:ODE}, which is equivalent to 
\begin{align*}
\mathrm{d} x_{\tau} = -\frac{1}{2(1-\tau)}\big(x_{\tau} + s_{\tau}^{\star}(x_{\tau})\big) \mathrm{d} \tau.
\end{align*}
This is the well-known probability ODE flow, which comes from \citet{song2021denoising} and is also used in \citet{Li2023Towards}.

The proof of \eqref{eq:lem-distribution} can be completed by using mathematical induction.
Recalling that $\widehat{X}_0\overset{d}{=}X_{\tau_{0,0}}$, the \eqref{eq:lem-distribution} holds for $k+1=0$.
Assume that \eqref{eq:lem-distribution} holds for $k+1=h$.
We have $\Phi_{\tau_{h-1, 0} \to \tau_{h-1, N}}(\widehat{X}_{h-1})\overset{d}{=}X_{\tau_{h-1,N}}$.
According to \eqref{eq:def-Xtau}, we could immediately get that \eqref{eq:lem-distribution} holds for $k=h$.

\subsection{Proof of Lemma \ref{lem:prop-seq}}
\label{subsec:proof-lem-prop-seq}

We shall complete the remaining proof by mathematical induction.
According to initializations, all inequalities in Lemma \ref{lem:prop-seq} hold for $k=0$.
Assume \eqref{eq:lem-prop-seq} hold for $k=h$.
For $x,y\in \mathcal{E}_{h+1}^{\rm c}$, we have
$p_{\widetilde{X}_{h+1}}(x) = 0 \le p_{\widehat{X}_{h+1}}(x)$, $p_{\widetilde{Y}_{h+1}}(y)=0\le p_{Y_{h+1}}(y)$.
For $x,y\in\mathcal{E}_{h+1}$,
\begin{align*}
p_{\widetilde{X}_{h+1}}(x) &= \int p_{\widetilde{X}_{h+1}\mymid \widetilde{X}_{h}}(x\mymid x_h) p_{\widetilde{X}_{h}}(x_h) \mathrm d x_h \le \int p_{\widehat{X}_{h+1}\mymid \widehat{X}_{h}}(x\mymid x_h) p_{\widehat{X}_{h}}(x_h) \mathrm d x_h 
= p_{\widehat{X}_h}(x), \\
% p_{\overline{X}_{h+1}}(x)&\le p_{\overline{X}_{h+1}^{-}}(x)= \int p_{\overline{X}_{h+1}^{-}\mymid \overline{X}_{h}}(x\mymid x_h)p_{\overline{X}_{h}}(x_h) \mathrm{d} x_h\le \int p_{\widetilde{X}_{h+1}\mymid \widetilde{X}_{h}}(x\mymid x_h)p_{\widetilde{X}_{h}}(x_h) \mathrm{d} x_h = p_{\widetilde{X}_{h+1}}(x),\\
p_{\widetilde{Y}_{h+1}}(y)&=\int p_{\widetilde{Y}_{h+1}\mymid \widetilde{Y}_h}(y\mymid y_h) p_{\widetilde{Y}_h}(y_h) \mathrm{d} y_h \le  \int p_{{Y}_{h+1}\mymid {Y}_h}(y\mymid y_h) p_{{Y}_h}(y_h) \mathrm{d} y_h = p_{{Y}_{h+1}}(y).
\end{align*}

% \subsection{Proof of Lemma \ref{lem:endpoint}}
% \label{sec:proof-lem:endpoint}

% According to Lemma \ref{lem:ODE}, we know that $\widehat{X}_0 \overset{\mathrm{d}}{=} X_{\tau_{0,0}} = \sqrt{1-\tau_{0,0}}X_0 + \sqrt{\tau_{0,0}}Z$ for some standard Gaussian variable $Z$. 
% Then we have
% \begin{align*}
% \mathsf{KL}\big(p_{\widehat{X}_{0}}\Vert p_{Y_{0}}\big) \le& \mathsf{KL}\big(p_{X_{\tau_{0,0}},X_0}\Vert p_{Y_{0}}p_{X_0}\big) = \mathbb{E}_{X_0\sim p_{\mathsf{data}}}[\mathsf{KL}\big(p_{\widehat{X}_{0}|X_0}\Vert p_{Y_{0}}\big)]\nonumber\\
% =&\frac12\left[(1-\tau_{0,0})\mathbb{E}_{X_0\sim p_{\mathsf{data}}}\left\|X_0\right\|^2-d\log(\tau_{0,0})+d\tau_{0,0}-d\right]\\
% \le& \frac12T^{c_R-c_0}+d\frac{(1-\tau_{0,0})^2}{2\tau_{0,0}}
% \le \frac12T^{c_R-c_0}+\frac{d}{T^{2c_0}} \le \frac{1}{T^{10}},
% \end{align*}
% as long as $c_0\ge \max\{c_R+10,10\}$, and $T\gtrsim d^{1/10}$.
% In addition, the analysis of $\mathsf{TV}(p_{\mathsf{data}},p_{\widehat{X}_K})$ can be found in the proof of Theorem 2 in \citep{Chen2023The}.

\subsection{Proof of Lemma \ref{lem:ratio-pxy}}
\label{subsec:proof-lem-ratio-pxy}
%Denote $\mathcal{R} = \{x_0\in\mathbb{R}^d: \left\|x-\sqrt{1-\tau}x_0\right\|\le\left\|y-\sqrt{1-\tau}x_0\right\|\}$.
%Without loss of generality, we assume that $p_\tau(x)\ge p_\tau (y)$.
Notice that
\begin{align*}
p_{X_\tau}(y) 
&= \int p_{X_\tau|X_0}(y|x_0) p_{X_0}(x_0) \mathrm d x_0\nonumber\\
&=\int p_{X_\tau\mymid X_0}(x\mymid x_0)p_{X_0}(x_0) \exp\left(\frac{(x-y)^{\top}(x+y-2\sqrt{1-\tau}x_0)}{2\tau}\right) \mathrm d x_0,\\
p_{X_\tau}(x) &=\int p_{X_\tau\mymid X_0}(y\mymid x_0)p_{X_0}(x_0) \exp\left(\frac{(y-x)^{\top}(x+y-2\sqrt{1-\tau}x_0)}{2\tau}\right) \mathrm d x_0.
\end{align*}
For $x\in\mathcal{S}_{\tau}$, we have
\begin{align*}
&\quad p_{X_\tau}(x) - p_{X_\tau}(y)\notag\\
&= \int p_{X_\tau|X_0}(x|x_0)P_{X_0}(x_0)\left(1-\exp\left(\frac{(x-y)^{\top}(x+y-2\sqrt{1-\tau}x_0)}{2\tau}\right)\right)  \mathrm d x_0\nonumber\\
&\le \int p_{X_\tau|X_0}(x|x_0)P_{X_0}(x_0)\frac{(y-x)^{\top}(x+y-2\sqrt{1-\tau}x_0)}{2\tau}  \mathrm d x_0\nonumber\\
&\le \int p_{X_\tau|X_0}(x|x_0)P_{X_0}(x_0)\frac{\|y-x\|^2}{2\tau}  \mathrm d x_0\nonumber\\
&\quad +\int p_{X_\tau|X_0}(x|x_0)P_{X_0}(x_0)\frac{(y-x)^{\top}(x-\sqrt{1-\tau}x_0)}{\tau}  \mathrm d x_0\nonumber\\
&\le\frac{\|y-x\|^2}{2\tau}p_{X_\tau}(x) + \frac{\left\|y-x\right\|}{\tau}\int p_{X_\tau|X_0}(x|x_0)P_{X_0}(x_0)\left\|x-\sqrt{1-\tau}x_0\right\|  \mathrm d x_0.
\end{align*}
Furthermore, according to Lemma \ref{lem:bound-s-J}, and the fact that $\|x-y\|\le 2\sqrt{(\theta+c_0)d\tau\log T}$, we have
\begin{align*}
&\quad p_{X_\tau}(x) - p_{X_\tau}(y)\notag\\
&\overset{\text{}}{\le} \left(\frac{\|y-x\|^2}{2\tau} + 5\|y-x\|\sqrt{\frac{(\theta+c_0)d\log T}{\tau}}\right)p_{X_{\tau}}(x) \nonumber\\
&\overset{\text{}}{\le} 6\|y-x\|\sqrt{\frac{(\theta+c_0)d\log T}{\tau}}p_{X_{\tau}}(x).
\end{align*}
%where (a) uses Lemma \ref{lem:bound-s-J}, and (b) uses $\|x-y\|\le 2\sqrt{(\theta+c_0)d\tau\log T}$.
Similarly, for $y\in\mathcal{S}_{\tau}$, we have
\begin{align*}
p_{X_\tau}(y) - p_{X_\tau}(x) \le  6\|y-x\|\sqrt{\frac{(\theta+c_0)d\log T}{\tau}}p_{X_{\tau}}(y).
\end{align*}
Thus we complete the proof.

\subsection{Proof of Lemma \ref{lem:learning}}
\label{sec:proof-lem:learning}
\noindent \textbf{Proof of \eqref{eq:diff-tau-1}.}
The first relation is immediately obtained by noticing that 
\begin{align*}
1-\tau_{0,0}=\overline{\alpha}_{T+1}\le \widehat{\alpha}_T \le 2\widehat{\alpha}_{T+1},
\end{align*}
provided that $\frac{c_1\log T}{T} < 1$.

Regarding the second inequality, for $T_0 = (1-\frac{3c_0}{c_1})T+1$ and $c_1>3c_0$, we claim that $\widehat{\alpha}_{T_0} \ge \frac{1}{2}$; otherwise, we have
\begin{align*}
\widehat{\alpha}_{t-1} 
= \widehat{\alpha}_{t} \Big(1 + \frac{c_1(1-\widehat{\alpha}_{t})\log T}{T}\Big)
> \widehat{\alpha}_{t} \Big(1 + \frac{c_1\log T}{2T}\Big),
\end{align*}
and then
\begin{align*}
\widehat{\alpha}_{T_0} 
> \widehat{\alpha}_{T+1} \Big(1 + \frac{c_1\log T}{2T}\Big)^{T-T_0+1} > \frac{1}{2}.
\end{align*}
Then if $c_1>5c_0$, we have
\begin{align*}
\tau_{K, 0} =1-\overline{\alpha}_1 \le 1- \widehat{\alpha}_{1}
\le (1-\widehat{\alpha}_{T_0})\left(1-\frac{c_1\log T}{2T}\right)^{T_0-1}
\le \frac{1}{T^{c_0}},
\end{align*}
where we make use of the observation that for $t \le T_0$,
\begin{align*}
1-\widehat{\alpha}_{t-1} = (1-\widehat{\alpha}_{t})\Big(1 - \frac{c_1\widehat{\alpha}_{t}\log T}{T}\Big) \le (1-\widehat{\alpha}_{t})\Big(1 - \frac{c_1\log T}{2T}\Big).
\end{align*}

The third equation follows immediately from the definition of $\widehat{\tau}_{k,n}$ that
\begin{align*}
\frac{\widehat{\tau}_{k,n-1} - \widehat{\tau}_{k,n}}{\widehat{\tau}_{k,n-1}(1-\widehat{\tau}_{k, n-1})}  = \frac{\widehat{\alpha}_{T-\frac{kN}{2}-n} - \widehat{\alpha}_{T-\frac{kN}{2}-n+1}}{(1-\widehat{\alpha}_{T-\frac{kN}{2}-n+1})\widehat{\alpha}_{T-\frac{kN}{2}-n+1}} 
= \frac{c_1\log T}{T}.
\end{align*}

\bibliographystyle{apalike}
\bibliography{refs}

\end{document}